\RequirePackage{fix-cm}
\makeatletter
\def\cl@chapter{}
\makeatother
\documentclass[smallcondensed]{svjour3}     
\smartqed  
\usepackage{amsmath,amssymb,amsfonts}
\usepackage{url}
\usepackage{cleveref}
\usepackage{graphicx}
\usepackage{natbib}
\usepackage{booktabs}
\usepackage{algorithm}
\usepackage[noend]{algpseudocode}

\usepackage{tikz}
\usepackage{pgfplots}
\usepgfplotslibrary{groupplots}
\usepackage{aircraftshapes}

\usepackage{makecell}
\usepackage{siunitx}

\usetikzlibrary{decorations.pathreplacing}

\pgfplotsset{compat=newest}
\pgfplotsset{every axis legend/.append style={legend cell align=left}}

\pgfplotsset{every axis/.append style={
                    title style={font=\small},
                    tick label style={font=\footnotesize}  
                    }}
\pgfplotsset{every axis label/.style={font=\small}}                    
\pgfplotsset{
legend image code/.code={
\draw[mark repeat=2,mark phase=2]
plot coordinates {
(0cm,0cm)
(0.15cm,0cm)        
(0.3cm,0cm)         
};%
}
}

\definecolor{ra_1}{rgb}{1.0, 1.0, 1.0}
\definecolor{ra_2}{rgb}{0.0, 1.0, 1.0}
\definecolor{ra_3}{rgb}{0.56, 0.93, 0.56}
\definecolor{ra_4}{rgb}{0.12, 0.56, 1.0}
\definecolor{ra_5}{rgb}{0.0, 1.0, 0.0}
\definecolor{ra_6}{rgb}{0.0, 0.0, 1.0}
\definecolor{ra_7}{rgb}{0.13, 0.53, 0.13}
\definecolor{ra_8}{rgb}{0.0, 0.0, 0.50}
\definecolor{ra_9}{rgb}{0.0, 0.39, 0.0}

\definecolor{up}{rgb}{0.75, 0.75, 0.75}
\definecolor{down}{rgb}{0.56, 0.93, 0.56}
\definecolor{left}{rgb}{0.898, 0.24, 0.52}
\definecolor{right}{rgb}{0.0, 0.50, 1.0}

\newcommand{\argmax}{\operatornamewithlimits{arg\,max}}

\renewcommand{\Pr}{\text{Pr}}

\begin{document}

\title{Generating Probabilistic Safety Guarantees \\ for Neural Network Controllers
}


\author{Sydney M. Katz  \and
        Kyle D. Julian  \and
        Christopher A. Strong \and
        Mykel J. Kochenderfer
}


\institute{S.M. Katz (corresponding author), K.D. Julian, M.J. Kochenderfer \at
              Stanford University, Department of Aeronautics and Astronautics \\
              Stanford, CA, USA \\
              \email{\{smkatz, kjulian3, mykel\}@stanford.edu}            \\
              ORCID: 0000-0001-8376-5145, 0000-0002-6247-1874, 0000-0002-7238-9663
           \and
           C.A. Strong \at
              Stanford University, Department of Electrical Engineering \\
              Stanford, CA, USA \\
              \email{castrong@stanford.edu}           \\
              ORCID: 0000-0002-8914-6852
}

\date{Received: date / Accepted: date}

\maketitle

\begin{abstract}
Neural networks serve as effective controllers in a variety of complex settings due to their ability to represent expressive policies. The complex nature of neural networks, however, makes their output difficult to verify and predict, which limits their use in safety-critical applications. While simulations provide insight into the performance of neural network controllers, they are not enough to guarantee that the controller will perform safely in all scenarios. To address this problem, recent work has focused on formal methods to verify properties of neural network outputs. For neural network controllers, we can use a dynamics model to determine the output properties that must hold for the controller to operate safely. In this work, we develop a method to use the results from neural network verification tools to provide probabilistic safety guarantees on a neural network controller. We develop an adaptive verification approach to efficiently generate an overapproximation of the neural network policy. Next, we modify the traditional formulation of Markov decision process (MDP) model checking to provide guarantees on the overapproximated policy given a stochastic dynamics model. Finally, we incorporate techniques in state abstraction to reduce overapproximation error during the model checking process. We show that our method is able to generate meaningful probabilistic safety guarantees for aircraft collision avoidance neural networks that are loosely inspired by Airborne Collision Avoidance System X (ACAS X), a family of collision avoidance systems that formulates the problem as a partially observable Markov decision process (POMDP).

\end{abstract}

\section{Introduction}
\label{intro}
Neural networks provide a means to represent complex control policies, 
making them particularly useful in complicated problem domains where an agent must make decisions over a large input space \citep{mnih2015human}. 
Recently, neural networks have been proposed as controllers in safety-critical applications such as aircraft collision avoidance and autonomous driving \citep{Julian2016dasc, Julian2019jgcd, Bouton2020phd, pan2017agile}. 
Using neural networks in these settings presents major challenges. Due to the inherent complexity and unpredictable nature of neural networks, they are difficult to certify for use in safety-critical applications. Performance in Monte Carlo simulations is not enough to guarantee that the network will provide safe actions in all scenarios. To this end, recent work in formal methods has resulted in tools for verifying properties of neural networks \citep{Katz2017, katz2019marabou, reluval, tjeng2017evaluating}. Given a bounded input set, these tools provide guarantees on characteristics of the output set.

Using neural network verification tools to prove properties in this manner represents a step towards the ability to certify neural networks as safe; however, these works simply check input-output properties specified by a human designer without considering the closed-loop behavior of the system. In order to guarantee safety for neural network controllers, there is a need for a more principled approach to selecting the properties that a neural network must satisfy for safe operation. By taking into account a dynamic model, we can create a closed-loop system that describes the effect of the neural network controller's actions on its environment. We can then use this system to better understand what constitutes a ``safe'' neural network output. Previous work has evaluated the safety of the closed-loop system using various forms of reachability analysis \citep{Julian2019reach, Julian2019dasc, huang2019reachnn, xiang2018reachability, xiang2018reachable, xiang2019reachable, dutta2019reachability, ivanov2019verisig, claviere2020safety, manzanas2021verification}.

One limitation of the reachability approaches used in previous work is their inability to properly account for stochasticity in the system dynamics, which is present in many real-world systems. For instance, a number of the approaches assume no uncertainty in the dynamics model when computing reachable sets \citep{claviere2020safety, manzanas2021verification}. While other work takes into account this uncertainty by overapproximating the system dynamics, the binary nature of the output of this analysis does not properly reflect the stochastic nature of the dynamics model \citep{Julian2019dasc}. In particular, this analysis simply flags states as reachable without specifying the likelihood of reaching them. Therefore, even if the probability of reaching an unsafe state is extremely low, this technique would mark the overall system as unsafe.

In this work, instead of determining solely whether unsafe states are reachable, we develop a method that takes as input a stochastic dynamics model and provides probabilistic safety guarantees on the closed-loop system. Similar to \citet{Julian2019reach}, we divide the input space into small cells and run each input region through a neural network verification tool \citep{Julian2019dasc, Julian2019reach}. 
Using the results of the neural network verification tool to define an action space, we formulate the closed-loop verification problem as a Markov decision process (MDP). 
This formulation allows us to draw upon techniques from MDP model checking to approximate the probability of reaching an unsafe state from any particular cell given a probabilistic model of the dynamics \citep{baier2008principles, lahijanian2011control, Bouton2020aaai, Bouton2020phd}. 

We modify the model checking formulation to ensure an overapproximation of the probabilities and outline both online and offline methods to reduce overapproximation error. Specifically, we develop an adaptive verification method that addresses limitations mentioned in previous work to efficiently divide the input region into cells \citep{Julian2019dasc}. We show that this method better approximates the decision boundaries of the neural network and processes the input space faster than a na\"ive approach to state space partition. We further reduce overapproximation error by using ideas from state abstraction to split safety-critical regions of the state space during the solving process \citep{munos2002variable}. Our contributions are summarized as follows.
\begin{itemize}
    \item We show how to adapt techniques in MDP model checking to generate probabilistic safety guarantees on neural network controllers operating in environments with stochastic dynamics.
    \item We create a method to obtain an overapproximated neural network policy using existing neural network verification tools. Specifically, we introduce an adaptive verification approach to automatically partition the input space in a way that reduces overapproximation error.
    \item We show how to use techniques in state abstraction to reduce overapproximation error in the estimated probability during the model checking process.
\end{itemize}
We apply our method to aircraft collision avoidance neural networks and show that we can use it to provide meaningful safety guarantees on a neural network controller.

\section{Background}
In this work, we formulate the closed-loop verification problem as a Markov decision process (MDP) and use this formulation to apply techniques in probabilistic model checking to evaluate the safety of a neural network controller. This section outlines the necessary background on MDPs and probabilistic model checking.

\subsection{Markov Decision Process}
An MDP is a way of encoding a sequential decision making problem where an agent's action at each time step depends only on its current state \citep{Kochenderfer2015}. An MDP is defined by the tuple $(\mathcal{S}, \mathcal{A}, T, R, \gamma)$, where $\mathcal{S}$ is the state space, $\mathcal{A}$ is the action space, $T(s, a, s')$ is the probability of transitioning to state $s'$ given that we are in state $s$ and take action $a$, $R(s,a)$ is the reward for taking action $a$ in state $s$, and $\gamma$ is the discount factor. Using this formulation, we can solve for a policy $\pi$ that maps states to actions. To do so, we define an action-value function $Q(s,a)$ that represents the discounted sum of expected future rewards when taking action $a$ from state $s$. The optimal action-value function $Q^\ast(s,a)$ can be found using a form of dynamic programming called value iteration. Value iteration relies on iterative updates using the Bellman equation \citep{bellman1952theory}:
\begin{equation}
\label{eq:bellman}
    Q^\ast(s,a) = R(s,a) + \gamma \sum_{s' \in S} T(s,a,s') \max_{a' \in A} Q^\ast(s',a')
\end{equation}
We can extract the policy from the action-value function by simply choosing the action with the maximum value at state $s$:
\begin{equation}
    \pi(s) = \argmax_{a \in A} Q^\ast(s,a)
    \label{detpol}
\end{equation}

\subsection{Probabilistic Model Checking}\label{sec:model_check}
Probabilistic model checking for MDPs has been well studied and often involves determining the probability of satisfying a property expressed using Linear Temporal Logic (LTL) \citep{baier2008principles, lahijanian2011control, Bouton2020aaai}. An LTL formula consists of atomic propositions connected by logical or temporal operators \citep{baier2008principles}. Our goal is to assign a probability $\Pr^{\pi}(s)$ of satisfying the LTL specification to each state $s \in \mathcal{S}$. For any LTL formula, this computation reduces to a reachability problem for a set of states $\mathcal{B}$ \citep{baier2008principles, Bouton2020phd}. In traditional MDP model checking, we seek to find the maximum probability of reaching states in $\mathcal{B}$ while following policy $\pi$ from each state $s \in \mathcal{S}$. This probability can be written recursively as
\begin{equation}
\label{eq:bellmanProbValStates}
    \Pr^{\pi}(s) = \sum_{s' \in \mathcal{S}} T(s' \mid s, \pi(s)) \Pr^{\pi}(s')
\end{equation}
for all states $s \notin \mathcal{B}$. All states $s \in \mathcal{B}$ are assigned a probability of one. 

\Cref{eq:bellmanProbValStates} can also be written in a form that is analogous to the action-value function in \cref{eq:bellman} to represent the probability of satisfying the LTL formula when action $a$ is taken from state $s$ as follows
\begin{equation}
\label{eq:bellmanProbQ}
    \Pr^{\pi}(s, a) =  \sum_{s' \in \mathcal{S}} T(s' \mid s, a) \Pr^{\pi}(s', \pi(s'))
\end{equation}
Noting the similarity between \cref{eq:bellman} and \cref{eq:bellmanProbQ}, we can solve for the probabilities using value iteration. The problem reduces to solving for the value function of an MDP with a modified reward function to represent probabilities \citep{Bouton2020aaai, Bouton2020phd}. The immediate reward is one for being in a state in $\mathcal{B}$ and zero for being in any other state.


\section{Approach}
We assume that we are given a neural network controller that represents the value function for an MDP policy $\pi$, which maps points in a bounded input space $\mathcal{S}$ to an action in the action space $\mathcal{A}$. We also assume that we are given a stochastic dynamics model in the form of a transition model $T(s' \mid s, a)$ and a safety specification on the closed-loop system written in the form of an LTL formula. Using this information, our goal is to determine the probability that the neural network controller satisfies this specification from each state $s \in \mathcal{S}$. If the input space $\mathcal{S}$ were discrete, we could directly apply the technique outlined in \cref{sec:model_check} to determine these probabilities. However, because neural network controllers typically take in a continuous range of states, we must introduce approximations into the model checking process to handle the continuous input space.

We make these approximations by partitioning the input space $\mathcal{S}$ into a finite number of smaller regions called cells, $c \in \mathcal{C}$, and modifying the model checking process to work with cells rather than states. We break the problem into two steps. The first step involves using a neural network verification tool to obtain an overapproximated neural network policy $\tilde{\pi}$ that operates on cells in $\mathcal{C}$. Using this policy, the second step uses probabilistic model checking to generate an overapproximated probability of reaching an unsafe state from each cell in $\mathcal{C}$. For both steps, we develop techniques to reduce overapproximation error in the estimated probabilities, which we outline in \cref{sec:reduce_overappox}.

\begin{figure}[htb]
    \begin{tikzpicture}[]
    
    \filldraw[color=black, fill=up ,line width=0.25mm] (8.2,3.0) rectangle (8.5,3.3);
    \node at (8.8, 3.15) {Up};
    
    \filldraw[color=black, fill=down ,line width=0.25mm] (8.2,2.5) rectangle (8.5,2.8);
    \node at (9.0, 2.65) {Down};
    
    \filldraw[color=black, fill=left ,line width=0.25mm] (8.2,2.0) rectangle (8.5,2.3);
    \node at (8.9, 2.15) {Left};
    
    \filldraw[color=black, fill=right ,line width=0.25mm] (8.2,1.5) rectangle (8.5,1.8);
    \node at (9.0, 1.65) {Right};
    
    \begin{groupplot}[group style={horizontal sep=1.0cm, group size=2 by 1}]
    
    \nextgroupplot [
      height = {5cm},
      ylabel = {$y$},
      title = {Continuum World Problem},
      xmin = {0.0},
      xmax = {20.0},
      axis equal image = {true},
      ymax = {20.0},
      xlabel = {$x$},
      ymin = {0.0}
    ]
    
    \fill [red, opacity=1.0] (axis cs:8.0,8.0) rectangle (axis cs:12.0,12.0);;
    
    \fill [green, opacity=1.0] (axis cs:19.0,19.0) rectangle (axis cs:20.0,20.0);;

    \nextgroupplot [
      height = {5cm},
      ylabel = {$y$},
      title = {Neural Network Actions},
      xmin = {0.0},
      xmax = {20.0},
      axis equal image = {true},
      ymax = {20.0},
      xlabel = {$x$},
      ymin = {0.0},
      enlargelimits = false,
      axis on top
    ]
    
    \addplot[
      ] graphics[
      xmin = 0.0,
      xmax = 20.0,
      ymin = 0.0,
      ymax = 20.0
    ] {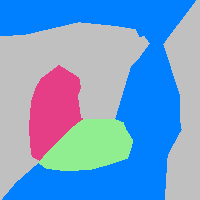};
    
    \draw[black,line width=0.25mm] (17.5,17.5) rectangle (20.0,20.0);;
    
    \draw[black,line width=0.25mm] (17.5,15.0) rectangle (20.0,17.5);;
    
    \draw[black,line width=0.25mm] (15.0,17.5) rectangle (17.5,20.0);;
    
    \draw[black,line width=0.25mm] (15.0,15.0) rectangle (17.5,17.5);;
    
    \draw[black,line width=0.25mm] (17.5,12.5) rectangle (20.0,15.0);;
    
    \draw[black,line width=0.25mm] (17.5,10.0) rectangle (20.0,12.5);;
    
    \draw[black,line width=0.25mm] (15.0,12.5) rectangle (17.5,15.0);;
    
    \draw[black,line width=0.25mm] (15.0,10.0) rectangle (17.5,12.5);;
    
    \draw[black,line width=0.25mm] (12.5,17.5) rectangle (15.0,20.0);;
    
    \draw[black,line width=0.25mm] (12.5,15.0) rectangle (15.0,17.5);;
    
    \draw[black,line width=0.25mm] (10.0,17.5) rectangle (12.5,20.0);;
    
    \draw[black,line width=0.25mm] (10.0,15.0) rectangle (12.5,17.5);;
    
    \draw[black,line width=0.25mm] (12.5,12.5) rectangle (15.0,15.0);;
    
    \draw[black,line width=0.25mm] (12.5,10.0) rectangle (15.0,12.5);;
    
    \draw[black,line width=0.25mm] (10.0,12.5) rectangle (12.5,15.0);;
    
    \draw[black,line width=0.25mm] (10.0,10.0) rectangle (12.5,12.5);;
    
    \draw[black,line width=0.25mm] (17.5,7.5) rectangle (20.0,10.0);;
    
    \draw[black,line width=0.25mm] (17.5,5.0) rectangle (20.0,7.5);;
    
    \draw[black,line width=0.25mm] (15.0,7.5) rectangle (17.5,10.0);;
    
    \draw[black,line width=0.25mm] (15.0,5.0) rectangle (17.5,7.5);;
    
    \draw[black,line width=0.25mm] (17.5,2.5) rectangle (20.0,5.0);;
    
    \draw[black,line width=0.25mm] (17.5,0.0) rectangle (20.0,2.5);;
    
    \draw[black,line width=0.25mm] (15.0,2.5) rectangle (17.5,5.0);;
    
    \draw[black,line width=0.25mm] (15.0,0.0) rectangle (17.5,2.5);;
    
    \draw[black,line width=0.25mm] (12.5,7.5) rectangle (15.0,10.0);;
    
    \draw[black,line width=0.25mm] (12.5,5.0) rectangle (15.0,7.5);;
    
    \draw[black,line width=0.25mm] (10.0,7.5) rectangle (12.5,10.0);;
    
    \draw[black,line width=0.25mm] (10.0,5.0) rectangle (12.5,7.5);;
    
    \draw[black,line width=0.25mm] (12.5,2.5) rectangle (15.0,5.0);;
    
    \draw[black,line width=0.25mm] (12.5,0.0) rectangle (15.0,2.5);;
    
    \draw[black,line width=0.25mm] (10.0,2.5) rectangle (12.5,5.0);;
    
    \draw[black,line width=0.25mm] (10.0,0.0) rectangle (12.5,2.5);;
    
    \draw[black,line width=0.25mm] (7.5,17.5) rectangle (10.0,20.0);;
    
    \draw[black,line width=0.25mm] (7.5,15.0) rectangle (10.0,17.5);;
    
    \draw[black,line width=0.25mm] (5.0,17.5) rectangle (7.5,20.0);;
    
    \draw[black,line width=0.25mm] (5.0,15.0) rectangle (7.5,17.5);;
    
    \draw[black,line width=0.25mm] (7.5,12.5) rectangle (10.0,15.0);;
    
    \draw[black,line width=0.25mm] (7.5,10.0) rectangle (10.0,12.5);;
    
    \draw[black,line width=0.25mm] (5.0,12.5) rectangle (7.5,15.0);;
    
    \draw[black,line width=0.25mm] (5.0,10.0) rectangle (7.5,12.5);;
    
    \draw[black,line width=0.25mm] (2.5,17.5) rectangle (5.0,20.0);;
    
    \draw[black,line width=0.25mm] (2.5,15.0) rectangle (5.0,17.5);;
    
    \draw[black,line width=0.25mm] (0.0,17.5) rectangle (2.5,20.0);;
    
    \draw[black,line width=0.25mm] (0.0,15.0) rectangle (2.5,17.5);;
    
    \draw[black,line width=0.25mm] (2.5,12.5) rectangle (5.0,15.0);;
    
    \draw[black,line width=0.25mm] (2.5,10.0) rectangle (5.0,12.5);;
    
    \draw[black,line width=0.25mm] (0.0,12.5) rectangle (2.5,15.0);;
    
    \draw[black,line width=0.25mm] (0.0,10.0) rectangle (2.5,12.5);;
    
    \draw[black,line width=0.25mm] (7.5,7.5) rectangle (10.0,10.0);;
    
    \draw[black,line width=0.25mm] (7.5,5.0) rectangle (10.0,7.5);;
    
    \draw[black,line width=0.25mm] (5.0,7.5) rectangle (7.5,10.0);;
    
    \draw[black,line width=0.25mm] (5.0,5.0) rectangle (7.5,7.5);;
    
    \draw[black,line width=0.25mm] (7.5,2.5) rectangle (10.0,5.0);;
    
    \draw[black,line width=0.25mm] (7.5,0.0) rectangle (10.0,2.5);;
    
    \draw[black,line width=0.25mm] (5.0,2.5) rectangle (7.5,5.0);;
    
    \draw[black,line width=0.25mm] (5.0,0.0) rectangle (7.5,2.5);;
    
    \draw[black,line width=0.25mm] (2.5,7.5) rectangle (5.0,10.0);;
    
    \draw[black,line width=0.25mm] (2.5,5.0) rectangle (5.0,7.5);;
    
    \draw[black,line width=0.25mm] (0.0,7.5) rectangle (2.5,10.0);;
    
    \draw[black,line width=0.25mm] (0.0,5.0) rectangle (2.5,7.5);;
    
    \draw[black,line width=0.25mm] (2.5,2.5) rectangle (5.0,5.0);;
    
    \draw[black,line width=0.25mm] (2.5,0.0) rectangle (5.0,2.5);;
    
    \draw[black,line width=0.25mm] (0.0,2.5) rectangle (2.5,5.0);;
    
    \draw[black,line width=0.25mm] (0.0,0.0) rectangle (2.5,2.5);;

    \end{groupplot}
    
    \end{tikzpicture}
    \caption{Continuum world explanatory example. The plot on the left shows the setup of the continuum world. The goal of the agent is to reach a point in the green area while avoiding the red area. The plot on the right shows an example neural network policy for this problem. \label{fig:cw_ex}}
\end{figure}
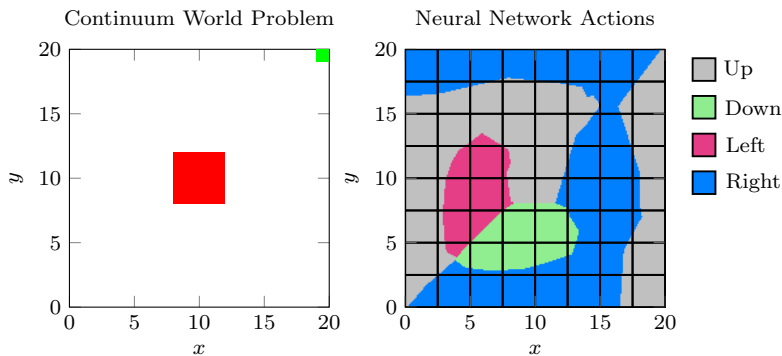

\Cref{fig:cw_ex} shows a simple example of a slippery continuum world that will be used to aid in our explanations of each step of our approach. In this example, the agent's objective is to reach a point within a set of goal states represented by the green region in the upper right corner without falling into a pit in the center of the world represented by the red region. The agent can select from four actions: up, down, left, or right. Because the world is slippery, taking an action results in a $70\%$ chance of moving one unit in the specified direction and a $10\%$ chance of moving one unit in each of the other directions. The plot on the right of \cref{fig:cw_ex} shows a sensible neural network policy for an agent in this world to follow along with a potential partition of the state space into cells. Our goal is to determine the overapproximated probability that an agent following this policy from each cell will fall into the pit.

\subsection{Policy Overapproximation}\label{sec:policy_overapprox}
The first step in modifying traditional MDP model checking to use cells involves defining a policy that takes a cell as input rather than a single state. Because each cell contains multiple states, it is possible for cells to map to multiple actions. For example, the cell in the bottom left corner of the policy plot in \cref{fig:cw_ex} contains some states that map to the right action and others that map to the up action. For each cell $c \in \mathcal{C}$, we use a neural verification tool to obtain the possible actions $\mathcal{A}_c \subseteq \mathcal{A}$ that the neural network could output for some point in $c$. The results provide an overapproximated policy $\tilde{\pi}$ that maps a cell $c$ to a subset of the action space $\mathcal{A}_c$ in contrast with $\pi$, which maps a specific state in the input space to a specific action. We assume that any point in $c$ could yield any action in $\mathcal{A}_c$. Therefore, any cell that has multiple actions in $\mathcal{A}_c$ contributes to an overapproximation of the neural network policy. Policy overapproximation is the first source of overapproximation error in the probability estimate.

\subsection{Model Checking}
Once we have an overapproximated policy, we can further modify the model checking framework to determine the probability of satisfying the LTL safety specification from each cell $c \in \mathcal{C}$. We first convert the LTL specification to a reachability problem for a set of states $\mathcal{B}$ either directly or by using the methods in \citet{baier2008principles}. In the continuum world example, $\mathcal{B}$ is the region of the state space that corresponds to the pit. Next, we adapt \cref{eq:bellmanProbValStates} to determine the probabilities for each cell $c \in \mathcal{C}$ using our overapproximated neural network policy $\tilde{\pi}$ as
\begin{equation}
\label{eq:bellmanProbVal}
    \Pr^{\tilde{\pi}}(c) = \max_{a \in \mathcal{A}_c} \sum_{c' \in \mathcal{C}} T(c' \mid c, a) \Pr^{\tilde{\pi}}(c')
\end{equation}
where $\mathcal{A}_c$ uses the neural network verification results and contains the set of actions that could be taken in cell $c$. All cells that overlap with $\mathcal{B}$ are assigned a probability of one. The maximization in \cref{eq:bellmanProbVal} corresponds to taking the worst-case action in $\mathcal{A}_c$.

The transition model, $T(c' \mid c, a)$, is modified to determine transitions between cells rather than states and to ensure that the resulting probabilities represent an overapproximation of the true probabilities. In this work, we restrict our approach to models in which taking an action results in a finite number of outcomes. We assume that taking action $a$ from state $s$ has $n$ possible outcomes with probabilities according to $T(s' \mid s, a)$. For example, in the continuum world, these outcomes would be the result of moving one unit up, down, left, and right. Let $p_i$ represent the probability of the $i$th outcome. 
Let $\mathcal{S}’_{1:n}$ represent regions of the state space that contain all possible next states from cell $c$ for each outcome $i \in 1, \ldots, n$. We define $\mathcal{C}'_{1:n}$ as the sets of cells that overlap with $\mathcal{S}'_{1:n}$. In order to preserve the overapproximation in our probabilities, we assign all of the probability for outcome $i$ to the worst-case cell in $\mathcal{C}'_i$ as follows
\begin{equation}
    \label{eq:cell_transition}
    T(c' \mid c, a) = \sum_i \begin{cases}
    \begin{split}
            p_i, & \text{   if } c' = \argmax_{c'' \in \mathcal{C}'_i} \Pr^{\tilde{\pi}}(c'') \\
            0, & \text{   otherwise}
    \end{split}
    \end{cases}
\end{equation}

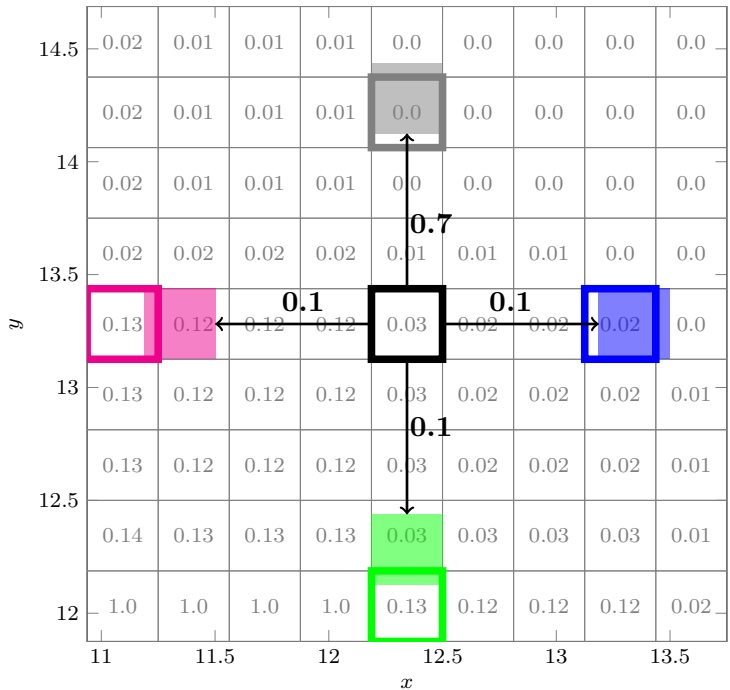
\begin{figure}[htb]
    \begin{tikzpicture}[]
    \begin{axis}[
      height = {10cm},
      ylabel = {$y$},
      xmin = {10.9375},
      xmax = {13.75},
      axis equal image = {true},
      ymax = {14.6875},
      xlabel = {$x$},
      ymin = {11.875}
    ]
    
    \draw[gray,line width=0.1mm] (13.4375,14.375) rectangle (13.75,14.6875);;
    
    \node[text = gray] at (13.59375, 14.53125) {0.0};;
    
    \draw[gray,line width=0.1mm] (13.125,14.375) rectangle (13.4375,14.6875);;
    
    \node[text = gray] at (13.28125, 14.53125) {0.0};;
    
    \draw[gray,line width=0.1mm] (13.4375,14.0625) rectangle (13.75,14.375);;
    
    \node[text = gray] at (13.59375, 14.21875) {0.0};;
    
    \draw[gray,line width=0.1mm] (13.4375,13.75) rectangle (13.75,14.0625);;
    
    \node[text = gray] at (13.59375, 13.90625) {0.0};;
    
    \draw[gray,line width=0.1mm] (13.125,14.0625) rectangle (13.4375,14.375);;
    
    \node[text = gray] at (13.28125, 14.21875) {0.0};;
    
    \draw[gray,line width=0.1mm] (13.125,13.75) rectangle (13.4375,14.0625);;
    
    \node[text = gray] at (13.28125, 13.90625) {0.0};;
    
    \draw[gray,line width=0.1mm] (12.8125,14.375) rectangle (13.125,14.6875);;
    
    \node[text = gray] at (12.96875, 14.53125) {0.0};;
    
    \draw[gray,line width=0.1mm] (12.5,14.375) rectangle (12.8125,14.6875);;
    
    \node[text = gray] at (12.65625, 14.53125) {0.0};;
    
    \draw[gray,line width=0.1mm] (12.8125,14.0625) rectangle (13.125,14.375);;
    
    \node[text = gray] at (12.96875, 14.21875) {0.0};;
    
    \draw[gray,line width=0.1mm] (12.8125,13.75) rectangle (13.125,14.0625);;
    
    \node[text = gray] at (12.96875, 13.90625) {0.0};;
    
    \draw[gray,line width=0.1mm] (12.5,14.0625) rectangle (12.8125,14.375);;
    
    \node[text = gray] at (12.65625, 14.21875) {0.0};;
    
    \draw[gray,line width=0.1mm] (12.5,13.75) rectangle (12.8125,14.0625);;
    
    \node[text = gray] at (12.65625, 13.90625) {0.0};;
    
    \draw[gray,line width=0.1mm] (13.4375,13.4375) rectangle (13.75,13.75);;
    
    \node[text = gray] at (13.59375, 13.59375) {0.0};;
    
    \draw[gray,line width=0.1mm] (13.4375,13.125) rectangle (13.75,13.4375);;
    
    \node[text = gray] at (13.59375, 13.28125) {0.0};;
    
    \draw[gray,line width=0.1mm] (13.125,13.4375) rectangle (13.4375,13.75);;
    
    \node[text = gray] at (13.28125, 13.59375) {0.0};;
    
    \draw[gray,line width=0.1mm] (13.125,13.125) rectangle (13.4375,13.4375);;
    
    \node[text = gray] at (13.28125, 13.28125) {0.02};;
    
    \draw[gray,line width=0.1mm] (13.4375,12.8125) rectangle (13.75,13.125);;
    
    \node[text = gray] at (13.59375, 12.96875) {0.01};;
    
    \draw[gray,line width=0.1mm] (13.4375,12.5) rectangle (13.75,12.8125);;
    
    \node[text = gray] at (13.59375, 12.65625) {0.01};;
    
    \draw[gray,line width=0.1mm] (13.125,12.8125) rectangle (13.4375,13.125);;
    
    \node[text = gray] at (13.28125, 12.96875) {0.02};;
    
    \draw[gray,line width=0.1mm] (13.125,12.5) rectangle (13.4375,12.8125);;
    
    \node[text = gray] at (13.28125, 12.65625) {0.02};;
    
    \draw[gray,line width=0.1mm] (12.8125,13.4375) rectangle (13.125,13.75);;
    
    \node[text = gray] at (12.96875, 13.59375) {0.01};;
    
    \draw[gray,line width=0.1mm] (12.8125,13.125) rectangle (13.125,13.4375);;
    
    \node[text = gray] at (12.96875, 13.28125) {0.02};;
    
    \draw[gray,line width=0.1mm] (12.5,13.4375) rectangle (12.8125,13.75);;
    
    \node[text = gray] at (12.65625, 13.59375) {0.01};;
    
    \draw[gray,line width=0.1mm] (12.5,13.125) rectangle (12.8125,13.4375);;
    
    \node[text = gray] at (12.65625, 13.28125) {0.02};;
    
    \draw[gray,line width=0.1mm] (12.8125,12.8125) rectangle (13.125,13.125);;
    
    \node[text = gray] at (12.96875, 12.96875) {0.02};;
    
    \draw[gray,line width=0.1mm] (12.8125,12.5) rectangle (13.125,12.8125);;
    
    \node[text = gray] at (12.96875, 12.65625) {0.02};;
    
    \draw[gray,line width=0.1mm] (12.5,12.8125) rectangle (12.8125,13.125);;
    
    \node[text = gray] at (12.65625, 12.96875) {0.02};;
    
    \draw[gray,line width=0.1mm] (12.5,12.5) rectangle (12.8125,12.8125);;
    
    \node[text = gray] at (12.65625, 12.65625) {0.02};;
    
    \draw[gray,line width=0.1mm] (13.4375,12.1875) rectangle (13.75,12.5);;
    
    \node[text = gray] at (13.59375, 12.34375) {0.01};;
    
    \draw[gray,line width=0.1mm] (13.4375,11.875) rectangle (13.75,12.1875);;
    
    \node[text = gray] at (13.59375, 12.03125) {0.02};;
    
    \draw[gray,line width=0.1mm] (13.125,12.1875) rectangle (13.4375,12.5);;
    
    \node[text = gray] at (13.28125, 12.34375) {0.03};;
    
    \draw[gray,line width=0.1mm] (13.125,11.875) rectangle (13.4375,12.1875);;
    
    \node[text = gray] at (13.28125, 12.03125) {0.12};;
    
    \draw[gray,line width=0.1mm] (12.8125,12.1875) rectangle (13.125,12.5);;
    
    \node[text = gray] at (12.96875, 12.34375) {0.03};;
    
    \draw[gray,line width=0.1mm] (12.8125,11.875) rectangle (13.125,12.1875);;
    
    \node[text = gray] at (12.96875, 12.03125) {0.12};;
    
    \draw[gray,line width=0.1mm] (12.5,12.1875) rectangle (12.8125,12.5);;
    
    \node[text = gray] at (12.65625, 12.34375) {0.03};;
    
    \draw[gray,line width=0.1mm] (12.5,11.875) rectangle (12.8125,12.1875);;
    
    \node[text = gray] at (12.65625, 12.03125) {0.12};;
    
    \draw[gray,line width=0.1mm] (12.1875,14.375) rectangle (12.5,14.6875);;
    
    \node[text = gray] at (12.34375, 14.53125) {0.0};;
    
    \draw[gray,line width=0.1mm] (11.875,14.375) rectangle (12.1875,14.6875);;
    
    \node[text = gray] at (12.03125, 14.53125) {0.01};;
    
    \draw[gray,line width=0.1mm] (12.1875,14.0625) rectangle (12.5,14.375);;
    
    \node[text = gray] at (12.34375, 14.21875) {0.0};;
    
    \draw[gray,line width=0.1mm] (12.1875,13.75) rectangle (12.5,14.0625);;
    
    \node[text = gray] at (12.34375, 13.90625) {0.0};;
    
    \draw[gray,line width=0.1mm] (11.875,14.0625) rectangle (12.1875,14.375);;
    
    \node[text = gray] at (12.03125, 14.21875) {0.01};;
    
    \draw[gray,line width=0.1mm] (11.875,13.75) rectangle (12.1875,14.0625);;
    
    \node[text = gray] at (12.03125, 13.90625) {0.01};;
    
    \draw[gray,line width=0.1mm] (11.5625,14.375) rectangle (11.875,14.6875);;
    
    \node[text = gray] at (11.71875, 14.53125) {0.01};;
    
    \draw[gray,line width=0.1mm] (11.25,14.375) rectangle (11.5625,14.6875);;
    
    \node[text = gray] at (11.40625, 14.53125) {0.01};;
    
    \draw[gray,line width=0.1mm] (11.5625,14.0625) rectangle (11.875,14.375);;
    
    \node[text = gray] at (11.71875, 14.21875) {0.01};;
    
    \draw[gray,line width=0.1mm] (11.5625,13.75) rectangle (11.875,14.0625);;
    
    \node[text = gray] at (11.71875, 13.90625) {0.01};;
    
    \draw[gray,line width=0.1mm] (11.25,14.0625) rectangle (11.5625,14.375);;
    
    \node[text = gray] at (11.40625, 14.21875) {0.01};;
    
    \draw[gray,line width=0.1mm] (11.25,13.75) rectangle (11.5625,14.0625);;
    
    \node[text = gray] at (11.40625, 13.90625) {0.01};;
    
    \draw[gray,line width=0.1mm] (12.1875,13.4375) rectangle (12.5,13.75);;
    
    \node[text = gray] at (12.34375, 13.59375) {0.01};;
    
    \draw[gray,line width=0.1mm] (12.1875,13.125) rectangle (12.5,13.4375);;
    
    \node[text = gray] at (12.34375, 13.28125) {0.03};;
    
    \draw[gray,line width=0.1mm] (11.875,13.4375) rectangle (12.1875,13.75);;
    
    \node[text = gray] at (12.03125, 13.59375) {0.02};;
    
    \draw[gray,line width=0.1mm] (11.875,13.125) rectangle (12.1875,13.4375);;
    
    \node[text = gray] at (12.03125, 13.28125) {0.12};;
    
    \draw[gray,line width=0.1mm] (12.1875,12.8125) rectangle (12.5,13.125);;
    
    \node[text = gray] at (12.34375, 12.96875) {0.03};;
    
    \draw[gray,line width=0.1mm] (12.1875,12.5) rectangle (12.5,12.8125);;
    
    \node[text = gray] at (12.34375, 12.65625) {0.03};;
    
    \draw[gray,line width=0.1mm] (11.875,12.8125) rectangle (12.1875,13.125);;
    
    \node[text = gray] at (12.03125, 12.96875) {0.12};;
    
    \draw[gray,line width=0.1mm] (11.875,12.5) rectangle (12.1875,12.8125);;
    
    \node[text = gray] at (12.03125, 12.65625) {0.12};;
    
    \draw[gray,line width=0.1mm] (11.5625,13.4375) rectangle (11.875,13.75);;
    
    \node[text = gray] at (11.71875, 13.59375) {0.02};;
    
    \draw[gray,line width=0.1mm] (11.5625,13.125) rectangle (11.875,13.4375);;
    
    \node[text = gray] at (11.71875, 13.28125) {0.12};;
    
    \draw[gray,line width=0.1mm] (11.25,13.4375) rectangle (11.5625,13.75);;
    
    \node[text = gray] at (11.40625, 13.59375) {0.02};;
    
    \draw[gray,line width=0.1mm] (11.25,13.125) rectangle (11.5625,13.4375);;
    
    \node[text = gray] at (11.40625, 13.28125) {0.12};;
    
    \draw[gray,line width=0.1mm] (11.5625,12.8125) rectangle (11.875,13.125);;
    
    \node[text = gray] at (11.71875, 12.96875) {0.12};;
    
    \draw[gray,line width=0.1mm] (11.5625,12.5) rectangle (11.875,12.8125);;
    
    \node[text = gray] at (11.71875, 12.65625) {0.12};;
    
    \draw[gray,line width=0.1mm] (11.25,12.8125) rectangle (11.5625,13.125);;
    
    \node[text = gray] at (11.40625, 12.96875) {0.12};;
    
    \draw[gray,line width=0.1mm] (11.25,12.5) rectangle (11.5625,12.8125);;
    
    \node[text = gray] at (11.40625, 12.65625) {0.12};;
    
    \draw[gray,line width=0.1mm] (10.9375,14.375) rectangle (11.25,14.6875);;
    
    \node[text = gray] at (11.09375, 14.53125) {0.02};;
    
    \draw[gray,line width=0.1mm] (10.9375,14.0625) rectangle (11.25,14.375);;
    
    \node[text = gray] at (11.09375, 14.21875) {0.02};;
    
    \draw[gray,line width=0.1mm] (10.9375,13.75) rectangle (11.25,14.0625);;
    
    \node[text = gray] at (11.09375, 13.90625) {0.02};;
    
    \draw[gray,line width=0.1mm] (10.9375,13.4375) rectangle (11.25,13.75);;
    
    \node[text = gray] at (11.09375, 13.59375) {0.02};;
    
    \draw[gray,line width=0.1mm] (10.9375,13.125) rectangle (11.25,13.4375);;
    
    \node[text = gray] at (11.09375, 13.28125) {0.13};;
    
    \draw[gray,line width=0.1mm] (10.9375,12.8125) rectangle (11.25,13.125);;
    
    \node[text = gray] at (11.09375, 12.96875) {0.13};;
    
    \draw[gray,line width=0.1mm] (10.9375,12.5) rectangle (11.25,12.8125);;
    
    \node[text = gray] at (11.09375, 12.65625) {0.13};;
    
    \draw[gray,line width=0.1mm] (12.1875,12.1875) rectangle (12.5,12.5);;
    
    \node[text = gray] at (12.34375, 12.34375) {0.03};;
    
    \draw[gray,line width=0.1mm] (12.1875,11.875) rectangle (12.5,12.1875);;
    
    \node[text = gray] at (12.34375, 12.03125) {0.13};;
    
    \draw[gray,line width=0.1mm] (11.875,12.1875) rectangle (12.1875,12.5);;
    
    \node[text = gray] at (12.03125, 12.34375) {0.13};;
    
    \draw[gray,line width=0.1mm] (11.875,11.875) rectangle (12.1875,12.1875);;
    
    \node[text = gray] at (12.03125, 12.03125) {1.0};;
    
    \draw[gray,line width=0.1mm] (11.5625,12.1875) rectangle (11.875,12.5);;
    
    \node[text = gray] at (11.71875, 12.34375) {0.13};;
    
    \draw[gray,line width=0.1mm] (11.5625,11.875) rectangle (11.875,12.1875);;
    
    \node[text = gray] at (11.71875, 12.03125) {1.0};;
    
    \draw[gray,line width=0.1mm] (11.25,12.1875) rectangle (11.5625,12.5);;
    
    \node[text = gray] at (11.40625, 12.34375) {0.13};;
    
    \draw[gray,line width=0.1mm] (11.25,11.875) rectangle (11.5625,12.1875);;
    
    \node[text = gray] at (11.40625, 12.03125) {1.0};;
    
    \draw[gray,line width=0.1mm] (10.9375,12.1875) rectangle (11.25,12.5);;
    
    \node[text = gray] at (11.09375, 12.34375) {0.14};;
    
    \draw[gray,line width=0.1mm] (10.9375,11.875) rectangle (11.25,12.1875);;
    
    \node[text = gray] at (11.09375, 12.03125) {1.0};;
    
    \fill [gray, opacity=0.5] (axis cs:12.1875,14.125) rectangle (axis cs:12.5,14.4375);;
    
    \fill [green, opacity=0.5] (axis cs:12.1875,12.125) rectangle (axis cs:12.5,12.4375);;
    
    \fill [magenta, opacity=0.5] (axis cs:11.1875,13.125) rectangle (axis cs:11.5,13.4375);;
    
    \fill [blue, opacity=0.5] (axis cs:13.1875,13.125) rectangle (axis cs:13.5,13.4375);;
    
    \draw[black,line width=1.0mm] (12.1875,13.125) rectangle (12.5,13.4375);;
    
    \draw[gray,line width=1.0mm] (12.1875,14.0625) rectangle (12.5,14.375);;
    
    \draw[green,line width=1.0mm] (12.1875,11.875) rectangle (12.5,12.1875);;
    
    \draw[magenta,line width=1.0mm] (10.9375,13.125) rectangle (11.25,13.4375);;
    
    \draw[blue,line width=1.0mm] (13.125,13.125) rectangle (13.4375,13.4375);;
    
    \draw[->, line width = 1.0] (12.34375, 13.4375) 
            -- (12.34375, 14.125);;
    
    \node[] at (12.45, 13.725) 
            {\large \textbf{0.7}};;
    
    \draw[->, line width = 1.0] (12.34375, 13.125) 
            -- (12.34375, 12.4375);;
    
    \node[] at (12.45, 12.825) 
            {\large \textbf{0.1}};;
    
    \draw[->, line width = 1.0] (12.1875, 13.28125) 
            -- (11.5, 13.28125);;
    
    \node[] at (11.8875, 13.38125) 
            {\large \textbf{0.1}};;
    
    \draw[->, line width = 1.0] (12.5, 13.28125) 
            -- (13.1875, 13.28125);;
    
    \node[] at (12.8, 13.38125) 
            {\large \textbf{0.1}};;
    
    \end{axis}
    \end{tikzpicture}
    \caption{Illustration of cell transitions for the continuum world example. Each cell is labeled with its corresponding probability estimate. Each shaded region shows the set of possible next states with their corresponding probabilities of being reached when the up action is taken from the cell highlighted in black. We assign all probability for each region to the worst-case cell that it overlaps with shown by the highlighted cells. \label{fig:transition_demo}}
\end{figure}

\Cref{eq:cell_transition} preserves the overapproximation by assuming that \emph{all} points in a cell transition to the worst-case next state realizable from \emph{any} point in the cell. \Cref{fig:transition_demo} shows a visual representation of the transition model for the continuum world adapted for use with cells. All cells are labeled with their probability estimates, and the figure shows the result of taking the up action from the cell highlighted in black. The shaded regions represent $\mathcal{S}'_{1:4}$. The highlighted cells represent the worst-case cells that overlap each region $\mathcal{S}'_i$.

\begin{figure}[htb]
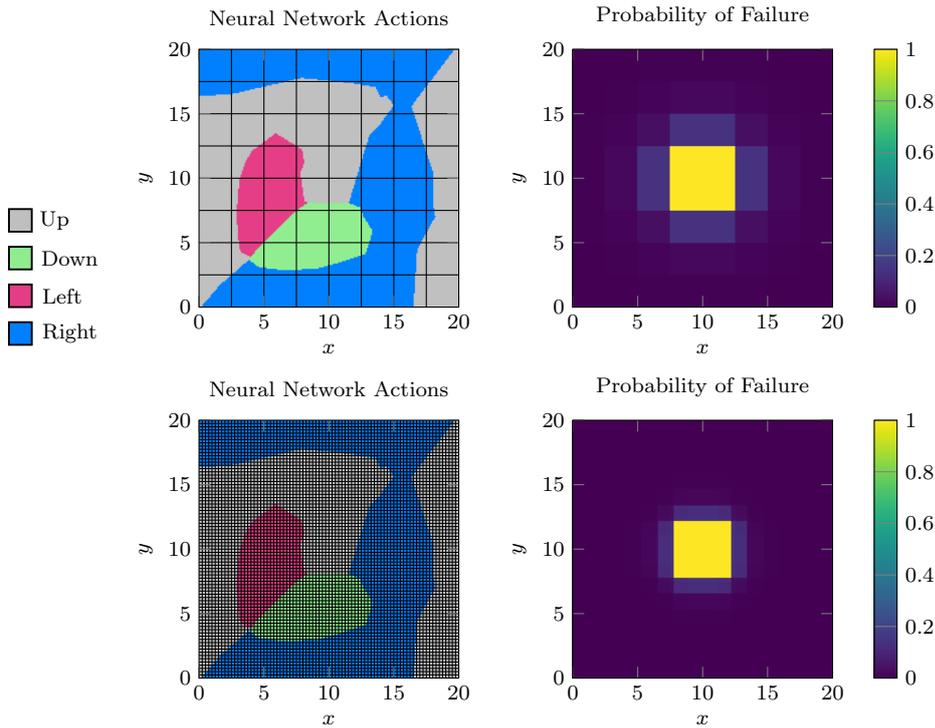


    \caption{Overapproximated probability of falling into the pit for different cell partitions. The plots in the left column show the cells plotted on top of the neural network policy, while the plots in the right column show the overapproximated probability of falling into the pit from each cell.  \label{fig:discre_comp}}
\end{figure}

With these modifications in place, we can apply dynamic programming using \cref{eq:bellmanProbVal} to determine the probability of reaching states in $\mathcal{B}$ for each cell $c \in \mathcal{C}$. \Cref{fig:discre_comp} shows the results of this process for two different cell partitions: a coarse partition (top row) and a fine partition (bottom row). In both cases, all cells that overlap with the pit have a probability of one assigned to them, and the probability of falling into the pit decreases as cells get further away from it. The coarse partitioning results in significantly more overapproximation error than the fine partitioning. However, partitioning the neural network input space uniformly into small cells significantly increases complexity, especially as the dimension of the input space increases. \Cref{sec:reduce_overappox} describes ways to reduce overapproximation error that only require a fine resolution in critical areas of the state space.

\section{Reducing Overapproximation Error}\label{sec:reduce_overappox}
The model checking formulation presented here results in two sources of overapproximation error. The first source of error, which we will refer to as policy overapproximation error, is the overapproximation of the neural network policy from the neural network verification tools. We assume that the actions in $\mathcal{A}_c$ may be taken at any point in the cell and that we always take the action with the worst-case probability (see the maximization in \cref{eq:bellmanProbQ}). Even if the worst-case action covers only a small portion of a cell, we must assume that the action is possible at any point in the cell. 

The second source of error, which we will refer to as worst-case transition error, is the overapproximation in the transition model shown in \cref{eq:cell_transition}. We assume that all points in the cell transition to the worst-case next cell for a given outcome. In order to produce meaningful probabilistic guarantees, it is crucial that we develop methods to reduce the overapproximation error. In this work, we present both offline and online error reduction methods. 

\subsection{Offline Reduction: Adaptive Verification}\label{sec:offline_reduc}
\begin{figure}[htb]
    \begin{tikzpicture}
    \filldraw[fill = left, draw = black] (0,-2.5) rectangle (2, -0.5);
    \filldraw[fill = left, draw = black] (3,-1.2) rectangle (5, 0.8);
    \filldraw[fill = left, draw = black] (6,-1.2) rectangle (8, 0.8);
    
    \fill[up] (2.0,-2.0) arc (270:180:0.5 and 1.5);
    \fill[up] (2.0,-2.0) -- (1.5,-0.5) -- (2.0,-0.5) -- cycle;
    
    \fill[up] (5.0,-0.7) arc (270:180:0.5 and 1.5);
    \fill[up] (5.0,-0.7) -- (4.5,0.8) -- (5.0,0.8) -- cycle;
    
    \fill[up] (8.0,-0.7) arc (270:180:0.5 and 1.5);
    \fill[up] (8.0,-0.7) -- (7.5,0.8) -- (8.0,0.8) -- cycle;
    
    \draw (0,-2.5) rectangle (2, -0.5);
    \draw (3,-1.2) rectangle (5, 0.8);
    \draw (6,-1.2) rectangle (8, 0.8);
    
    
    \draw (3, -1.2) rectangle (4, -0.53);
    \draw (3, -0.53) rectangle (4, 0.13);
    \draw (3, 0.13) rectangle (4, 0.8);
    \draw (4, -1.2) rectangle (5, -0.53);
    \draw (4, -0.53) rectangle (5, 0.13);
    \draw (4, 0.13) rectangle (5, 0.8);
    
    \draw (7.0, -0.2) rectangle (7.5, 0.8);
    \draw (7.5, -0.2) rectangle (8, 0.8);
    \draw (7.0, -1.2) rectangle (8.0, -0.7);
    
    \draw (7.0, -0.7) rectangle (7.5, -0.2);
    \draw (7.5, -0.7) rectangle (8, -0.2);
    
    
    \filldraw[fill = blue!80, draw = black] (3,-3.8) rectangle (5, -1.8);
    \filldraw[fill = blue!80, draw = black] (6,-3.8) rectangle (8, -1.8);
    
    \filldraw[fill = blue!80, draw = black] (3, -3.8) rectangle (4, -3.13);
    \filldraw[fill = blue!80, draw = black] (3, -3.13) rectangle (4, -2.47);
    \filldraw[fill = blue!80, draw = black] (3, -2.47) rectangle (4, -1.8);
    \filldraw[fill = red, draw = black] (4, -3.8) rectangle (5, -3.13);
    \filldraw[fill = red, draw = black] (4, -3.13) rectangle (5, -2.47);
    \filldraw[fill = red, draw = black] (4, -2.47) rectangle (5, -1.8);
    
    
    
    \filldraw[fill = blue!80, draw = black] (7.0, -2.8) rectangle (7.5, -1.8);
    \filldraw[fill = red, draw = black] (7.5, -2.8) rectangle (8, -1.8);
    \filldraw[fill = blue!80, draw = black] (7.0, -3.8) rectangle (8.0, -3.3);
    
    \filldraw[fill = blue!80, draw = black] (7.0, -3.3) rectangle (7.5, -2.8);
    \filldraw[fill = red, draw = black] (7.5, -3.3) rectangle (8, -2.8);
    
    \node [below] at (1, 0) {Policy};
    
    \node [below] at (4, 1.3) {Uniform};
    \node [below] at (7, 1.3) {Efficient};

    
    
    
    
    
    
    \end{tikzpicture}
    \caption{Two possible partitions of the input space for the simple policy with two possible actions shown on the left. The pink region corresponds to the first action, and the gray region corresponds to the second action. The top row shows an overlay of the partition on the policy, while bottom row shows the corresponding number of actions in each cell. Blue cells have one possible action, while red cells have two possible actions. \label{fig:adap_demo}}
\end{figure}
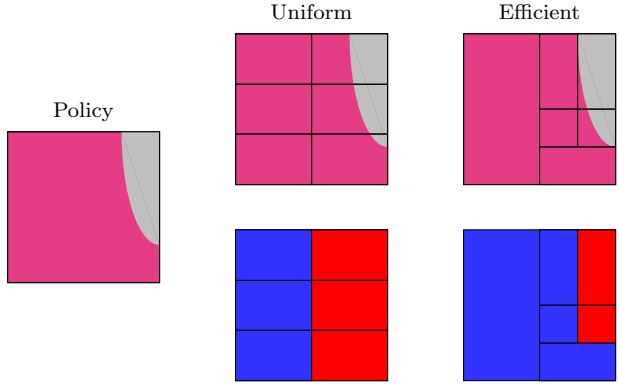

As described in \cref{sec:policy_overapprox}, overapproximation error grows with the number of possible actions in a cell, and cells with only one possible action will have no overapproximation error in the policy. Therefore, in order to partition our space in a way that minimizes policy overapproximation, we seek to minimize the volume of the input space occupied by cells that have multiple possible actions. \Cref{fig:adap_demo} shows an example of two possible partitions of the input space for an example policy. While both partitions contain the same number of cells, the second partition has a smaller area of the input space covered by cells with multiple possible actions and therefore a smaller overapproximation error. Our goal is to develop a verification strategy that will automatically generate a partition similar to the rightmost partition in \cref{fig:adap_demo}.

We use an adaptive verification strategy summarized in \cref{alg:adaptive_verification} to obtain $\tilde{\pi}$.
We begin with a single cell that encompasses all of $\mathcal{S}$. Each time we evaluate a cell, we run the verification tool to check which actions are possible in the specified cell to obtain $\mathcal{A}_c$. If $\mathcal{A}_c$ contains more than one action and the cell exceeds the minimum cell size, we split the cell into smaller cells. Splitting continues until all cells are either below the minimum cell width in the splitting dimension or have a single action in $\mathcal{A}_c$.
\begin{algorithm}
\caption{Adaptive Verification \label{alg:adaptive_verification}}
\begin{algorithmic}[1]
\Function{AdaptiveVerification}{neuralNetwork, inputCell, minCellSize}
    \State initialize $s$ to empty stack
    \State push inputCell onto $s$
    \While{$s$ is not empty}
       \State $c \gets \text{pop}(s)$
       \State $\mathcal{A}_c \gets$ \Call{VerificationTool}{neuralNetwork, $c$}
       \If{length of $\mathcal{A}_c > 1$ and size of $c >$ minCellSize}
            \State split $c$ according to splitting strategy
            \State push the resulting cells to $s$
        \EndIf
    \EndWhile
    \State \textbf{return} $\tilde{\pi}$
\EndFunction
\end{algorithmic}
\end{algorithm}

\begin{figure}[htb]
    \begin{tikzpicture}
    \draw (0,-1) rectangle (2, 1);
    \draw (3,-1) rectangle (5, 1);
    \draw (6,-1) rectangle (8, 1);
    
    \fill[down] (0,-1) circle (0.1);
    \fill[right] (0,1) circle (0.1);
    \fill[down] (2,-1) circle (0.1);
    \fill[right] (2,1) circle (0.1);
    
    \draw[dashed] (-0.2, 0) -- (2.2, 0);
    
    \fill[down] (3,-1) circle (0.1);
    \fill[down] (3,1) circle (0.1);
    \fill[right] (5,-1) circle (0.1);
    \fill[right] (5,1) circle (0.1);
    
    \draw[dashed] (4, -1.2) -- (4, 1.2);
    
    \fill[down] (6,-1) circle (0.1);
    \fill[right] (6,1) circle (0.1);
    \fill[right] (8,-1) circle (0.1);
    \fill[right] (8,1) circle (0.1);
    
    \draw[dashed] (5.8, 0) -- (8.2, 0);
    \draw[dashed] (7, -1.2) -- (7, 1.2);
    
    \end{tikzpicture}
    \caption{Example splitting of three cells for the informed splitting strategy based on the actions at the corners. The colored dots represent the actions at the corners with different colors indicating different actions.  \label{fig:corner_method}}
\end{figure}
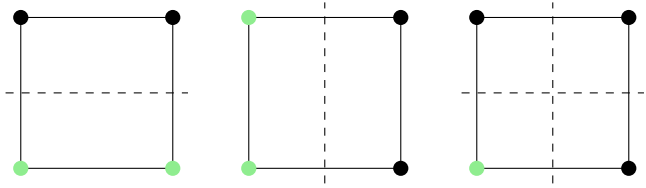

Because calls to neural verification tools are computationally expensive, we want to select a splitting strategy that will allow us to minimize the number of calls. We tested the following two splitting strategies. 
\begin{itemize}
    \item \emph{All split}: splits the cell along all dimensions at the midpoint
    \item \emph{Informed split}: attempts to speed up the verification process by first evaluating the neural network at the corners of the cell. If the corner points evaluate to different actions, we know that the verification tool would return multiple possible actions. Therefore, we can split the cell without calling it. Furthermore, we can use the evaluations of the corner points to select the dimensions to split.
\end{itemize}
\Cref{fig:corner_method} demonstrates the informed split strategy for different corner evaluations. If the adjacent actions are the same across a particular dimension, we do not split along that dimension. For example, the adjacent actions in the first dimension in the leftmost cell of \cref{fig:corner_method} are equal, so we do not split in the first dimension.

\begin{figure}[htb]
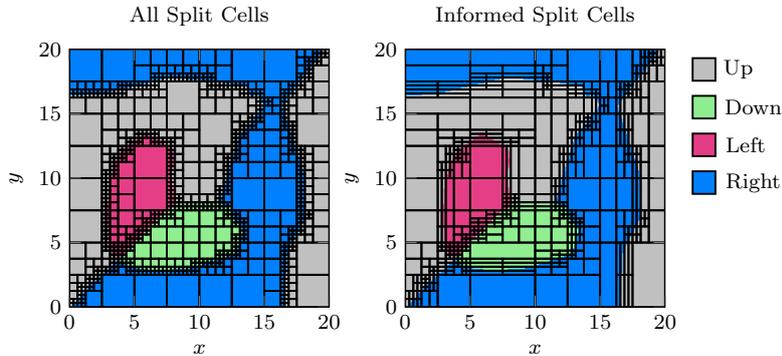


    \caption{Resulting cell partition when applying \cref{alg:adaptive_verification} to the continuum world explanatory example plotted on top of the neural network policy for each splitting strategy. \label{fig:adap_cw}}
\end{figure}

The adaptive verification algorithm encourages small cells near the decision boundaries of the neural network policy, and larger cells in continuous regions of the same action. This result is illustrated by \cref{fig:adap_cw}, which shows the result of applying \cref{alg:adaptive_verification} to the continuum world example with each splitting strategy. Both splitting strategies focus on the decision boundaries of the network. By only splitting along particular dimensions, the informed split strategy results in fewer cells overall.

\subsection{Online Reduction: State Abstraction}\label{sec:online_reduc}
While offline error reduction addresses policy overapproximation error, it does not address the second source of error. In order to reduce both types of error, we add online error reduction techniques to the model checking process. The online overapproximation error reduction techniques presented in this work are inspired by state abstraction techniques developed to solve for the value function of MDPs with large state spaces \citep{munos2002variable}. State abstraction relies on the assumption that large portions of the state space will have low variability in the policy or value function, while other more critical regions of the state space will require a finer resolution for accuracy. Some portions of the state space are safety-critical; however, a large portion of the state space will have a low probability estimate regardless of the action taken. During the solving process, we use heuristics based on our current probability estimate to determine critical portions of the state space. Splitting cells in these critical regions allows us to significantly reduce overapproximation error during the solving process without making cells in the input space unnecessarily small.

We address the worst-case transition error with an online splitting heuristic based on the maximum range of probability values for next cells. This range is computed as
\begin{equation}
\label{eq:transition_range}
    \text{transitionRange} = \max_i \left[ \max_{c' \in \mathcal{C}'_i} \Pr^{\tilde{\pi}}(c') - \min_{c' \in \mathcal{C}'_i} \Pr^{\tilde{\pi}}(c') \right]
\end{equation}
As an example, the worst-case transition range for the example shown in \cref{fig:transition_demo} is calculated as
\begin{equation}
\label{eq:transition_range_cw}
    \text{transitionRange} = \max (0.0 - 0.0, 0.13 - 0.03, 0.13 - 0.12, 0.02 - 0.0) = 0.1
\end{equation}
If this range is high, the worst-case transition overapproximation error is likely to be high. Therefore, we split the cell if this range exceeds a specified threshold and the cell is larger than a minimum cell size. The threshold can be tuned by the user to achieve a desired resolution. Splitting the cell will shrink the region of next states for each new cell and reduce the spread of overapproximation error.

The second online splitting heuristic aims to reduce policy overapproximation error and applies to cells that have multiple actions in $\mathcal{A}_c$. If a cell has multiple actions, we compute the range of the probabilities when taking each action as
\begin{equation}
    \label{eq:action_range}
    \text{actionRange} = \max_{a \in \mathcal{A}_c} \Pr^{\tilde{\pi}}(c, a) - \min_{a \in \mathcal{A}_c} \Pr^{\tilde{\pi}}(c, a)
\end{equation}
If this range exceeds a threshold and the cell is larger than the minimum cell size, we split the cell and rerun the verification on the resulting smaller cells. 

\subsection{Algorithm Summary}
The neural network model checking methods established in this work are summarized in \cref{alg:nnmc}.
Inputs to the algorithm include the neural networks to verify, the reachable set encoding the property we wish to verify, an initial cell that covers the entire network input space, the minimum cell size, and the transition and action thresholds. The last three input parameters can be tuned to achieve desired accuracy. A smaller minimum cell size and lower values for the thresholds will result in smaller overapproximation error and therefore a more accurate estimate of the probabilities.
\begin{algorithm}
\caption{Neural Network Model Checking \label{alg:nnmc}}
\begin{algorithmic}[1]
\Function{Check}{neuralNetwork, $\mathcal{B}$, inputCell, minCellSize, transitionThreshold, actionThreshold}
    \State $\tilde{\pi} \gets$ \Call{AdaptiveVerification}{neuralNetwork, inputCell, minCellSize}
    \State $\Pr^{\tilde{\pi}}(c) \gets 0$ for all $c \notin \mathcal{B}$
    \State $\Pr^{\tilde{\pi}}(c) \gets 1$ for all $c \in \mathcal{B}$
    \Repeat
       \State initialize $s$ to an empty stack
       \State push $c$ onto $s$ for all $c \in \mathcal{C}$
       \While{$s$ is not empty}
       \State $c \gets \text{pop}(s)$
       \State ranges $\gets \emptyset$
       \For $a \in \mathcal{A}_c$
           \State compute $\mathcal{C}'_{1:n}$ from $c$ for action $a$
           \State add $\max_i \left[ \max_{c' \in \mathcal{C}'_i} \Pr^{\tilde{\pi}}(c') - \min_{c' \in \mathcal{C}'_i} \Pr^{\tilde{\pi}}(c') \right]$ to ranges
       \EndFor
       transitionRange $\gets$ maximum of ranges
       \If{transitionRange $>$ transitionThreshold and size of $c >$ minCellSize}
            \State split $c$
            \State add resulting cells to $s$
        \Else
            \State $\Pr^{\tilde{\pi}}(c, a) \gets \sum_{c' \in \mathcal{C}} T(c' \mid c, a) \max_{a' \in \mathcal{A}_c} \Pr^{\tilde{\pi}}(c', a')$ for all $a \in \mathcal{A}_c$
            \State actionRange $\gets \max_{a \in \mathcal{A}_c} \Pr^{\tilde{\pi}}(c, a) - \min_{a \in \mathcal{A}_c} \Pr^{\tilde{\pi}}(c, a)$
            \If{actionRange $>$ actionThreshold and size of $c >$ minCellSize}
                \State split $c$
                \State add resulting cells to $s$
            \EndIf
        \EndIf
    \EndWhile
    \Until{convergence}
    \State \textbf{return} $\Pr^{\tilde{\pi}}$
\EndFunction
\end{algorithmic}
\end{algorithm}

The first step in the algorithm is to obtain an overapproximation of the neural network policy using the adaptive verification method in \cref{alg:adaptive_verification}. Next, the probability is initialized to zero for all cells except those that overlap with the reachable set $\mathcal{B}$, which are initialized to a probability of one. After these preprocessing steps, we begin value iteration with our online splitting heuristics. 
For each cell, we first compute the transition range according to \cref{eq:transition_range}. If the cell satisfies the worst-case transition splitting criterion, we split the cell. Otherwise, we perform the Bellman update (\cref{eq:bellmanProbQ}) to compute the probability of collision for taking each action in $\mathcal{A}_c$ from the cell. Using these probabilities and \cref{eq:action_range}, we can compute the action range and decide once again whether to split the cell.

\begin{figure}[htb]
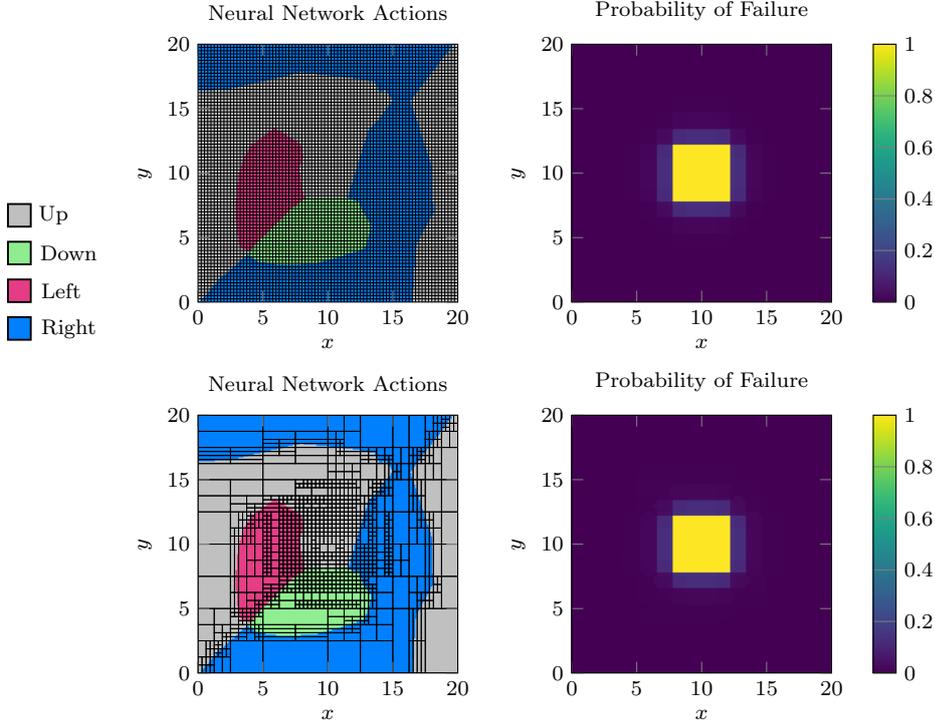


    \caption{Overapproximated probability of falling into the pit using a na\"ive approach to partitioning the state space into cells (top) compared to using adaptive verification with the informed splitting strategy along with the online heuristics. The plots in the left column show the cells plotted on top of the neural network policy, while the plots in the right column show the overapproximated probability of falling into the pit from each cell.  \label{fig:naive_v_smart_comp}}
\end{figure}

\Cref{fig:naive_v_smart_comp} compares the overapproximated probability of falling into the pit when taking two different approaches to partitioning the state space into cells. The first approach represents the na\"ive approach in which the space is uniformly partitioned into small cells. The second approach follows our algorithm, applying adaptive verification with the informed split strategy and using the online overapproximation error reduction methods during the solving process. The partition using the second approach results in small cells in regions of the state space near the pit and the decision boundaries of the network. Using our overapproximation error reduction methods, we are able to obtain similar probability estimates with significantly fewer cells in the partition.

\subsection{Complexity}
The overapproximation error reduction methods presented in \cref{sec:offline_reduc} and \cref{sec:online_reduc} are key contributors to the overall complexity of the algorithm. In the worst case, \cref{alg:adaptive_verification} will result in a uniform splitting of the input space in which all cells have the minimum cell size. Therefore, the worst-case complexity of \cref{alg:adaptive_verification} is exponential in the input dimension of the neural network controller. Despite this worst-case complexity, control policies in practice tend to be structured such that large, continuous regions of the state space correspond to the same action. Our adaptive verification approach takes advantage of this structure by allowing large regions of the same action to be grouped into a single cell while only decreasing cells on the decision boundaries of the network to the minimum cell size. 

The complexity of \cref{alg:adaptive_verification} observed in practice depends on the splitting strategy. Because the all split strategy simply splits along every dimension, the number of new cells created on each split is always exponential in the input dimension. In contrast, while still exponential in the worst case, the informed split strategy limits the number of new cells created in practice by selecting a subset of dimensions to split using the evaluations at the corners. We note that the number of corner points of a cell grows exponentially with the input dimension, so this strategy will be intractable for controllers with high-dimensional inputs. In this case, one could consider using the gradient-based splitting heuristics used in prior work on neural network verification to select a subset of dimensions to split \citep{reluval}. 

The online reduction strategies described in \cref{sec:online_reduc} also have worst-case exponential complexity in the input dimension. However, similarly to \cref{alg:adaptive_verification}, the strategies exploit the structure of the problem to limit this complexity in practice by only splitting cells in safety-critical regions. Another contributor to algorithm complexity is calls to the neural network verification tool. Depending on the verification algorithm, queries on larger cells tend to have greater complexity. Therefore, there is a tradeoff between using large cells to limit the total number of cells in the partition and keeping the cells small enough to be effectively handed by a neural network verification tool. Furthermore, the size and type of neural networks we can verify with our algorithm is limited by the current capabilities of neural network verification tools. These tools tend to perform best on small neural networks with simple architectures and rectified linear unit (ReLU) activations \citep{liu2019algorithms}. 

As a result of the worst-case complexity in the input dimension, our algorithm is most effective on neural network controllers with relatively low-dimensional inputs. Controllers that take the physical state of the system as input tend to have this property; however, our method will not scale well to controllers with high-dimensional inputs such as images.

\section{Application: Collision Avoidance Neural Networks} \label{sec:CAS}
We use the aircraft collision avoidance problem as an example application for our methods. Aircraft collision avoidance provides a compelling, real-world example of a safety-critical application in which neural network controllers provide a substantial benefit and has been used as a benchmark in previous work on neural network verification. Specifically, neural networks have been demonstrated as space-efficient controllers for a family of aircraft collision avoidance systems called the Airborne Collision Avoidance System X (ACAS X) \citep{Julian2016dasc, Julian2019jgcd}. ACAS X relies on a large numeric lookup table to provide optimized advisories during flight \citep{kochenderfer2011robust, kochenderfer2012next, olson2015airborne, owen2019acas}. \Citet{Julian2019jgcd} showed that it is possible to decrease the memory footprint of the table by training a neural network to take its place. The neural network representation decreases the required storage by a factor of 1,000 while maintaining comparable performance to the table in simulation. 

The collision avoidance neural networks used in this work are based on the networks in the VerticalCAS repository developed by \citet{Julian2019reach}. The repository contains an open source collision avoidance logic loosely modeled after the vertical logic used in ACAS X \citep{Julian2019reach}. The logic is designed to prevent near mid-air collisions (NMACs), which are defined as a simultaneous loss of separation to less than 500 ft horizontally and 100 ft vertically. To apply model checking to the problem, we must first formulate it as a Markov decision process. We use a similar formulation to what was used to generate the lookup table for ACAS X.

\paragraph{State Space} The state space for VerticalCAS consists of five state variables. \Cref{tab:vert_ss} summarizes the variables and their ranges, and \cref{fig:vert_state} provides a visual representation of the state variables.
The first three variables summarize the relative positioning and vertical rate of the ownship and intruder aircraft. The next state variable, $\tau$, compactly summarizes the horizontal geometry by representing the time until the  horizontal separation between the two aircraft is less than 500 ft. Finally, adding the previous advisory to the state space allows us to penalize a reversal or strengthening of an advisory while still satisfying the Markov property.
\begin{table}[htb]
    \caption{VerticalCAS state variables \label{tab:vert_ss}}
    \begin{tabular}{@{}llcc@{}}
         \toprule
         \textbf{Variable} & \textbf{Description} & \textbf{Units} & \textbf{Range} (low:high) \\
         \midrule
         $h$ & relative altitude of intruder & ft & $-8000:8000$\\
         $\dot{h}_0$ & ownship vertical rate & ft/s & $-100:100$ \\
         $\dot{h}_1$ & intruder vertical rate & ft/s & $-100:100$ \\
         $\tau$ & time to loss of lateral separation & s & 0:40 \\
         $a_{\text{prev}}$ & previous advisory & N/A & N/A \\
         \bottomrule
    \end{tabular}
\end{table}
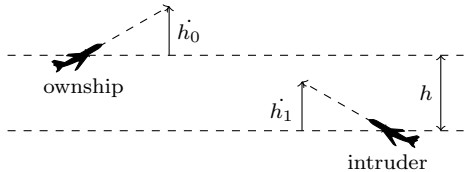
\begin{figure}[htb]
    \begin{tikzpicture}
    \node [aircraft side,fill=black,draw=black,minimum width=1cm,rotate=30,scale=0.7] at (0,0) {};
    \node [aircraft side,fill=black,draw=black,minimum width=1cm,rotate=-30,scale=0.7,xscale=-1] at (4,-1) {}; 
    \draw [dashed] (-1,0) -- (5,0); 
    \draw [dashed] (-1,-1) -- (5,-1); 
    \draw [<->] (4.7,-1) -- (4.7,0);
    \draw [dashed, <->] (0,0) -- (1.125833, 0.65);
    \draw [dashed, <->] (4,-1) -- (2.874167, -0.35);
    \draw [->] (2.874167, -1) -- (2.874167, -0.35);
    \draw [->] (1.125833, 0) -- (1.125833, 0.65);
    \node [left] at (4.7,-0.5) {$h$};
    \node [right] at (1.125833,0.325) {$\dot{h_0}$};
    \node [left] at (2.874167,-0.725) {$\dot{h_1}$};
    \node [below] at (0,-0.2) {ownship};
    \node [below] at (4,-1.2) {intruder};
    \end{tikzpicture}
    \caption{Visual representation of VerticalCAS state variables. \label{fig:vert_state}}
\end{figure}

\paragraph{Action Space} The action space consists of the advisories that the collision avoidance system will provide to the aircraft during flight. The logic has nine possible advisories, which are summarized in \cref{tab:vert_as}. All advisories except $\textsc{coc}$ represent an alert and command the aircraft to a particular vertical rate range. The $\textsc{coc}$ advisory indicates that there is currently no threat of collision with an intruding aircraft.
\begin{table}[htb]
    \caption{VerticalCAS action space \label{tab:vert_as}}
    \begin{tabular}{@{}ll@{}}
         \toprule
         \textbf{Action} & \textbf{Description} \\ 
         \midrule
         $\textsc{coc}$ & clear of conflict \\ 
         $\textsc{dnc}$ & do not climb \\ 
         $\textsc{dnd}$ & do not descend \\ 
         $\textsc{des1500}$ & descend $\geq$ 1500 ft/min \\ 
         $\textsc{cl1500}$ & climb $\geq$ 1500 ft/min \\ 
         $\textsc{sdes1500}$ & strengthen descent to $\geq$ 1500 ft/min \\ 
         $\textsc{scl1500}$ & strengthen climb to $\geq$ 1500 ft/min \\ 
         $\textsc{sdes2500}$ & strengthen descent to $\geq$ 2500 ft/min \\ 
         $\textsc{scl2500}$ & strengthen climb to $\geq$ 2500 ft/min \\ 
         \bottomrule
    \end{tabular}
\end{table}

\paragraph{Transition Model} The transition model uses the following linear dynamics model
\begin{equation}
\label{eq:dynamics}
\begin{split}
    & h \gets h + \dot{h}_1 + 0.5\ddot{h}_1 - \dot{h}_0 - 0.5\ddot{h}_0 \\
    & \dot{h}_0 \gets \dot{h}_0 + \ddot{h}_0 \\
    & \dot{h}_1 \gets \dot{h}_1 + \ddot{h}_1 \\
    & \tau \gets \tau - 1 \\  
    & a_{\text{prev}} \gets a \\
\end{split}
\end{equation}
We assume a one second time step corresponding to the 1 Hz update frequency of the collision avoidance system. The dynamics model is made stochastic by assuming distributions over the accelerations of the ownship and intruder. The intruder is assumed to have an equal chance of following accelerations $-g/8$, $g/8$, and 0 ft/s$^2$ as in \citet{Julian2019dasc}. The ownship follows accelerations $\ddot{h}_{1-3}$ based on its previous advisory with associated probabilities $p_{1:3}$ shown in \cref{tab:own_accel}. This transition model contains two features that add robustness. First, the ownship is assumed to follow the accelerations associated with its previous advisory (rather than its current advisory), which represents a short delay in the aircraft's response to the advisory. Additionally, the ownship is assumed to accelerate in the opposite direction of its advisory $20 \%$ of the time to further incorporate errors in aircraft response. For example, if a human pilot is responsible for executing the collision avoidance maneuver, they may not respond instantaneously to an advisory and could be accelerating in a direction opposite the advisory. The probabilistic safety guarantees presented in this paper are based on this stochastic dynamics model.
\begin{table}[htb]
    \caption{Transition accelerations \label{tab:own_accel}}
    \begin{tabular}{@{}lcc@{}}
         \toprule
         \textbf{Previous Action} & \textbf{Probabilities ($p_{1-3})$} & \textbf{Accelerations ($\ddot{h}_{1-3}$)} \\
         \midrule
         $\textsc{coc}$ & $[0.34, 0.33, 0.33] $ & $[0.0, -g/3, g/3]$ \\
         $\textsc{dnc}$ & $[0.50, 0.30, 0.20]$ & $[-g/3, -g/2, g/3]$ \\
         $\textsc{dnd}$ & $[0.50, 0.30, 0.20]$ & $[g/3, g/2, -g/3]$ \\
         $\textsc{des1500}$ & $[0.50, 0.30, 0.20]$ & $[-g/3, -g/2, g/3]$ \\
         $\textsc{cl1500}$ & $[0.50, 0.30, 0.20]$ & $[g/3, g/2, -g/3]$ \\
         $\textsc{sdes1500}$ & $[0.50, 0.30, 0.20]$ & $[-g/2.5, -g/2, g/3]$ \\
         $\textsc{scl1500}$ & $[0.50, 0.30, 0.20]$ & $[g/2.5, g/2, -g/3]$ \\
         $\textsc{sdes2500}$ & $[0.50, 0.30, 0.20]$ & $[-g/2.5, -g/2, g/3]$ \\
         $\textsc{scl2500}$ & $[0.50, 0.30, 0.20]$ & $[g/2.5, g/2, -g/3]$ \\
         \bottomrule
    \end{tabular}
\end{table}

The reward model balances between safety and efficiency with a high penalty for an NMAC and relatively smaller penalties for alerting advisories. Using this formulation, we can solve for the optimal policy using value iteration. Traditionally, the final action-values result in a large numeric lookup table. To reduce the on-board memory requirements, \Citet{Julian2019jgcd} train a neural network representation to approximate the action-value function. One network is trained for each discrete previous advisory. The values of the other four state variables in \cref{tab:vert_ss} make up the four-dimensional input to the network, and the approximate value of each action makes up the nine-dimensional output. Each network has five hidden layers with 25 units each that use rectified linear unit (ReLU) activation functions. The networks used in this work can be found at \url{https://github.com/sisl/AdaptiveVerification/}.

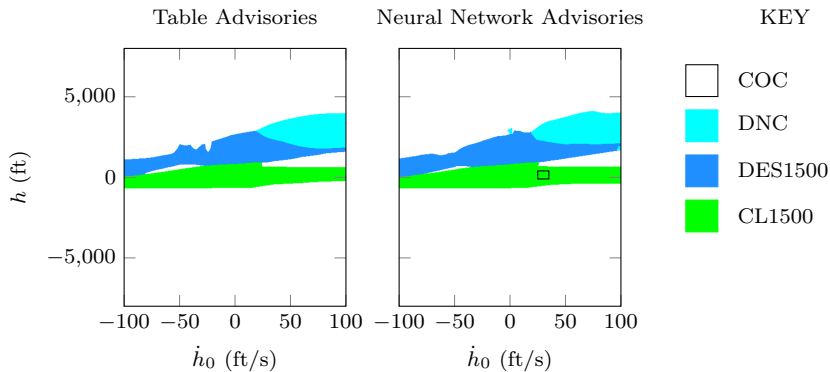
\begin{figure}[htb]
    \begin{tikzpicture}
    \begin{groupplot}[group style={horizontal sep = 0.7cm, vertical sep = 2.5cm, group size=3 by 1}]
    \nextgroupplot [height = {5cm}, ylabel = {$h$ (ft)}, title = {Table Advisories}, xmin = {-100.0}, xmax = {100.0}, ymax = {8000.0}, xlabel = {$\dot{h}_0$ (ft/s)}, ymin = {-8000.0}, width = {4.5cm}, enlargelimits = false, axis on top]\addplot [point meta min=0, point meta max=8] graphics [xmin=-100.0, xmax=100.0, ymin=-8000.0, ymax=8000.0] {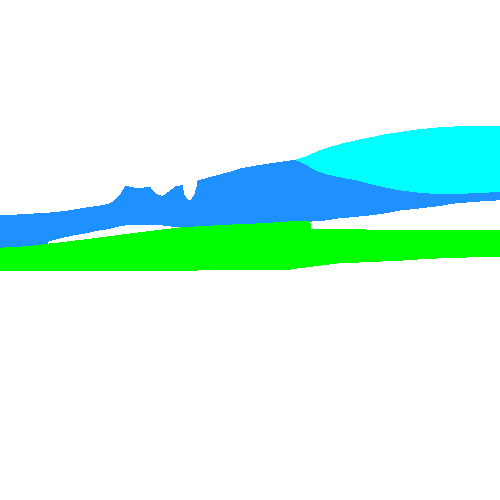};
    \nextgroupplot [height = {5cm}, title = {Neural Network Advisories}, xmin = {-100}, xmax = {100}, ymax = {8000}, xlabel = {$\dot{h}_0$ (ft/s)}, ymin = {-8000}, width = {4.5cm}, enlargelimits = false, axis on top, yticklabels={,,}]\addplot [point meta min=0, point meta max=8] graphics [xmin=-100, xmax=100, ymin=-8000, ymax=8000] {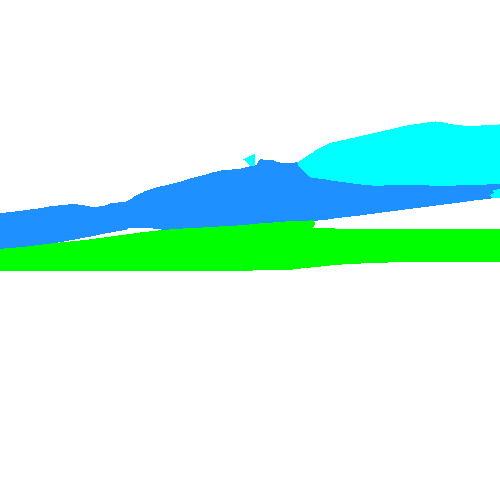};
    \draw (axis cs:25.0,-100.0) rectangle (axis cs:35,400.0);
    \nextgroupplot [height = {5cm}, title = {KEY}, hide axis = {true}, width = {4.5cm}, enlargelimits = false, axis on top]\addplot [point meta min=-2.0, point meta max=2.0] graphics [xmin=-2, xmax=2, ymin=-2, ymax=2] {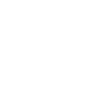};
    \addplot+[scatter, scatter src=explicit symbolic, only marks = {true}, scatter/classes = {{ra_1={mark=square, style={black, mark options={fill=ra_1}, mark size=6}},ra_2={style={ra_2, mark size=6}},ra_3={style={ra_3, mark size=6}},ra_4={style={ra_4, mark size=6}},ra_5={style={ra_5, mark size=6}},ra_6={style={ra_6, mark size=6}},ra_7={style={ra_7, mark size=6}},ra_8={style={ra_8, mark size=6}},ra_9={style={ra_9, mark size=6}}}}] coordinates {
    (-1.5, 1.5) [ra_1]
    (-1.5, 0.8) [ra_2]
    (-1.5, 0.1) [ra_4]
    (-1.5, -0.6) [ra_5]
    };
    \node at (axis description cs:0.25, 0.88) [black,anchor=west] {COC };
    \node at (axis description cs:0.25, 0.71) [black,anchor=west] {DNC };
    \node at (axis description cs:0.25, 0.53) [black,anchor=west] {DES1500};
    \node at (axis description cs:0.25, 0.35) [black,anchor=west] {CL1500 };
    \end{groupplot}
    
    \end{tikzpicture}
    \caption{Comparison of table policy to neural network policy for a slice of the state space. The intruder vertical rate is fixed at $-90$ ft/s, $\tau$ is fixed at 5 seconds, and the previous advisory is $\textsc{coc}$. \label{fig:table_nn_compare}}
\end{figure}

\Cref{fig:table_nn_compare} shows a comparison of the neural network policy and lookup table policy for a slice of the state space.
The neural network policy closely approximates the table policy with a few subtle differences that are visible in the plot. Even though the alerting regions in the neural network policy appear continuous, the plot is generated by evaluating a finite number of points in the state space and does not guarantee this property. For example, while the region highlighted on the neural network policy in \cref{fig:table_nn_compare} appears to evaluate to $\textsc{cl1500}$ at all points within it, we cannot guarantee this property just by examining the plotted points. To provide a guarantee that all advisories in that region are in fact $\textsc{cl1500}$, we use a neural verification tool to obtain $\mathcal{A}_c$. For this application, we use the Reluval algorithm to verify each cell due to its fast performance on the collision avoidance networks \citep{liu2019algorithms, reluval}.

We can craft an LTL formula to determine the probability of an NMAC by using the temporal operator ``eventually,'' written as $\mathsf{F}$. Let the atomic proposition $N$ represent whether or not a cell belongs to the NMAC region ($\tau = 0$ seconds and $h <$ 100 ft). The LTL formula of interest is $\mathsf{F} N$, read as ``eventually N.'' The probability of satisfying this formula corresponds to the probability that a state will eventually reach an NMAC state. This problem is easily converted to a reachability problem. Let $\mathcal{B}$ represent the collection of all cells that satisfy $N$. We seek to find the maximum probability of reaching cells in $\mathcal{B}$ while following the overapproximated neural network policy.

\section{Results}\label{sec:res}
For ease of computation and visualization, we tested our methods on a two-dimensional version of the collision avoidance problem in which the intruder vertical rate is fixed at $-90$ ft/s before testing on the full scale model. Since most of the overapproximation error in our initial experiments seemed to be concentrated at high vertical rates, we selected $-90$ ft/s for the intruder vertical rate to understand how our method performs in a challenging area of the state space. All other aspects of the problem, including the neural networks used for verification, remain the same. By taking this approach, we were able to better understand the effects of each aspect of the algorithm. Therefore, the results summarizing the effects of the overapproximation reduction techniques were generated using the two-dimensional model. After providing intuition with these results, we present the results of the full scale model.

\subsection{Adaptive Verification}
\begin{figure}[htb]
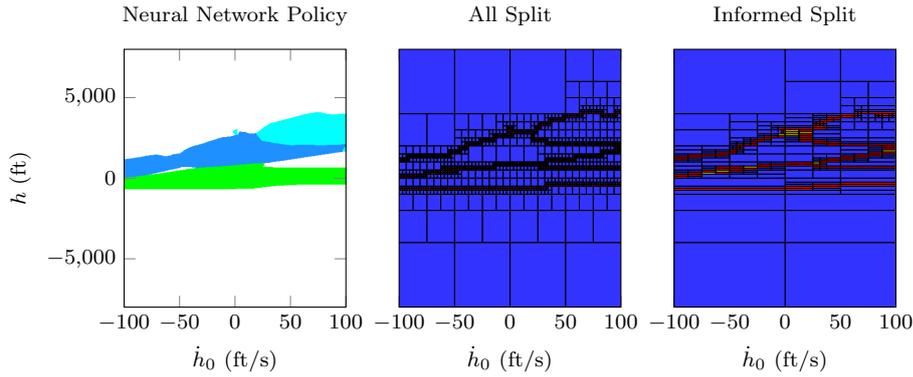


    \caption{Comparison of the resulting state space partition for the adaptive splitting strategies. The intruder vertical rate is fixed at $-90$ ft/s, $\tau$ is 5 seconds, and the previous advisory is $\textsc{coc}$. Cells that are colored blue have only one possible advisory in $\mathcal{A}_c$, red cells have two possible advisories, and yellow cells have three or more possible advisories. \label{fig:adap_comparison}}
\end{figure}

We tested both the all split and informed split adaptive verification splitting strategies to obtain the overapproximated policy $\tilde{\pi}$. \Cref{fig:adap_comparison} shows the results of both splitting strategies when the intruder vertical rate is fixed at $-90$ ft/s, and \cref{tab:adap_timing} shows a comparison of runtime for both the two-dimensional and full scale model for each strategy. We also include the runtime of a uniform, non-adaptive splitting strategy in which all cells are the minimum cell size used in the adaptive strategies. For the full-scale model, the non-adaptive runtime was estimated based on the time required to verify a single cell of the minimum cell size. Time trials were run on a single 4.20 GHz Intel Core i7 processor.
Both adaptive splitting strategies result in small cells around the decision boundaries of the network; however, the informed splitting strategy results in fewer cells.
\begin{table}[htb]
    \caption{Adaptive verification timing results \label{tab:adap_timing}}
    \begin{tabular}{@{}lcc@{}}
         \toprule
         \textbf{Splitting Strategy} & \textbf{$\mathbf{\dot{h}_1}$ Fixed Time (s)} & \textbf{Full Scale Time (s)} \\
         \midrule
         Uniform & 235.5 & 16,000 (est.) \\
         All Split & 21.9 & 2,500.3 \\
         Informed Split & 5.5 & 655.6 \\
         \bottomrule
    \end{tabular}
\end{table}

The adaptive strategies require fewer calls to the verification tool and therefore perform faster than a non-adaptive strategy. Additionally, whenever a cell is split and the resulting cells are reverified, the verification tool only needs to check for the actions that were possible in the larger cell. Informed split is faster than all split because it makes even fewer Reluval calls. Checking the corners to inform the split prevents unnecessary calls and results in fewer cells in the final partition. Due to its speed, the informed split strategy was selected for the rest of the analysis in this work.

\begin{figure}[htb]
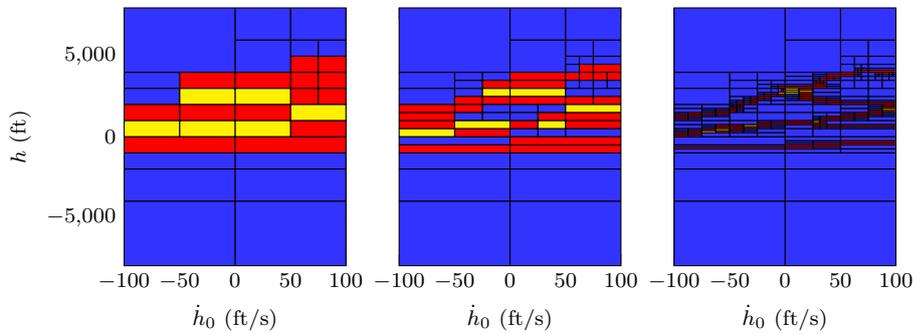


    \caption{Adaptive verification results using informed split for decreasing minimum cell sizes (left to right). The intruder vertical rate is fixed at $-90$ ft/s, $\tau$ is 5 seconds, and the previous advisory is $\textsc{coc}$. Cells that are colored blue have only one possible advisory in $\mathcal{A}_c$, red cells have two possible advisories, and yellow cells have three or more possible advisories. \label{fig:adap_resolution}}
\end{figure}

While the adaptive verification technique presented here addresses policy overapproximation error for model checking, it may also be used on its own to analyze neural network policies and detect decision boundaries. The algorithm is an anytime algorithm, and the resolution can be controlled using the minimum cell size parameter. \Cref{fig:adap_resolution} shows the results of running adaptive verification as the parameter is decreased. To minimize overapproximation error, we want to maximize the amount of the state space covered by cells with a single advisory in them (blue cells). \Cref{fig:adap_resolution} demonstrates that as the minimum cell size decreases, overapproximation error in the policy also decreases.

\begin{figure}[htb]
    \begin{tikzpicture}
    \begin{groupplot}[group style={horizontal sep = 0.7cm, vertical sep = 2.5cm, group size=2 by 1}]
    \nextgroupplot [height = {5cm}, ylabel = {$h$ (ft)}, title = {Number of Advisories}, xmin = {0.0}, xmax = {40.0}, ymax = {8000}, xlabel = {$\tau$ (s)}, ymin = {-8000}, width = {5.7cm}, enlargelimits = false, axis on top]\addplot [point meta min=0, point meta max=3] graphics [xmin=0.0, xmax=40.0, ymin=-8000, ymax=8000] {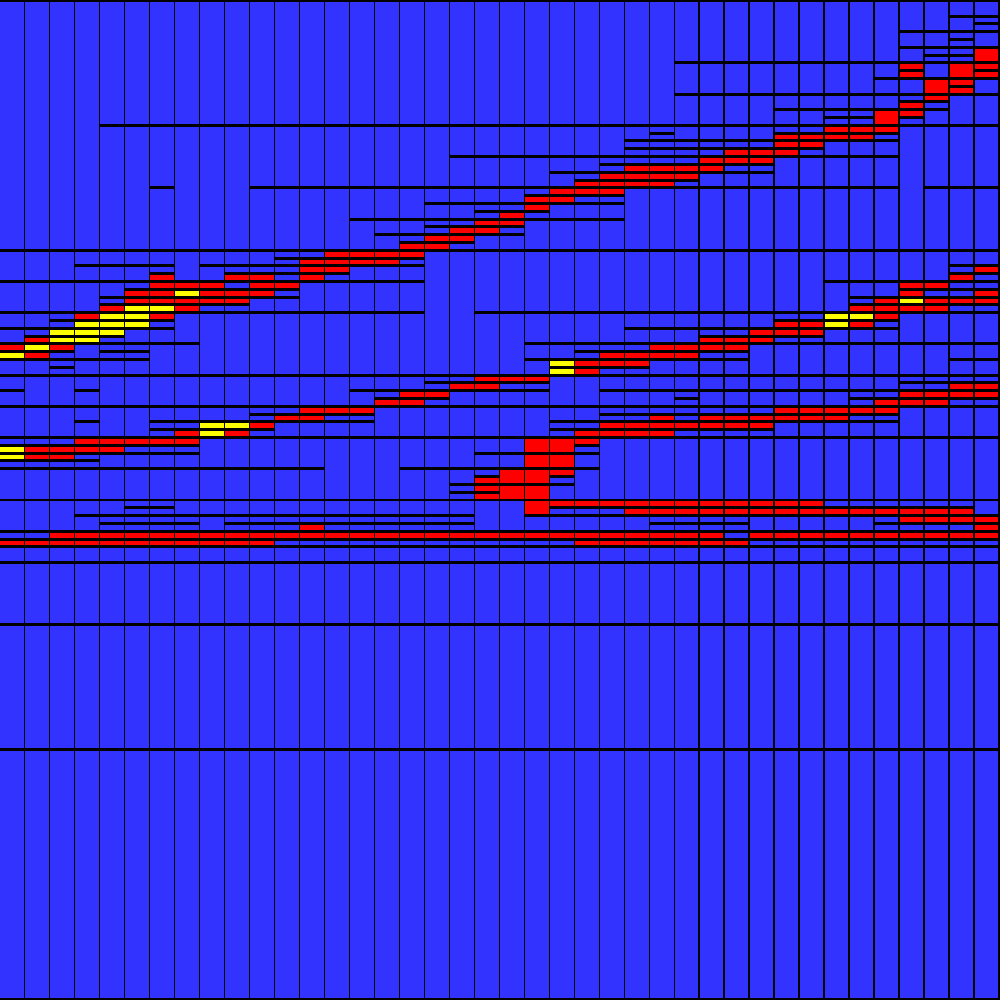};
    \nextgroupplot [height = {5cm}, title = {Probability of NMAC}, xmin = {0.0}, xmax = {40.0}, ymax = {8000}, xlabel = {$\tau$ (s)}, ymin = {-8000}, width = {5.7cm}, enlargelimits = false, axis on top, yticklabels={,,}, colorbar style={
                width=0.3cm}, colormap={mycolormap}{ rgb(0cm)=(0.267004,0.004874,0.329415) rgb(1cm)=(0.26851,0.009605,0.335427) rgb(2cm)=(0.269944,0.014625,0.341379) rgb(3cm)=(0.271305,0.019942,0.347269) rgb(4cm)=(0.272594,0.025563,0.353093) rgb(5cm)=(0.273809,0.031497,0.358853) rgb(6cm)=(0.274952,0.037752,0.364543) rgb(7cm)=(0.276022,0.044167,0.370164) rgb(8cm)=(0.277018,0.050344,0.375715) rgb(9cm)=(0.277941,0.056324,0.381191) rgb(10cm)=(0.278791,0.062145,0.386592) rgb(11cm)=(0.279566,0.067836,0.391917) rgb(12cm)=(0.280267,0.073417,0.397163) rgb(13cm)=(0.280894,0.078907,0.402329) rgb(14cm)=(0.281446,0.08432,0.407414) rgb(15cm)=(0.281924,0.089666,0.412415) rgb(16cm)=(0.282327,0.094955,0.417331) rgb(17cm)=(0.282656,0.100196,0.42216) rgb(18cm)=(0.28291,0.105393,0.426902) rgb(19cm)=(0.283091,0.110553,0.431554) rgb(20cm)=(0.283197,0.11568,0.436115) rgb(21cm)=(0.283229,0.120777,0.440584) rgb(22cm)=(0.283187,0.125848,0.44496) rgb(23cm)=(0.283072,0.130895,0.449241) rgb(24cm)=(0.282884,0.13592,0.453427) rgb(25cm)=(0.282623,0.140926,0.457517) rgb(26cm)=(0.28229,0.145912,0.46151) rgb(27cm)=(0.281887,0.150881,0.465405) rgb(28cm)=(0.281412,0.155834,0.469201) rgb(29cm)=(0.280868,0.160771,0.472899) rgb(30cm)=(0.280255,0.165693,0.476498) rgb(31cm)=(0.279574,0.170599,0.479997) rgb(32cm)=(0.278826,0.17549,0.483397) rgb(33cm)=(0.278012,0.180367,0.486697) rgb(34cm)=(0.277134,0.185228,0.489898) rgb(35cm)=(0.276194,0.190074,0.493001) rgb(36cm)=(0.275191,0.194905,0.496005) rgb(37cm)=(0.274128,0.199721,0.498911) rgb(38cm)=(0.273006,0.20452,0.501721) rgb(39cm)=(0.271828,0.209303,0.504434) rgb(40cm)=(0.270595,0.214069,0.507052) rgb(41cm)=(0.269308,0.218818,0.509577) rgb(42cm)=(0.267968,0.223549,0.512008) rgb(43cm)=(0.26658,0.228262,0.514349) rgb(44cm)=(0.265145,0.232956,0.516599) rgb(45cm)=(0.263663,0.237631,0.518762) rgb(46cm)=(0.262138,0.242286,0.520837) rgb(47cm)=(0.260571,0.246922,0.522828) rgb(48cm)=(0.258965,0.251537,0.524736) rgb(49cm)=(0.257322,0.25613,0.526563) rgb(50cm)=(0.255645,0.260703,0.528312) rgb(51cm)=(0.253935,0.265254,0.529983) rgb(52cm)=(0.252194,0.269783,0.531579) rgb(53cm)=(0.250425,0.27429,0.533103) rgb(54cm)=(0.248629,0.278775,0.534556) rgb(55cm)=(0.246811,0.283237,0.535941) rgb(56cm)=(0.244972,0.287675,0.53726) rgb(57cm)=(0.243113,0.292092,0.538516) rgb(58cm)=(0.241237,0.296485,0.539709) rgb(59cm)=(0.239346,0.300855,0.540844) rgb(60cm)=(0.237441,0.305202,0.541921) rgb(61cm)=(0.235526,0.309527,0.542944) rgb(62cm)=(0.233603,0.313828,0.543914) rgb(63cm)=(0.231674,0.318106,0.544834) rgb(64cm)=(0.229739,0.322361,0.545706) rgb(65cm)=(0.227802,0.326594,0.546532) rgb(66cm)=(0.225863,0.330805,0.547314) rgb(67cm)=(0.223925,0.334994,0.548053) rgb(68cm)=(0.221989,0.339161,0.548752) rgb(69cm)=(0.220057,0.343307,0.549413) rgb(70cm)=(0.21813,0.347432,0.550038) rgb(71cm)=(0.21621,0.351535,0.550627) rgb(72cm)=(0.214298,0.355619,0.551184) rgb(73cm)=(0.212395,0.359683,0.55171) rgb(74cm)=(0.210503,0.363727,0.552206) rgb(75cm)=(0.208623,0.367752,0.552675) rgb(76cm)=(0.206756,0.371758,0.553117) rgb(77cm)=(0.204903,0.375746,0.553533) rgb(78cm)=(0.203063,0.379716,0.553925) rgb(79cm)=(0.201239,0.38367,0.554294) rgb(80cm)=(0.19943,0.387607,0.554642) rgb(81cm)=(0.197636,0.391528,0.554969) rgb(82cm)=(0.19586,0.395433,0.555276) rgb(83cm)=(0.1941,0.399323,0.555565) rgb(84cm)=(0.192357,0.403199,0.555836) rgb(85cm)=(0.190631,0.407061,0.556089) rgb(86cm)=(0.188923,0.41091,0.556326) rgb(87cm)=(0.187231,0.414746,0.556547) rgb(88cm)=(0.185556,0.41857,0.556753) rgb(89cm)=(0.183898,0.422383,0.556944) rgb(90cm)=(0.182256,0.426184,0.55712) rgb(91cm)=(0.180629,0.429975,0.557282) rgb(92cm)=(0.179019,0.433756,0.55743) rgb(93cm)=(0.177423,0.437527,0.557565) rgb(94cm)=(0.175841,0.44129,0.557685) rgb(95cm)=(0.174274,0.445044,0.557792) rgb(96cm)=(0.172719,0.448791,0.557885) rgb(97cm)=(0.171176,0.45253,0.557965) rgb(98cm)=(0.169646,0.456262,0.55803) rgb(99cm)=(0.168126,0.459988,0.558082) rgb(100cm)=(0.166617,0.463708,0.558119) rgb(101cm)=(0.165117,0.467423,0.558141) rgb(102cm)=(0.163625,0.471133,0.558148) rgb(103cm)=(0.162142,0.474838,0.55814) rgb(104cm)=(0.160665,0.47854,0.558115) rgb(105cm)=(0.159194,0.482237,0.558073) rgb(106cm)=(0.157729,0.485932,0.558013) rgb(107cm)=(0.15627,0.489624,0.557936) rgb(108cm)=(0.154815,0.493313,0.55784) rgb(109cm)=(0.153364,0.497,0.557724) rgb(110cm)=(0.151918,0.500685,0.557587) rgb(111cm)=(0.150476,0.504369,0.55743) rgb(112cm)=(0.149039,0.508051,0.55725) rgb(113cm)=(0.147607,0.511733,0.557049) rgb(114cm)=(0.14618,0.515413,0.556823) rgb(115cm)=(0.144759,0.519093,0.556572) rgb(116cm)=(0.143343,0.522773,0.556295) rgb(117cm)=(0.141935,0.526453,0.555991) rgb(118cm)=(0.140536,0.530132,0.555659) rgb(119cm)=(0.139147,0.533812,0.555298) rgb(120cm)=(0.13777,0.537492,0.554906) rgb(121cm)=(0.136408,0.541173,0.554483) rgb(122cm)=(0.135066,0.544853,0.554029) rgb(123cm)=(0.133743,0.548535,0.553541) rgb(124cm)=(0.132444,0.552216,0.553018) rgb(125cm)=(0.131172,0.555899,0.552459) rgb(126cm)=(0.129933,0.559582,0.551864) rgb(127cm)=(0.128729,0.563265,0.551229) rgb(128cm)=(0.127568,0.566949,0.550556) rgb(129cm)=(0.126453,0.570633,0.549841) rgb(130cm)=(0.125394,0.574318,0.549086) rgb(131cm)=(0.124395,0.578002,0.548287) rgb(132cm)=(0.123463,0.581687,0.547445) rgb(133cm)=(0.122606,0.585371,0.546557) rgb(134cm)=(0.121831,0.589055,0.545623) rgb(135cm)=(0.121148,0.592739,0.544641) rgb(136cm)=(0.120565,0.596422,0.543611) rgb(137cm)=(0.120092,0.600104,0.54253) rgb(138cm)=(0.119738,0.603785,0.5414) rgb(139cm)=(0.119512,0.607464,0.540218) rgb(140cm)=(0.119423,0.611141,0.538982) rgb(141cm)=(0.119483,0.614817,0.537692) rgb(142cm)=(0.119699,0.61849,0.536347) rgb(143cm)=(0.120081,0.622161,0.534946) rgb(144cm)=(0.120638,0.625828,0.533488) rgb(145cm)=(0.12138,0.629492,0.531973) rgb(146cm)=(0.122312,0.633153,0.530398) rgb(147cm)=(0.123444,0.636809,0.528763) rgb(148cm)=(0.12478,0.640461,0.527068) rgb(149cm)=(0.126326,0.644107,0.525311) rgb(150cm)=(0.128087,0.647749,0.523491) rgb(151cm)=(0.130067,0.651384,0.521608) rgb(152cm)=(0.132268,0.655014,0.519661) rgb(153cm)=(0.134692,0.658636,0.517649) rgb(154cm)=(0.137339,0.662252,0.515571) rgb(155cm)=(0.14021,0.665859,0.513427) rgb(156cm)=(0.143303,0.669459,0.511215) rgb(157cm)=(0.146616,0.67305,0.508936) rgb(158cm)=(0.150148,0.676631,0.506589) rgb(159cm)=(0.153894,0.680203,0.504172) rgb(160cm)=(0.157851,0.683765,0.501686) rgb(161cm)=(0.162016,0.687316,0.499129) rgb(162cm)=(0.166383,0.690856,0.496502) rgb(163cm)=(0.170948,0.694384,0.493803) rgb(164cm)=(0.175707,0.6979,0.491033) rgb(165cm)=(0.180653,0.701402,0.488189) rgb(166cm)=(0.185783,0.704891,0.485273) rgb(167cm)=(0.19109,0.708366,0.482284) rgb(168cm)=(0.196571,0.711827,0.479221) rgb(169cm)=(0.202219,0.715272,0.476084) rgb(170cm)=(0.20803,0.718701,0.472873) rgb(171cm)=(0.214,0.722114,0.469588) rgb(172cm)=(0.220124,0.725509,0.466226) rgb(173cm)=(0.226397,0.728888,0.462789) rgb(174cm)=(0.232815,0.732247,0.459277) rgb(175cm)=(0.239374,0.735588,0.455688) rgb(176cm)=(0.24607,0.73891,0.452024) rgb(177cm)=(0.252899,0.742211,0.448284) rgb(178cm)=(0.259857,0.745492,0.444467) rgb(179cm)=(0.266941,0.748751,0.440573) rgb(180cm)=(0.274149,0.751988,0.436601) rgb(181cm)=(0.281477,0.755203,0.432552) rgb(182cm)=(0.288921,0.758394,0.428426) rgb(183cm)=(0.296479,0.761561,0.424223) rgb(184cm)=(0.304148,0.764704,0.419943) rgb(185cm)=(0.311925,0.767822,0.415586) rgb(186cm)=(0.319809,0.770914,0.411152) rgb(187cm)=(0.327796,0.77398,0.40664) rgb(188cm)=(0.335885,0.777018,0.402049) rgb(189cm)=(0.344074,0.780029,0.397381) rgb(190cm)=(0.35236,0.783011,0.392636) rgb(191cm)=(0.360741,0.785964,0.387814) rgb(192cm)=(0.369214,0.788888,0.382914) rgb(193cm)=(0.377779,0.791781,0.377939) rgb(194cm)=(0.386433,0.794644,0.372886) rgb(195cm)=(0.395174,0.797475,0.367757) rgb(196cm)=(0.404001,0.800275,0.362552) rgb(197cm)=(0.412913,0.803041,0.357269) rgb(198cm)=(0.421908,0.805774,0.35191) rgb(199cm)=(0.430983,0.808473,0.346476) rgb(200cm)=(0.440137,0.811138,0.340967) rgb(201cm)=(0.449368,0.813768,0.335384) rgb(202cm)=(0.458674,0.816363,0.329727) rgb(203cm)=(0.468053,0.818921,0.323998) rgb(204cm)=(0.477504,0.821444,0.318195) rgb(205cm)=(0.487026,0.823929,0.312321) rgb(206cm)=(0.496615,0.826376,0.306377) rgb(207cm)=(0.506271,0.828786,0.300362) rgb(208cm)=(0.515992,0.831158,0.294279) rgb(209cm)=(0.525776,0.833491,0.288127) rgb(210cm)=(0.535621,0.835785,0.281908) rgb(211cm)=(0.545524,0.838039,0.275626) rgb(212cm)=(0.555484,0.840254,0.269281) rgb(213cm)=(0.565498,0.84243,0.262877) rgb(214cm)=(0.575563,0.844566,0.256415) rgb(215cm)=(0.585678,0.846661,0.249897) rgb(216cm)=(0.595839,0.848717,0.243329) rgb(217cm)=(0.606045,0.850733,0.236712) rgb(218cm)=(0.616293,0.852709,0.230052) rgb(219cm)=(0.626579,0.854645,0.223353) rgb(220cm)=(0.636902,0.856542,0.21662) rgb(221cm)=(0.647257,0.8584,0.209861) rgb(222cm)=(0.657642,0.860219,0.203082) rgb(223cm)=(0.668054,0.861999,0.196293) rgb(224cm)=(0.678489,0.863742,0.189503) rgb(225cm)=(0.688944,0.865448,0.182725) rgb(226cm)=(0.699415,0.867117,0.175971) rgb(227cm)=(0.709898,0.868751,0.169257) rgb(228cm)=(0.720391,0.87035,0.162603) rgb(229cm)=(0.730889,0.871916,0.156029) rgb(230cm)=(0.741388,0.873449,0.149561) rgb(231cm)=(0.751884,0.874951,0.143228) rgb(232cm)=(0.762373,0.876424,0.137064) rgb(233cm)=(0.772852,0.877868,0.131109) rgb(234cm)=(0.783315,0.879285,0.125405) rgb(235cm)=(0.79376,0.880678,0.120005) rgb(236cm)=(0.804182,0.882046,0.114965) rgb(237cm)=(0.814576,0.883393,0.110347) rgb(238cm)=(0.82494,0.88472,0.106217) rgb(239cm)=(0.83527,0.886029,0.102646) rgb(240cm)=(0.845561,0.887322,0.099702) rgb(241cm)=(0.85581,0.888601,0.097452) rgb(242cm)=(0.866013,0.889868,0.095953) rgb(243cm)=(0.876168,0.891125,0.09525) rgb(244cm)=(0.886271,0.892374,0.095374) rgb(245cm)=(0.89632,0.893616,0.096335) rgb(246cm)=(0.906311,0.894855,0.098125) rgb(247cm)=(0.916242,0.896091,0.100717) rgb(248cm)=(0.926106,0.89733,0.104071) rgb(249cm)=(0.935904,0.89857,0.108131) rgb(250cm)=(0.945636,0.899815,0.112838) rgb(251cm)=(0.9553,0.901065,0.118128) rgb(252cm)=(0.964894,0.902323,0.123941) rgb(253cm)=(0.974417,0.90359,0.130215) rgb(254cm)=(0.983868,0.904867,0.136897) rgb(255cm)=(0.993248,0.906157,0.143936) }, colorbar]\addplot [point meta min=0.0, point meta max=1.0] graphics [xmin=0.0, xmax=40.0, ymin=-8000, ymax=8000] {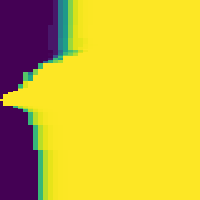};
    \end{groupplot}
        
    \end{tikzpicture}
    \caption{State space partition (same color scheme as \cref{fig:adap_comparison}) and overapproximated probability of NMAC with no online overapproximation reduction. The intruder vertical rate is fixed at $-90$ ft/s, the ownship vertical rate is 0 ft/s, and the previous advisory is $\textsc{coc}$. \label{fig:adap_only}}
\end{figure}

After obtaining the overapproximated network policy using adaptive verification, we can run model checking on the cells. \Cref{fig:adap_only} shows the results for a slice of the state space. Assuming that an encounter gets initiated with 40 seconds until loss of horizontal separation, a safe policy should have a low probability of NMAC for all states at $\tau = 40$ seconds; however, the probability values at $\tau = 40$ seconds after model checking are close to one, and we therefore cannot provide a safety guarantee with this adaptive verification method alone. We need to introduce the online error reduction techniques.

\subsection{Online Reduction}
\begin{figure}[htb]
    \begin{tikzpicture}
    \begin{groupplot}[group style={horizontal sep = 0.7cm, vertical sep = 2.5cm, group size=2 by 1}]
    \nextgroupplot [height = {5cm}, ylabel = {$h$ (ft)}, title = {Number of Advisories}, xmin = {0.0}, xmax = {40.0}, ymax = {8000}, xlabel = {$\tau$ (s)}, ymin = {-8000}, width = {5.7cm}, enlargelimits = false, axis on top]\addplot [point meta min=0, point meta max=3] graphics [xmin=0.0, xmax=40.0, ymin=-8000, ymax=8000] {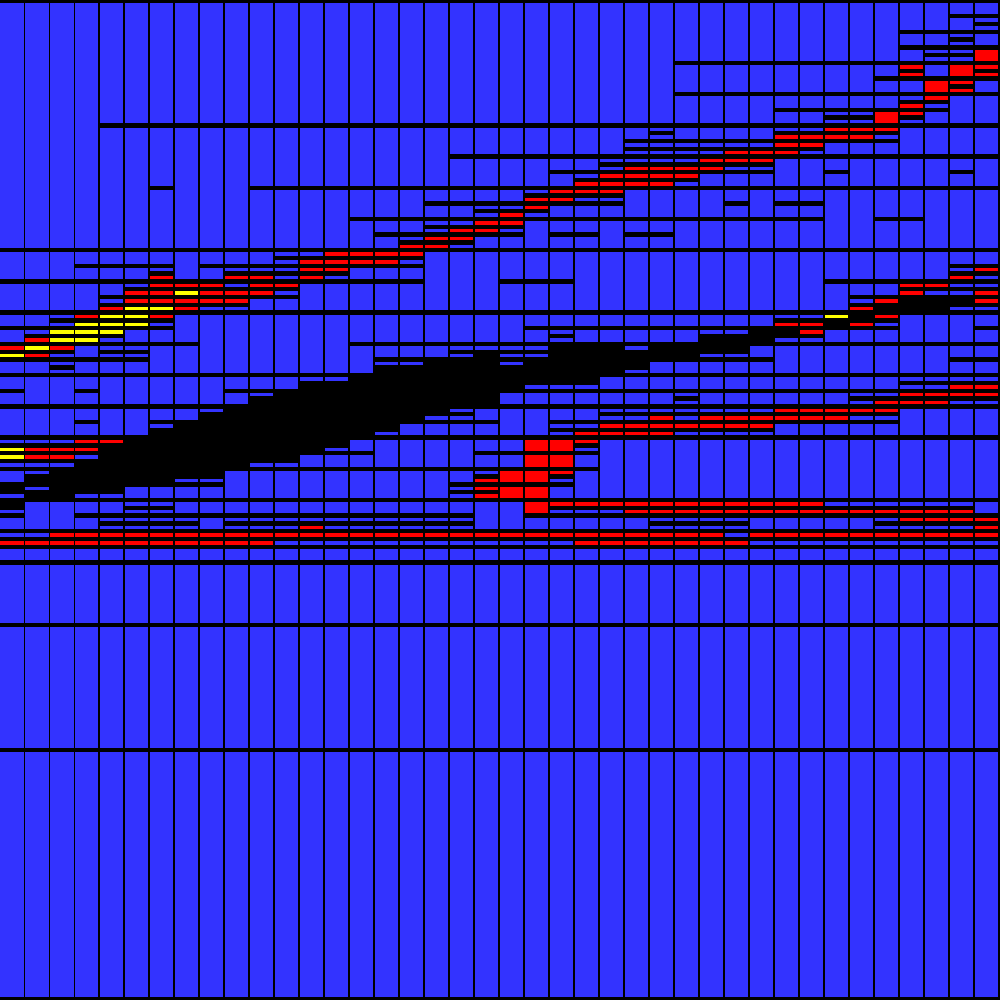};
    \nextgroupplot [height = {5cm}, title = {Probability of NMAC}, xmin = {0.0}, xmax = {40.0}, ymax = {8000}, xlabel = {$\tau$ (s)}, ymin = {-8000}, width = {5.7cm}, enlargelimits = false, axis on top, yticklabels={,,}, colorbar style={
                width=0.3cm}, colormap={mycolormap}{ rgb(0cm)=(0.267004,0.004874,0.329415) rgb(1cm)=(0.26851,0.009605,0.335427) rgb(2cm)=(0.269944,0.014625,0.341379) rgb(3cm)=(0.271305,0.019942,0.347269) rgb(4cm)=(0.272594,0.025563,0.353093) rgb(5cm)=(0.273809,0.031497,0.358853) rgb(6cm)=(0.274952,0.037752,0.364543) rgb(7cm)=(0.276022,0.044167,0.370164) rgb(8cm)=(0.277018,0.050344,0.375715) rgb(9cm)=(0.277941,0.056324,0.381191) rgb(10cm)=(0.278791,0.062145,0.386592) rgb(11cm)=(0.279566,0.067836,0.391917) rgb(12cm)=(0.280267,0.073417,0.397163) rgb(13cm)=(0.280894,0.078907,0.402329) rgb(14cm)=(0.281446,0.08432,0.407414) rgb(15cm)=(0.281924,0.089666,0.412415) rgb(16cm)=(0.282327,0.094955,0.417331) rgb(17cm)=(0.282656,0.100196,0.42216) rgb(18cm)=(0.28291,0.105393,0.426902) rgb(19cm)=(0.283091,0.110553,0.431554) rgb(20cm)=(0.283197,0.11568,0.436115) rgb(21cm)=(0.283229,0.120777,0.440584) rgb(22cm)=(0.283187,0.125848,0.44496) rgb(23cm)=(0.283072,0.130895,0.449241) rgb(24cm)=(0.282884,0.13592,0.453427) rgb(25cm)=(0.282623,0.140926,0.457517) rgb(26cm)=(0.28229,0.145912,0.46151) rgb(27cm)=(0.281887,0.150881,0.465405) rgb(28cm)=(0.281412,0.155834,0.469201) rgb(29cm)=(0.280868,0.160771,0.472899) rgb(30cm)=(0.280255,0.165693,0.476498) rgb(31cm)=(0.279574,0.170599,0.479997) rgb(32cm)=(0.278826,0.17549,0.483397) rgb(33cm)=(0.278012,0.180367,0.486697) rgb(34cm)=(0.277134,0.185228,0.489898) rgb(35cm)=(0.276194,0.190074,0.493001) rgb(36cm)=(0.275191,0.194905,0.496005) rgb(37cm)=(0.274128,0.199721,0.498911) rgb(38cm)=(0.273006,0.20452,0.501721) rgb(39cm)=(0.271828,0.209303,0.504434) rgb(40cm)=(0.270595,0.214069,0.507052) rgb(41cm)=(0.269308,0.218818,0.509577) rgb(42cm)=(0.267968,0.223549,0.512008) rgb(43cm)=(0.26658,0.228262,0.514349) rgb(44cm)=(0.265145,0.232956,0.516599) rgb(45cm)=(0.263663,0.237631,0.518762) rgb(46cm)=(0.262138,0.242286,0.520837) rgb(47cm)=(0.260571,0.246922,0.522828) rgb(48cm)=(0.258965,0.251537,0.524736) rgb(49cm)=(0.257322,0.25613,0.526563) rgb(50cm)=(0.255645,0.260703,0.528312) rgb(51cm)=(0.253935,0.265254,0.529983) rgb(52cm)=(0.252194,0.269783,0.531579) rgb(53cm)=(0.250425,0.27429,0.533103) rgb(54cm)=(0.248629,0.278775,0.534556) rgb(55cm)=(0.246811,0.283237,0.535941) rgb(56cm)=(0.244972,0.287675,0.53726) rgb(57cm)=(0.243113,0.292092,0.538516) rgb(58cm)=(0.241237,0.296485,0.539709) rgb(59cm)=(0.239346,0.300855,0.540844) rgb(60cm)=(0.237441,0.305202,0.541921) rgb(61cm)=(0.235526,0.309527,0.542944) rgb(62cm)=(0.233603,0.313828,0.543914) rgb(63cm)=(0.231674,0.318106,0.544834) rgb(64cm)=(0.229739,0.322361,0.545706) rgb(65cm)=(0.227802,0.326594,0.546532) rgb(66cm)=(0.225863,0.330805,0.547314) rgb(67cm)=(0.223925,0.334994,0.548053) rgb(68cm)=(0.221989,0.339161,0.548752) rgb(69cm)=(0.220057,0.343307,0.549413) rgb(70cm)=(0.21813,0.347432,0.550038) rgb(71cm)=(0.21621,0.351535,0.550627) rgb(72cm)=(0.214298,0.355619,0.551184) rgb(73cm)=(0.212395,0.359683,0.55171) rgb(74cm)=(0.210503,0.363727,0.552206) rgb(75cm)=(0.208623,0.367752,0.552675) rgb(76cm)=(0.206756,0.371758,0.553117) rgb(77cm)=(0.204903,0.375746,0.553533) rgb(78cm)=(0.203063,0.379716,0.553925) rgb(79cm)=(0.201239,0.38367,0.554294) rgb(80cm)=(0.19943,0.387607,0.554642) rgb(81cm)=(0.197636,0.391528,0.554969) rgb(82cm)=(0.19586,0.395433,0.555276) rgb(83cm)=(0.1941,0.399323,0.555565) rgb(84cm)=(0.192357,0.403199,0.555836) rgb(85cm)=(0.190631,0.407061,0.556089) rgb(86cm)=(0.188923,0.41091,0.556326) rgb(87cm)=(0.187231,0.414746,0.556547) rgb(88cm)=(0.185556,0.41857,0.556753) rgb(89cm)=(0.183898,0.422383,0.556944) rgb(90cm)=(0.182256,0.426184,0.55712) rgb(91cm)=(0.180629,0.429975,0.557282) rgb(92cm)=(0.179019,0.433756,0.55743) rgb(93cm)=(0.177423,0.437527,0.557565) rgb(94cm)=(0.175841,0.44129,0.557685) rgb(95cm)=(0.174274,0.445044,0.557792) rgb(96cm)=(0.172719,0.448791,0.557885) rgb(97cm)=(0.171176,0.45253,0.557965) rgb(98cm)=(0.169646,0.456262,0.55803) rgb(99cm)=(0.168126,0.459988,0.558082) rgb(100cm)=(0.166617,0.463708,0.558119) rgb(101cm)=(0.165117,0.467423,0.558141) rgb(102cm)=(0.163625,0.471133,0.558148) rgb(103cm)=(0.162142,0.474838,0.55814) rgb(104cm)=(0.160665,0.47854,0.558115) rgb(105cm)=(0.159194,0.482237,0.558073) rgb(106cm)=(0.157729,0.485932,0.558013) rgb(107cm)=(0.15627,0.489624,0.557936) rgb(108cm)=(0.154815,0.493313,0.55784) rgb(109cm)=(0.153364,0.497,0.557724) rgb(110cm)=(0.151918,0.500685,0.557587) rgb(111cm)=(0.150476,0.504369,0.55743) rgb(112cm)=(0.149039,0.508051,0.55725) rgb(113cm)=(0.147607,0.511733,0.557049) rgb(114cm)=(0.14618,0.515413,0.556823) rgb(115cm)=(0.144759,0.519093,0.556572) rgb(116cm)=(0.143343,0.522773,0.556295) rgb(117cm)=(0.141935,0.526453,0.555991) rgb(118cm)=(0.140536,0.530132,0.555659) rgb(119cm)=(0.139147,0.533812,0.555298) rgb(120cm)=(0.13777,0.537492,0.554906) rgb(121cm)=(0.136408,0.541173,0.554483) rgb(122cm)=(0.135066,0.544853,0.554029) rgb(123cm)=(0.133743,0.548535,0.553541) rgb(124cm)=(0.132444,0.552216,0.553018) rgb(125cm)=(0.131172,0.555899,0.552459) rgb(126cm)=(0.129933,0.559582,0.551864) rgb(127cm)=(0.128729,0.563265,0.551229) rgb(128cm)=(0.127568,0.566949,0.550556) rgb(129cm)=(0.126453,0.570633,0.549841) rgb(130cm)=(0.125394,0.574318,0.549086) rgb(131cm)=(0.124395,0.578002,0.548287) rgb(132cm)=(0.123463,0.581687,0.547445) rgb(133cm)=(0.122606,0.585371,0.546557) rgb(134cm)=(0.121831,0.589055,0.545623) rgb(135cm)=(0.121148,0.592739,0.544641) rgb(136cm)=(0.120565,0.596422,0.543611) rgb(137cm)=(0.120092,0.600104,0.54253) rgb(138cm)=(0.119738,0.603785,0.5414) rgb(139cm)=(0.119512,0.607464,0.540218) rgb(140cm)=(0.119423,0.611141,0.538982) rgb(141cm)=(0.119483,0.614817,0.537692) rgb(142cm)=(0.119699,0.61849,0.536347) rgb(143cm)=(0.120081,0.622161,0.534946) rgb(144cm)=(0.120638,0.625828,0.533488) rgb(145cm)=(0.12138,0.629492,0.531973) rgb(146cm)=(0.122312,0.633153,0.530398) rgb(147cm)=(0.123444,0.636809,0.528763) rgb(148cm)=(0.12478,0.640461,0.527068) rgb(149cm)=(0.126326,0.644107,0.525311) rgb(150cm)=(0.128087,0.647749,0.523491) rgb(151cm)=(0.130067,0.651384,0.521608) rgb(152cm)=(0.132268,0.655014,0.519661) rgb(153cm)=(0.134692,0.658636,0.517649) rgb(154cm)=(0.137339,0.662252,0.515571) rgb(155cm)=(0.14021,0.665859,0.513427) rgb(156cm)=(0.143303,0.669459,0.511215) rgb(157cm)=(0.146616,0.67305,0.508936) rgb(158cm)=(0.150148,0.676631,0.506589) rgb(159cm)=(0.153894,0.680203,0.504172) rgb(160cm)=(0.157851,0.683765,0.501686) rgb(161cm)=(0.162016,0.687316,0.499129) rgb(162cm)=(0.166383,0.690856,0.496502) rgb(163cm)=(0.170948,0.694384,0.493803) rgb(164cm)=(0.175707,0.6979,0.491033) rgb(165cm)=(0.180653,0.701402,0.488189) rgb(166cm)=(0.185783,0.704891,0.485273) rgb(167cm)=(0.19109,0.708366,0.482284) rgb(168cm)=(0.196571,0.711827,0.479221) rgb(169cm)=(0.202219,0.715272,0.476084) rgb(170cm)=(0.20803,0.718701,0.472873) rgb(171cm)=(0.214,0.722114,0.469588) rgb(172cm)=(0.220124,0.725509,0.466226) rgb(173cm)=(0.226397,0.728888,0.462789) rgb(174cm)=(0.232815,0.732247,0.459277) rgb(175cm)=(0.239374,0.735588,0.455688) rgb(176cm)=(0.24607,0.73891,0.452024) rgb(177cm)=(0.252899,0.742211,0.448284) rgb(178cm)=(0.259857,0.745492,0.444467) rgb(179cm)=(0.266941,0.748751,0.440573) rgb(180cm)=(0.274149,0.751988,0.436601) rgb(181cm)=(0.281477,0.755203,0.432552) rgb(182cm)=(0.288921,0.758394,0.428426) rgb(183cm)=(0.296479,0.761561,0.424223) rgb(184cm)=(0.304148,0.764704,0.419943) rgb(185cm)=(0.311925,0.767822,0.415586) rgb(186cm)=(0.319809,0.770914,0.411152) rgb(187cm)=(0.327796,0.77398,0.40664) rgb(188cm)=(0.335885,0.777018,0.402049) rgb(189cm)=(0.344074,0.780029,0.397381) rgb(190cm)=(0.35236,0.783011,0.392636) rgb(191cm)=(0.360741,0.785964,0.387814) rgb(192cm)=(0.369214,0.788888,0.382914) rgb(193cm)=(0.377779,0.791781,0.377939) rgb(194cm)=(0.386433,0.794644,0.372886) rgb(195cm)=(0.395174,0.797475,0.367757) rgb(196cm)=(0.404001,0.800275,0.362552) rgb(197cm)=(0.412913,0.803041,0.357269) rgb(198cm)=(0.421908,0.805774,0.35191) rgb(199cm)=(0.430983,0.808473,0.346476) rgb(200cm)=(0.440137,0.811138,0.340967) rgb(201cm)=(0.449368,0.813768,0.335384) rgb(202cm)=(0.458674,0.816363,0.329727) rgb(203cm)=(0.468053,0.818921,0.323998) rgb(204cm)=(0.477504,0.821444,0.318195) rgb(205cm)=(0.487026,0.823929,0.312321) rgb(206cm)=(0.496615,0.826376,0.306377) rgb(207cm)=(0.506271,0.828786,0.300362) rgb(208cm)=(0.515992,0.831158,0.294279) rgb(209cm)=(0.525776,0.833491,0.288127) rgb(210cm)=(0.535621,0.835785,0.281908) rgb(211cm)=(0.545524,0.838039,0.275626) rgb(212cm)=(0.555484,0.840254,0.269281) rgb(213cm)=(0.565498,0.84243,0.262877) rgb(214cm)=(0.575563,0.844566,0.256415) rgb(215cm)=(0.585678,0.846661,0.249897) rgb(216cm)=(0.595839,0.848717,0.243329) rgb(217cm)=(0.606045,0.850733,0.236712) rgb(218cm)=(0.616293,0.852709,0.230052) rgb(219cm)=(0.626579,0.854645,0.223353) rgb(220cm)=(0.636902,0.856542,0.21662) rgb(221cm)=(0.647257,0.8584,0.209861) rgb(222cm)=(0.657642,0.860219,0.203082) rgb(223cm)=(0.668054,0.861999,0.196293) rgb(224cm)=(0.678489,0.863742,0.189503) rgb(225cm)=(0.688944,0.865448,0.182725) rgb(226cm)=(0.699415,0.867117,0.175971) rgb(227cm)=(0.709898,0.868751,0.169257) rgb(228cm)=(0.720391,0.87035,0.162603) rgb(229cm)=(0.730889,0.871916,0.156029) rgb(230cm)=(0.741388,0.873449,0.149561) rgb(231cm)=(0.751884,0.874951,0.143228) rgb(232cm)=(0.762373,0.876424,0.137064) rgb(233cm)=(0.772852,0.877868,0.131109) rgb(234cm)=(0.783315,0.879285,0.125405) rgb(235cm)=(0.79376,0.880678,0.120005) rgb(236cm)=(0.804182,0.882046,0.114965) rgb(237cm)=(0.814576,0.883393,0.110347) rgb(238cm)=(0.82494,0.88472,0.106217) rgb(239cm)=(0.83527,0.886029,0.102646) rgb(240cm)=(0.845561,0.887322,0.099702) rgb(241cm)=(0.85581,0.888601,0.097452) rgb(242cm)=(0.866013,0.889868,0.095953) rgb(243cm)=(0.876168,0.891125,0.09525) rgb(244cm)=(0.886271,0.892374,0.095374) rgb(245cm)=(0.89632,0.893616,0.096335) rgb(246cm)=(0.906311,0.894855,0.098125) rgb(247cm)=(0.916242,0.896091,0.100717) rgb(248cm)=(0.926106,0.89733,0.104071) rgb(249cm)=(0.935904,0.89857,0.108131) rgb(250cm)=(0.945636,0.899815,0.112838) rgb(251cm)=(0.9553,0.901065,0.118128) rgb(252cm)=(0.964894,0.902323,0.123941) rgb(253cm)=(0.974417,0.90359,0.130215) rgb(254cm)=(0.983868,0.904867,0.136897) rgb(255cm)=(0.993248,0.906157,0.143936) }, colorbar]\addplot [point meta min=0.0, point meta max=1.0] graphics [xmin=0.0, xmax=40.0, ymin=-8000, ymax=8000] {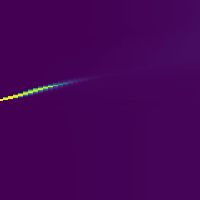};
    \end{groupplot}
    
    \end{tikzpicture}
    \caption{State space partition (same color scheme as \cref{fig:adap_comparison}) and overapproximated probability of NMAC with the worst-case transition splitting heuristic. The intruder vertical rate is fixed at $-90$ ft/s, the ownship vertical rate is 0 ft/s, and the previous advisory is $\textsc{coc}$. \label{fig:dynamics}}
\end{figure}

\Cref{fig:dynamics} displays the model checking results with the online worst-case transition splitting heuristic for the same slice of the state space shown in \cref{fig:adap_only}.
The algorithm splits cells in a densely packed band along the safety-critical region of the state space where a collision is imminent. The intruder is rapidly descending at 90 ft/s, so the region directly above the ownship is most dangerous. Outside of this band, the state space partition remains untouched, indicating that the online splitting heuristic only splits states that require a finer resolution.

\begin{figure}[htb]
    \begin{tikzpicture}
    \begin{groupplot}[group style={horizontal sep = 0.4cm, vertical sep = 0.5cm, group size=3 by 2}]
    \nextgroupplot [height = {5cm}, ylabel = {$h$ (ft)}, title = {Threshold = 0.5}, xmin = {0.0}, xmax = {40.0}, ymax = {8000}, ymin = {-8000}, width = {4.35cm}, enlargelimits = false, axis on top, xticklabels={,,}]\addplot [point meta min=0, point meta max=3] graphics [xmin=0.0, xmax=40.0, ymin=-8000, ymax=8000] {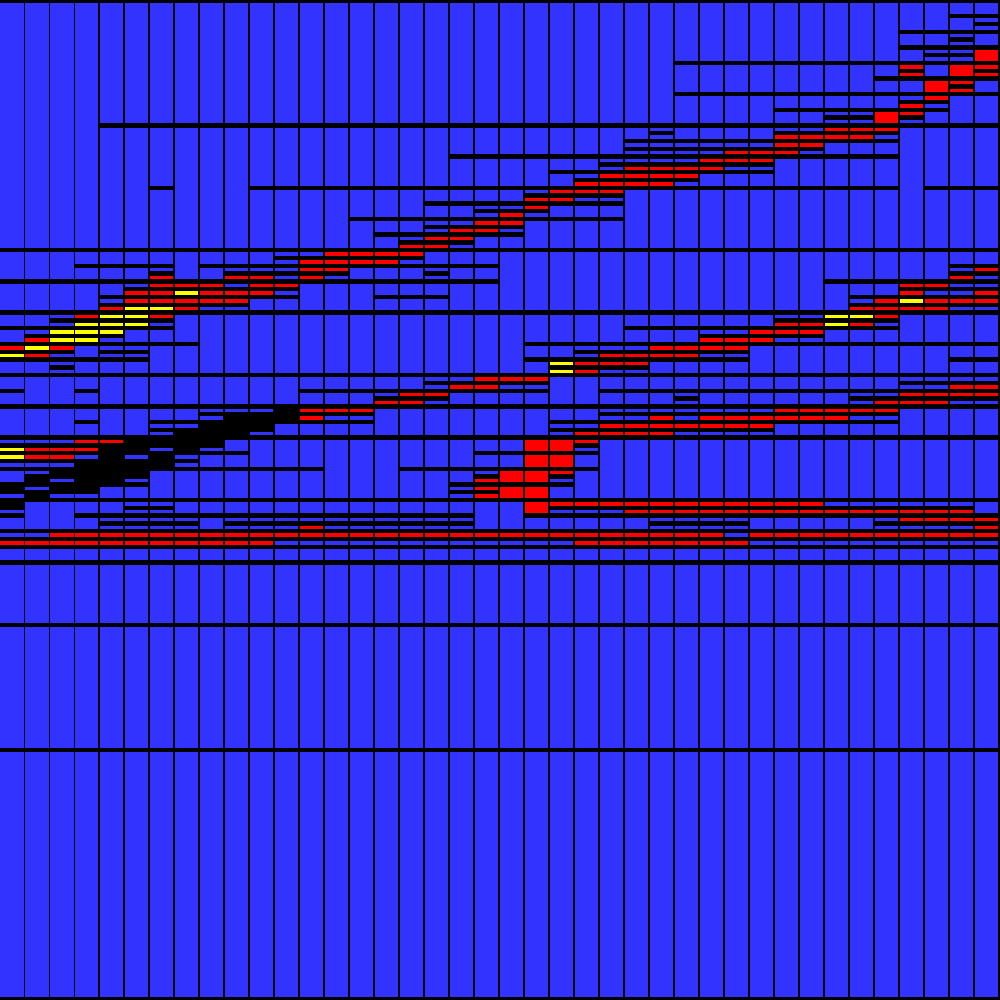};
    \nextgroupplot [height = {5cm}, title = {Threshold = 0.1}, xmin = {0.0}, xmax = {40.0}, ymax = {8000}, ymin = {-8000}, width = {4.35cm}, enlargelimits = false, axis on top, yticklabels={,,}, xticklabels={,,}]\addplot [point meta min=0, point meta max=3] graphics [xmin=0.0, xmax=40.0, ymin=-8000, ymax=8000] {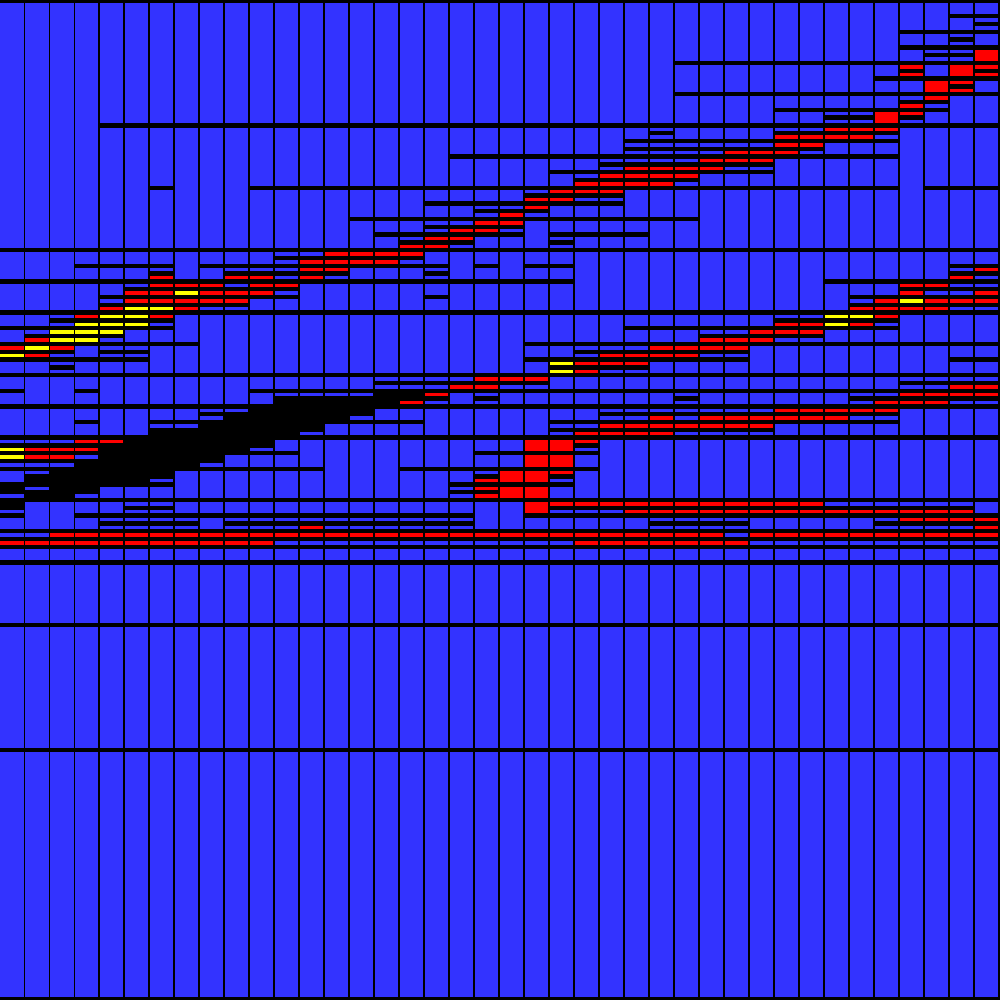};
    \nextgroupplot [height = {5cm}, title = {Threshold = 0.005}, xmin = {0.0}, xmax = {40.0}, ymax = {8000}, ymin = {-8000}, width = {4.35cm}, enlargelimits = false, axis on top, yticklabels={,,}, xticklabels={,,}]\addplot [point meta min=0, point meta max=3] graphics [xmin=0.0, xmax=40.0, ymin=-8000, ymax=8000] {dynamics1.png};
    \nextgroupplot [height = {5cm}, ylabel = {$h$ (ft)}, xmin = {0.0}, xmax = {40.0}, ymax = {8000}, xlabel = {$\tau$ (s)}, ymin = {-8000}, width = {4.35cm}, enlargelimits = false, axis on top]\addplot [point meta min=0.0, point meta max=1.0] graphics [xmin=0.0, xmax=40.0, ymin=-8000, ymax=8000] {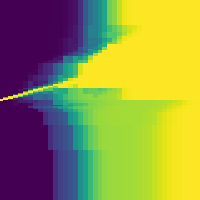};
    \nextgroupplot [height = {5cm}, xmin = {0.0}, xmax = {40.0}, ymax = {8000}, xlabel = {$\tau$ (s)}, ymin = {-8000}, width = {4.35cm}, enlargelimits = false, axis on top, yticklabels={,,}]\addplot [point meta min=0.0, point meta max=1.0] graphics [xmin=0.0, xmax=40.0, ymin=-8000, ymax=8000] {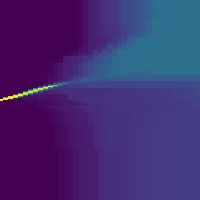};
    \nextgroupplot [height = {5cm}, xmin = {0.0}, xmax = {40.0}, ymax = {8000}, xlabel = {$\tau$ (s)}, ymin = {-8000}, width = {4.35cm}, enlargelimits = false, axis on top, yticklabels={,,}, colorbar style={
                width=0.3cm}, colormap={mycolormap}{ rgb(0cm)=(0.267004,0.004874,0.329415) rgb(1cm)=(0.26851,0.009605,0.335427) rgb(2cm)=(0.269944,0.014625,0.341379) rgb(3cm)=(0.271305,0.019942,0.347269) rgb(4cm)=(0.272594,0.025563,0.353093) rgb(5cm)=(0.273809,0.031497,0.358853) rgb(6cm)=(0.274952,0.037752,0.364543) rgb(7cm)=(0.276022,0.044167,0.370164) rgb(8cm)=(0.277018,0.050344,0.375715) rgb(9cm)=(0.277941,0.056324,0.381191) rgb(10cm)=(0.278791,0.062145,0.386592) rgb(11cm)=(0.279566,0.067836,0.391917) rgb(12cm)=(0.280267,0.073417,0.397163) rgb(13cm)=(0.280894,0.078907,0.402329) rgb(14cm)=(0.281446,0.08432,0.407414) rgb(15cm)=(0.281924,0.089666,0.412415) rgb(16cm)=(0.282327,0.094955,0.417331) rgb(17cm)=(0.282656,0.100196,0.42216) rgb(18cm)=(0.28291,0.105393,0.426902) rgb(19cm)=(0.283091,0.110553,0.431554) rgb(20cm)=(0.283197,0.11568,0.436115) rgb(21cm)=(0.283229,0.120777,0.440584) rgb(22cm)=(0.283187,0.125848,0.44496) rgb(23cm)=(0.283072,0.130895,0.449241) rgb(24cm)=(0.282884,0.13592,0.453427) rgb(25cm)=(0.282623,0.140926,0.457517) rgb(26cm)=(0.28229,0.145912,0.46151) rgb(27cm)=(0.281887,0.150881,0.465405) rgb(28cm)=(0.281412,0.155834,0.469201) rgb(29cm)=(0.280868,0.160771,0.472899) rgb(30cm)=(0.280255,0.165693,0.476498) rgb(31cm)=(0.279574,0.170599,0.479997) rgb(32cm)=(0.278826,0.17549,0.483397) rgb(33cm)=(0.278012,0.180367,0.486697) rgb(34cm)=(0.277134,0.185228,0.489898) rgb(35cm)=(0.276194,0.190074,0.493001) rgb(36cm)=(0.275191,0.194905,0.496005) rgb(37cm)=(0.274128,0.199721,0.498911) rgb(38cm)=(0.273006,0.20452,0.501721) rgb(39cm)=(0.271828,0.209303,0.504434) rgb(40cm)=(0.270595,0.214069,0.507052) rgb(41cm)=(0.269308,0.218818,0.509577) rgb(42cm)=(0.267968,0.223549,0.512008) rgb(43cm)=(0.26658,0.228262,0.514349) rgb(44cm)=(0.265145,0.232956,0.516599) rgb(45cm)=(0.263663,0.237631,0.518762) rgb(46cm)=(0.262138,0.242286,0.520837) rgb(47cm)=(0.260571,0.246922,0.522828) rgb(48cm)=(0.258965,0.251537,0.524736) rgb(49cm)=(0.257322,0.25613,0.526563) rgb(50cm)=(0.255645,0.260703,0.528312) rgb(51cm)=(0.253935,0.265254,0.529983) rgb(52cm)=(0.252194,0.269783,0.531579) rgb(53cm)=(0.250425,0.27429,0.533103) rgb(54cm)=(0.248629,0.278775,0.534556) rgb(55cm)=(0.246811,0.283237,0.535941) rgb(56cm)=(0.244972,0.287675,0.53726) rgb(57cm)=(0.243113,0.292092,0.538516) rgb(58cm)=(0.241237,0.296485,0.539709) rgb(59cm)=(0.239346,0.300855,0.540844) rgb(60cm)=(0.237441,0.305202,0.541921) rgb(61cm)=(0.235526,0.309527,0.542944) rgb(62cm)=(0.233603,0.313828,0.543914) rgb(63cm)=(0.231674,0.318106,0.544834) rgb(64cm)=(0.229739,0.322361,0.545706) rgb(65cm)=(0.227802,0.326594,0.546532) rgb(66cm)=(0.225863,0.330805,0.547314) rgb(67cm)=(0.223925,0.334994,0.548053) rgb(68cm)=(0.221989,0.339161,0.548752) rgb(69cm)=(0.220057,0.343307,0.549413) rgb(70cm)=(0.21813,0.347432,0.550038) rgb(71cm)=(0.21621,0.351535,0.550627) rgb(72cm)=(0.214298,0.355619,0.551184) rgb(73cm)=(0.212395,0.359683,0.55171) rgb(74cm)=(0.210503,0.363727,0.552206) rgb(75cm)=(0.208623,0.367752,0.552675) rgb(76cm)=(0.206756,0.371758,0.553117) rgb(77cm)=(0.204903,0.375746,0.553533) rgb(78cm)=(0.203063,0.379716,0.553925) rgb(79cm)=(0.201239,0.38367,0.554294) rgb(80cm)=(0.19943,0.387607,0.554642) rgb(81cm)=(0.197636,0.391528,0.554969) rgb(82cm)=(0.19586,0.395433,0.555276) rgb(83cm)=(0.1941,0.399323,0.555565) rgb(84cm)=(0.192357,0.403199,0.555836) rgb(85cm)=(0.190631,0.407061,0.556089) rgb(86cm)=(0.188923,0.41091,0.556326) rgb(87cm)=(0.187231,0.414746,0.556547) rgb(88cm)=(0.185556,0.41857,0.556753) rgb(89cm)=(0.183898,0.422383,0.556944) rgb(90cm)=(0.182256,0.426184,0.55712) rgb(91cm)=(0.180629,0.429975,0.557282) rgb(92cm)=(0.179019,0.433756,0.55743) rgb(93cm)=(0.177423,0.437527,0.557565) rgb(94cm)=(0.175841,0.44129,0.557685) rgb(95cm)=(0.174274,0.445044,0.557792) rgb(96cm)=(0.172719,0.448791,0.557885) rgb(97cm)=(0.171176,0.45253,0.557965) rgb(98cm)=(0.169646,0.456262,0.55803) rgb(99cm)=(0.168126,0.459988,0.558082) rgb(100cm)=(0.166617,0.463708,0.558119) rgb(101cm)=(0.165117,0.467423,0.558141) rgb(102cm)=(0.163625,0.471133,0.558148) rgb(103cm)=(0.162142,0.474838,0.55814) rgb(104cm)=(0.160665,0.47854,0.558115) rgb(105cm)=(0.159194,0.482237,0.558073) rgb(106cm)=(0.157729,0.485932,0.558013) rgb(107cm)=(0.15627,0.489624,0.557936) rgb(108cm)=(0.154815,0.493313,0.55784) rgb(109cm)=(0.153364,0.497,0.557724) rgb(110cm)=(0.151918,0.500685,0.557587) rgb(111cm)=(0.150476,0.504369,0.55743) rgb(112cm)=(0.149039,0.508051,0.55725) rgb(113cm)=(0.147607,0.511733,0.557049) rgb(114cm)=(0.14618,0.515413,0.556823) rgb(115cm)=(0.144759,0.519093,0.556572) rgb(116cm)=(0.143343,0.522773,0.556295) rgb(117cm)=(0.141935,0.526453,0.555991) rgb(118cm)=(0.140536,0.530132,0.555659) rgb(119cm)=(0.139147,0.533812,0.555298) rgb(120cm)=(0.13777,0.537492,0.554906) rgb(121cm)=(0.136408,0.541173,0.554483) rgb(122cm)=(0.135066,0.544853,0.554029) rgb(123cm)=(0.133743,0.548535,0.553541) rgb(124cm)=(0.132444,0.552216,0.553018) rgb(125cm)=(0.131172,0.555899,0.552459) rgb(126cm)=(0.129933,0.559582,0.551864) rgb(127cm)=(0.128729,0.563265,0.551229) rgb(128cm)=(0.127568,0.566949,0.550556) rgb(129cm)=(0.126453,0.570633,0.549841) rgb(130cm)=(0.125394,0.574318,0.549086) rgb(131cm)=(0.124395,0.578002,0.548287) rgb(132cm)=(0.123463,0.581687,0.547445) rgb(133cm)=(0.122606,0.585371,0.546557) rgb(134cm)=(0.121831,0.589055,0.545623) rgb(135cm)=(0.121148,0.592739,0.544641) rgb(136cm)=(0.120565,0.596422,0.543611) rgb(137cm)=(0.120092,0.600104,0.54253) rgb(138cm)=(0.119738,0.603785,0.5414) rgb(139cm)=(0.119512,0.607464,0.540218) rgb(140cm)=(0.119423,0.611141,0.538982) rgb(141cm)=(0.119483,0.614817,0.537692) rgb(142cm)=(0.119699,0.61849,0.536347) rgb(143cm)=(0.120081,0.622161,0.534946) rgb(144cm)=(0.120638,0.625828,0.533488) rgb(145cm)=(0.12138,0.629492,0.531973) rgb(146cm)=(0.122312,0.633153,0.530398) rgb(147cm)=(0.123444,0.636809,0.528763) rgb(148cm)=(0.12478,0.640461,0.527068) rgb(149cm)=(0.126326,0.644107,0.525311) rgb(150cm)=(0.128087,0.647749,0.523491) rgb(151cm)=(0.130067,0.651384,0.521608) rgb(152cm)=(0.132268,0.655014,0.519661) rgb(153cm)=(0.134692,0.658636,0.517649) rgb(154cm)=(0.137339,0.662252,0.515571) rgb(155cm)=(0.14021,0.665859,0.513427) rgb(156cm)=(0.143303,0.669459,0.511215) rgb(157cm)=(0.146616,0.67305,0.508936) rgb(158cm)=(0.150148,0.676631,0.506589) rgb(159cm)=(0.153894,0.680203,0.504172) rgb(160cm)=(0.157851,0.683765,0.501686) rgb(161cm)=(0.162016,0.687316,0.499129) rgb(162cm)=(0.166383,0.690856,0.496502) rgb(163cm)=(0.170948,0.694384,0.493803) rgb(164cm)=(0.175707,0.6979,0.491033) rgb(165cm)=(0.180653,0.701402,0.488189) rgb(166cm)=(0.185783,0.704891,0.485273) rgb(167cm)=(0.19109,0.708366,0.482284) rgb(168cm)=(0.196571,0.711827,0.479221) rgb(169cm)=(0.202219,0.715272,0.476084) rgb(170cm)=(0.20803,0.718701,0.472873) rgb(171cm)=(0.214,0.722114,0.469588) rgb(172cm)=(0.220124,0.725509,0.466226) rgb(173cm)=(0.226397,0.728888,0.462789) rgb(174cm)=(0.232815,0.732247,0.459277) rgb(175cm)=(0.239374,0.735588,0.455688) rgb(176cm)=(0.24607,0.73891,0.452024) rgb(177cm)=(0.252899,0.742211,0.448284) rgb(178cm)=(0.259857,0.745492,0.444467) rgb(179cm)=(0.266941,0.748751,0.440573) rgb(180cm)=(0.274149,0.751988,0.436601) rgb(181cm)=(0.281477,0.755203,0.432552) rgb(182cm)=(0.288921,0.758394,0.428426) rgb(183cm)=(0.296479,0.761561,0.424223) rgb(184cm)=(0.304148,0.764704,0.419943) rgb(185cm)=(0.311925,0.767822,0.415586) rgb(186cm)=(0.319809,0.770914,0.411152) rgb(187cm)=(0.327796,0.77398,0.40664) rgb(188cm)=(0.335885,0.777018,0.402049) rgb(189cm)=(0.344074,0.780029,0.397381) rgb(190cm)=(0.35236,0.783011,0.392636) rgb(191cm)=(0.360741,0.785964,0.387814) rgb(192cm)=(0.369214,0.788888,0.382914) rgb(193cm)=(0.377779,0.791781,0.377939) rgb(194cm)=(0.386433,0.794644,0.372886) rgb(195cm)=(0.395174,0.797475,0.367757) rgb(196cm)=(0.404001,0.800275,0.362552) rgb(197cm)=(0.412913,0.803041,0.357269) rgb(198cm)=(0.421908,0.805774,0.35191) rgb(199cm)=(0.430983,0.808473,0.346476) rgb(200cm)=(0.440137,0.811138,0.340967) rgb(201cm)=(0.449368,0.813768,0.335384) rgb(202cm)=(0.458674,0.816363,0.329727) rgb(203cm)=(0.468053,0.818921,0.323998) rgb(204cm)=(0.477504,0.821444,0.318195) rgb(205cm)=(0.487026,0.823929,0.312321) rgb(206cm)=(0.496615,0.826376,0.306377) rgb(207cm)=(0.506271,0.828786,0.300362) rgb(208cm)=(0.515992,0.831158,0.294279) rgb(209cm)=(0.525776,0.833491,0.288127) rgb(210cm)=(0.535621,0.835785,0.281908) rgb(211cm)=(0.545524,0.838039,0.275626) rgb(212cm)=(0.555484,0.840254,0.269281) rgb(213cm)=(0.565498,0.84243,0.262877) rgb(214cm)=(0.575563,0.844566,0.256415) rgb(215cm)=(0.585678,0.846661,0.249897) rgb(216cm)=(0.595839,0.848717,0.243329) rgb(217cm)=(0.606045,0.850733,0.236712) rgb(218cm)=(0.616293,0.852709,0.230052) rgb(219cm)=(0.626579,0.854645,0.223353) rgb(220cm)=(0.636902,0.856542,0.21662) rgb(221cm)=(0.647257,0.8584,0.209861) rgb(222cm)=(0.657642,0.860219,0.203082) rgb(223cm)=(0.668054,0.861999,0.196293) rgb(224cm)=(0.678489,0.863742,0.189503) rgb(225cm)=(0.688944,0.865448,0.182725) rgb(226cm)=(0.699415,0.867117,0.175971) rgb(227cm)=(0.709898,0.868751,0.169257) rgb(228cm)=(0.720391,0.87035,0.162603) rgb(229cm)=(0.730889,0.871916,0.156029) rgb(230cm)=(0.741388,0.873449,0.149561) rgb(231cm)=(0.751884,0.874951,0.143228) rgb(232cm)=(0.762373,0.876424,0.137064) rgb(233cm)=(0.772852,0.877868,0.131109) rgb(234cm)=(0.783315,0.879285,0.125405) rgb(235cm)=(0.79376,0.880678,0.120005) rgb(236cm)=(0.804182,0.882046,0.114965) rgb(237cm)=(0.814576,0.883393,0.110347) rgb(238cm)=(0.82494,0.88472,0.106217) rgb(239cm)=(0.83527,0.886029,0.102646) rgb(240cm)=(0.845561,0.887322,0.099702) rgb(241cm)=(0.85581,0.888601,0.097452) rgb(242cm)=(0.866013,0.889868,0.095953) rgb(243cm)=(0.876168,0.891125,0.09525) rgb(244cm)=(0.886271,0.892374,0.095374) rgb(245cm)=(0.89632,0.893616,0.096335) rgb(246cm)=(0.906311,0.894855,0.098125) rgb(247cm)=(0.916242,0.896091,0.100717) rgb(248cm)=(0.926106,0.89733,0.104071) rgb(249cm)=(0.935904,0.89857,0.108131) rgb(250cm)=(0.945636,0.899815,0.112838) rgb(251cm)=(0.9553,0.901065,0.118128) rgb(252cm)=(0.964894,0.902323,0.123941) rgb(253cm)=(0.974417,0.90359,0.130215) rgb(254cm)=(0.983868,0.904867,0.136897) rgb(255cm)=(0.993248,0.906157,0.143936) }, colorbar]\addplot [point meta min=0.0, point meta max=1.0] graphics [xmin=0.0, xmax=40.0, ymin=-8000, ymax=8000] {dynamics2.png};
    \end{groupplot}
    
    \end{tikzpicture}
    \caption{State space partition (same color scheme as \cref{fig:adap_comparison}) and overapproximated probability of NMAC using various thresholds for the worst-case transition splitting heuristic. The intruder vertical rate is fixed at $-90$ ft/s, the ownship vertical rate is 0 ft/s, and the previous advisory is $\textsc{coc}$. \label{fig:threshold_effect}}
\end{figure}

The overapproximation error in \cref{fig:dynamics} is much lower than the error in \cref{fig:adap_only}, and we can now obtain meaningful information from the estimated probabilities. For example, the probabilities are highest at low values of $\tau$ when the intruder is above the ownship. The high intruder descent rate results in the upward sloping band of high probability extending away from the NMAC region at $\tau = 0$. At $\tau = 40$ seconds, the probabilities of collision are significantly lower with a maximum probability of 0.0305 among all cells. The splitting threshold can be tuned to achieve a desired resolution as shown in \cref{fig:threshold_effect}.
The highest threshold only splits cells in the extremely safety-critical region near $\tau = 0$. As the threshold is decreased, the band of tightly packed cells extends further away from $\tau = 0$ and overapproximation error decreases.

\begin{figure}[htb]
    \begin{tikzpicture}[]
    \begin{groupplot}[group style={horizontal sep = 0.7cm, vertical sep = 2.5cm, group size=2 by 1}]
    \nextgroupplot [height = {5cm}, ylabel = {$h$ (ft)}, title = {Number of Advisories}, xmin = {0.0}, xmax = {40.0}, ymax = {8000}, xlabel = {$\tau$ (s)}, ymin = {-8000}, width = {5.7cm}, enlargelimits = false, axis on top]\addplot [point meta min=0, point meta max=3] graphics [xmin=0.0, xmax=40.0, ymin=-8000, ymax=8000] {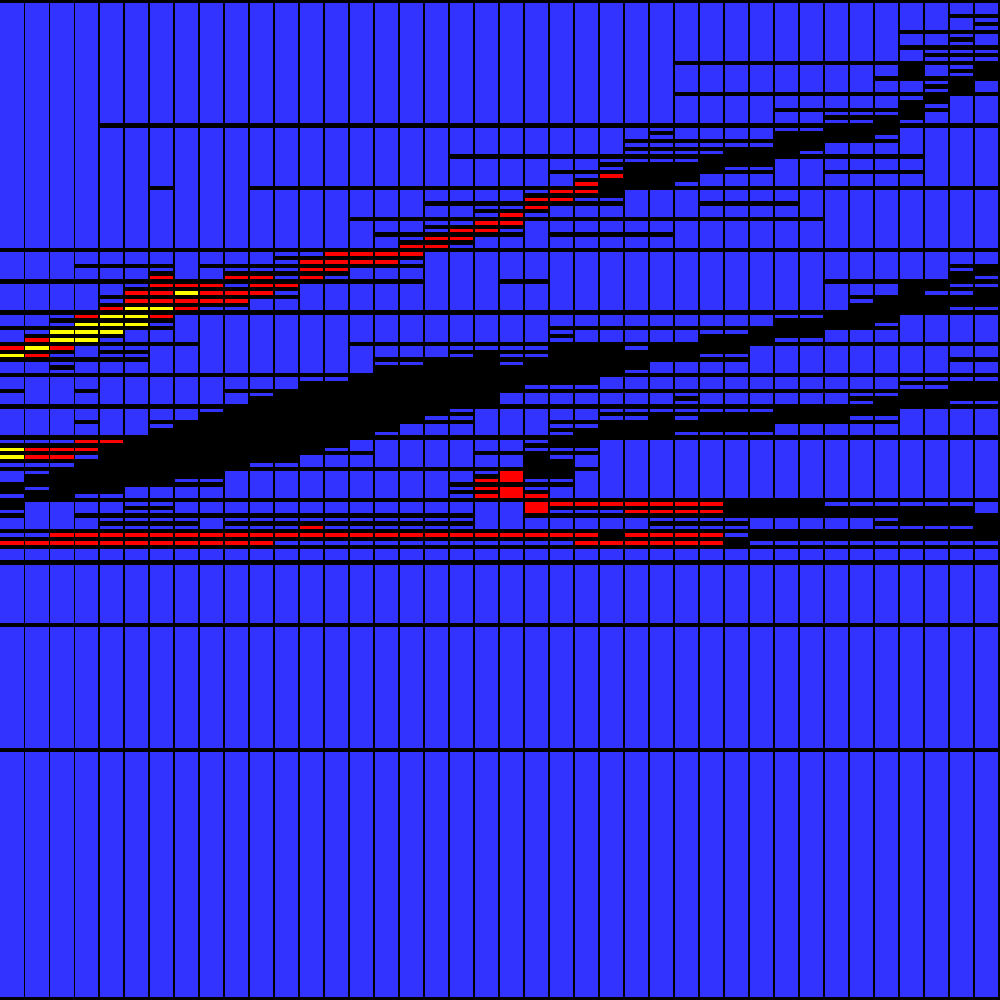};
    \nextgroupplot [height = {5cm}, title = {Probability of NMAC}, xmin = {0.0}, xmax = {40.0}, ymax = {8000}, xlabel = {$\tau$ (s)}, ymin = {-8000}, width = {5.7cm}, enlargelimits = false, axis on top, yticklabels={,,}, colorbar style={
                width=0.3cm}, colormap={mycolormap}{ rgb(0cm)=(0.267004,0.004874,0.329415) rgb(1cm)=(0.26851,0.009605,0.335427) rgb(2cm)=(0.269944,0.014625,0.341379) rgb(3cm)=(0.271305,0.019942,0.347269) rgb(4cm)=(0.272594,0.025563,0.353093) rgb(5cm)=(0.273809,0.031497,0.358853) rgb(6cm)=(0.274952,0.037752,0.364543) rgb(7cm)=(0.276022,0.044167,0.370164) rgb(8cm)=(0.277018,0.050344,0.375715) rgb(9cm)=(0.277941,0.056324,0.381191) rgb(10cm)=(0.278791,0.062145,0.386592) rgb(11cm)=(0.279566,0.067836,0.391917) rgb(12cm)=(0.280267,0.073417,0.397163) rgb(13cm)=(0.280894,0.078907,0.402329) rgb(14cm)=(0.281446,0.08432,0.407414) rgb(15cm)=(0.281924,0.089666,0.412415) rgb(16cm)=(0.282327,0.094955,0.417331) rgb(17cm)=(0.282656,0.100196,0.42216) rgb(18cm)=(0.28291,0.105393,0.426902) rgb(19cm)=(0.283091,0.110553,0.431554) rgb(20cm)=(0.283197,0.11568,0.436115) rgb(21cm)=(0.283229,0.120777,0.440584) rgb(22cm)=(0.283187,0.125848,0.44496) rgb(23cm)=(0.283072,0.130895,0.449241) rgb(24cm)=(0.282884,0.13592,0.453427) rgb(25cm)=(0.282623,0.140926,0.457517) rgb(26cm)=(0.28229,0.145912,0.46151) rgb(27cm)=(0.281887,0.150881,0.465405) rgb(28cm)=(0.281412,0.155834,0.469201) rgb(29cm)=(0.280868,0.160771,0.472899) rgb(30cm)=(0.280255,0.165693,0.476498) rgb(31cm)=(0.279574,0.170599,0.479997) rgb(32cm)=(0.278826,0.17549,0.483397) rgb(33cm)=(0.278012,0.180367,0.486697) rgb(34cm)=(0.277134,0.185228,0.489898) rgb(35cm)=(0.276194,0.190074,0.493001) rgb(36cm)=(0.275191,0.194905,0.496005) rgb(37cm)=(0.274128,0.199721,0.498911) rgb(38cm)=(0.273006,0.20452,0.501721) rgb(39cm)=(0.271828,0.209303,0.504434) rgb(40cm)=(0.270595,0.214069,0.507052) rgb(41cm)=(0.269308,0.218818,0.509577) rgb(42cm)=(0.267968,0.223549,0.512008) rgb(43cm)=(0.26658,0.228262,0.514349) rgb(44cm)=(0.265145,0.232956,0.516599) rgb(45cm)=(0.263663,0.237631,0.518762) rgb(46cm)=(0.262138,0.242286,0.520837) rgb(47cm)=(0.260571,0.246922,0.522828) rgb(48cm)=(0.258965,0.251537,0.524736) rgb(49cm)=(0.257322,0.25613,0.526563) rgb(50cm)=(0.255645,0.260703,0.528312) rgb(51cm)=(0.253935,0.265254,0.529983) rgb(52cm)=(0.252194,0.269783,0.531579) rgb(53cm)=(0.250425,0.27429,0.533103) rgb(54cm)=(0.248629,0.278775,0.534556) rgb(55cm)=(0.246811,0.283237,0.535941) rgb(56cm)=(0.244972,0.287675,0.53726) rgb(57cm)=(0.243113,0.292092,0.538516) rgb(58cm)=(0.241237,0.296485,0.539709) rgb(59cm)=(0.239346,0.300855,0.540844) rgb(60cm)=(0.237441,0.305202,0.541921) rgb(61cm)=(0.235526,0.309527,0.542944) rgb(62cm)=(0.233603,0.313828,0.543914) rgb(63cm)=(0.231674,0.318106,0.544834) rgb(64cm)=(0.229739,0.322361,0.545706) rgb(65cm)=(0.227802,0.326594,0.546532) rgb(66cm)=(0.225863,0.330805,0.547314) rgb(67cm)=(0.223925,0.334994,0.548053) rgb(68cm)=(0.221989,0.339161,0.548752) rgb(69cm)=(0.220057,0.343307,0.549413) rgb(70cm)=(0.21813,0.347432,0.550038) rgb(71cm)=(0.21621,0.351535,0.550627) rgb(72cm)=(0.214298,0.355619,0.551184) rgb(73cm)=(0.212395,0.359683,0.55171) rgb(74cm)=(0.210503,0.363727,0.552206) rgb(75cm)=(0.208623,0.367752,0.552675) rgb(76cm)=(0.206756,0.371758,0.553117) rgb(77cm)=(0.204903,0.375746,0.553533) rgb(78cm)=(0.203063,0.379716,0.553925) rgb(79cm)=(0.201239,0.38367,0.554294) rgb(80cm)=(0.19943,0.387607,0.554642) rgb(81cm)=(0.197636,0.391528,0.554969) rgb(82cm)=(0.19586,0.395433,0.555276) rgb(83cm)=(0.1941,0.399323,0.555565) rgb(84cm)=(0.192357,0.403199,0.555836) rgb(85cm)=(0.190631,0.407061,0.556089) rgb(86cm)=(0.188923,0.41091,0.556326) rgb(87cm)=(0.187231,0.414746,0.556547) rgb(88cm)=(0.185556,0.41857,0.556753) rgb(89cm)=(0.183898,0.422383,0.556944) rgb(90cm)=(0.182256,0.426184,0.55712) rgb(91cm)=(0.180629,0.429975,0.557282) rgb(92cm)=(0.179019,0.433756,0.55743) rgb(93cm)=(0.177423,0.437527,0.557565) rgb(94cm)=(0.175841,0.44129,0.557685) rgb(95cm)=(0.174274,0.445044,0.557792) rgb(96cm)=(0.172719,0.448791,0.557885) rgb(97cm)=(0.171176,0.45253,0.557965) rgb(98cm)=(0.169646,0.456262,0.55803) rgb(99cm)=(0.168126,0.459988,0.558082) rgb(100cm)=(0.166617,0.463708,0.558119) rgb(101cm)=(0.165117,0.467423,0.558141) rgb(102cm)=(0.163625,0.471133,0.558148) rgb(103cm)=(0.162142,0.474838,0.55814) rgb(104cm)=(0.160665,0.47854,0.558115) rgb(105cm)=(0.159194,0.482237,0.558073) rgb(106cm)=(0.157729,0.485932,0.558013) rgb(107cm)=(0.15627,0.489624,0.557936) rgb(108cm)=(0.154815,0.493313,0.55784) rgb(109cm)=(0.153364,0.497,0.557724) rgb(110cm)=(0.151918,0.500685,0.557587) rgb(111cm)=(0.150476,0.504369,0.55743) rgb(112cm)=(0.149039,0.508051,0.55725) rgb(113cm)=(0.147607,0.511733,0.557049) rgb(114cm)=(0.14618,0.515413,0.556823) rgb(115cm)=(0.144759,0.519093,0.556572) rgb(116cm)=(0.143343,0.522773,0.556295) rgb(117cm)=(0.141935,0.526453,0.555991) rgb(118cm)=(0.140536,0.530132,0.555659) rgb(119cm)=(0.139147,0.533812,0.555298) rgb(120cm)=(0.13777,0.537492,0.554906) rgb(121cm)=(0.136408,0.541173,0.554483) rgb(122cm)=(0.135066,0.544853,0.554029) rgb(123cm)=(0.133743,0.548535,0.553541) rgb(124cm)=(0.132444,0.552216,0.553018) rgb(125cm)=(0.131172,0.555899,0.552459) rgb(126cm)=(0.129933,0.559582,0.551864) rgb(127cm)=(0.128729,0.563265,0.551229) rgb(128cm)=(0.127568,0.566949,0.550556) rgb(129cm)=(0.126453,0.570633,0.549841) rgb(130cm)=(0.125394,0.574318,0.549086) rgb(131cm)=(0.124395,0.578002,0.548287) rgb(132cm)=(0.123463,0.581687,0.547445) rgb(133cm)=(0.122606,0.585371,0.546557) rgb(134cm)=(0.121831,0.589055,0.545623) rgb(135cm)=(0.121148,0.592739,0.544641) rgb(136cm)=(0.120565,0.596422,0.543611) rgb(137cm)=(0.120092,0.600104,0.54253) rgb(138cm)=(0.119738,0.603785,0.5414) rgb(139cm)=(0.119512,0.607464,0.540218) rgb(140cm)=(0.119423,0.611141,0.538982) rgb(141cm)=(0.119483,0.614817,0.537692) rgb(142cm)=(0.119699,0.61849,0.536347) rgb(143cm)=(0.120081,0.622161,0.534946) rgb(144cm)=(0.120638,0.625828,0.533488) rgb(145cm)=(0.12138,0.629492,0.531973) rgb(146cm)=(0.122312,0.633153,0.530398) rgb(147cm)=(0.123444,0.636809,0.528763) rgb(148cm)=(0.12478,0.640461,0.527068) rgb(149cm)=(0.126326,0.644107,0.525311) rgb(150cm)=(0.128087,0.647749,0.523491) rgb(151cm)=(0.130067,0.651384,0.521608) rgb(152cm)=(0.132268,0.655014,0.519661) rgb(153cm)=(0.134692,0.658636,0.517649) rgb(154cm)=(0.137339,0.662252,0.515571) rgb(155cm)=(0.14021,0.665859,0.513427) rgb(156cm)=(0.143303,0.669459,0.511215) rgb(157cm)=(0.146616,0.67305,0.508936) rgb(158cm)=(0.150148,0.676631,0.506589) rgb(159cm)=(0.153894,0.680203,0.504172) rgb(160cm)=(0.157851,0.683765,0.501686) rgb(161cm)=(0.162016,0.687316,0.499129) rgb(162cm)=(0.166383,0.690856,0.496502) rgb(163cm)=(0.170948,0.694384,0.493803) rgb(164cm)=(0.175707,0.6979,0.491033) rgb(165cm)=(0.180653,0.701402,0.488189) rgb(166cm)=(0.185783,0.704891,0.485273) rgb(167cm)=(0.19109,0.708366,0.482284) rgb(168cm)=(0.196571,0.711827,0.479221) rgb(169cm)=(0.202219,0.715272,0.476084) rgb(170cm)=(0.20803,0.718701,0.472873) rgb(171cm)=(0.214,0.722114,0.469588) rgb(172cm)=(0.220124,0.725509,0.466226) rgb(173cm)=(0.226397,0.728888,0.462789) rgb(174cm)=(0.232815,0.732247,0.459277) rgb(175cm)=(0.239374,0.735588,0.455688) rgb(176cm)=(0.24607,0.73891,0.452024) rgb(177cm)=(0.252899,0.742211,0.448284) rgb(178cm)=(0.259857,0.745492,0.444467) rgb(179cm)=(0.266941,0.748751,0.440573) rgb(180cm)=(0.274149,0.751988,0.436601) rgb(181cm)=(0.281477,0.755203,0.432552) rgb(182cm)=(0.288921,0.758394,0.428426) rgb(183cm)=(0.296479,0.761561,0.424223) rgb(184cm)=(0.304148,0.764704,0.419943) rgb(185cm)=(0.311925,0.767822,0.415586) rgb(186cm)=(0.319809,0.770914,0.411152) rgb(187cm)=(0.327796,0.77398,0.40664) rgb(188cm)=(0.335885,0.777018,0.402049) rgb(189cm)=(0.344074,0.780029,0.397381) rgb(190cm)=(0.35236,0.783011,0.392636) rgb(191cm)=(0.360741,0.785964,0.387814) rgb(192cm)=(0.369214,0.788888,0.382914) rgb(193cm)=(0.377779,0.791781,0.377939) rgb(194cm)=(0.386433,0.794644,0.372886) rgb(195cm)=(0.395174,0.797475,0.367757) rgb(196cm)=(0.404001,0.800275,0.362552) rgb(197cm)=(0.412913,0.803041,0.357269) rgb(198cm)=(0.421908,0.805774,0.35191) rgb(199cm)=(0.430983,0.808473,0.346476) rgb(200cm)=(0.440137,0.811138,0.340967) rgb(201cm)=(0.449368,0.813768,0.335384) rgb(202cm)=(0.458674,0.816363,0.329727) rgb(203cm)=(0.468053,0.818921,0.323998) rgb(204cm)=(0.477504,0.821444,0.318195) rgb(205cm)=(0.487026,0.823929,0.312321) rgb(206cm)=(0.496615,0.826376,0.306377) rgb(207cm)=(0.506271,0.828786,0.300362) rgb(208cm)=(0.515992,0.831158,0.294279) rgb(209cm)=(0.525776,0.833491,0.288127) rgb(210cm)=(0.535621,0.835785,0.281908) rgb(211cm)=(0.545524,0.838039,0.275626) rgb(212cm)=(0.555484,0.840254,0.269281) rgb(213cm)=(0.565498,0.84243,0.262877) rgb(214cm)=(0.575563,0.844566,0.256415) rgb(215cm)=(0.585678,0.846661,0.249897) rgb(216cm)=(0.595839,0.848717,0.243329) rgb(217cm)=(0.606045,0.850733,0.236712) rgb(218cm)=(0.616293,0.852709,0.230052) rgb(219cm)=(0.626579,0.854645,0.223353) rgb(220cm)=(0.636902,0.856542,0.21662) rgb(221cm)=(0.647257,0.8584,0.209861) rgb(222cm)=(0.657642,0.860219,0.203082) rgb(223cm)=(0.668054,0.861999,0.196293) rgb(224cm)=(0.678489,0.863742,0.189503) rgb(225cm)=(0.688944,0.865448,0.182725) rgb(226cm)=(0.699415,0.867117,0.175971) rgb(227cm)=(0.709898,0.868751,0.169257) rgb(228cm)=(0.720391,0.87035,0.162603) rgb(229cm)=(0.730889,0.871916,0.156029) rgb(230cm)=(0.741388,0.873449,0.149561) rgb(231cm)=(0.751884,0.874951,0.143228) rgb(232cm)=(0.762373,0.876424,0.137064) rgb(233cm)=(0.772852,0.877868,0.131109) rgb(234cm)=(0.783315,0.879285,0.125405) rgb(235cm)=(0.79376,0.880678,0.120005) rgb(236cm)=(0.804182,0.882046,0.114965) rgb(237cm)=(0.814576,0.883393,0.110347) rgb(238cm)=(0.82494,0.88472,0.106217) rgb(239cm)=(0.83527,0.886029,0.102646) rgb(240cm)=(0.845561,0.887322,0.099702) rgb(241cm)=(0.85581,0.888601,0.097452) rgb(242cm)=(0.866013,0.889868,0.095953) rgb(243cm)=(0.876168,0.891125,0.09525) rgb(244cm)=(0.886271,0.892374,0.095374) rgb(245cm)=(0.89632,0.893616,0.096335) rgb(246cm)=(0.906311,0.894855,0.098125) rgb(247cm)=(0.916242,0.896091,0.100717) rgb(248cm)=(0.926106,0.89733,0.104071) rgb(249cm)=(0.935904,0.89857,0.108131) rgb(250cm)=(0.945636,0.899815,0.112838) rgb(251cm)=(0.9553,0.901065,0.118128) rgb(252cm)=(0.964894,0.902323,0.123941) rgb(253cm)=(0.974417,0.90359,0.130215) rgb(254cm)=(0.983868,0.904867,0.136897) rgb(255cm)=(0.993248,0.906157,0.143936) }, colorbar]\addplot [point meta min=0.0, point meta max=1.0] graphics [xmin=0.0, xmax=40.0, ymin=-8000, ymax=8000] {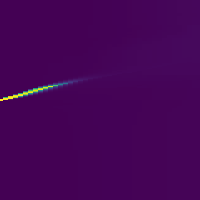};
    \end{groupplot}
    
    \end{tikzpicture}
    \caption{State space partition (same color scheme as \cref{fig:adap_comparison}) and overapproximated probability of NMAC with both the worst-case transition splitting heuristic and the policy overapproximation splitting heuristic. The intruder vertical rate is fixed at $-90$ ft/s, the ownship vertical rate is 0 ft/s, and the previous advisory is $\textsc{coc}$. \label{fig:all}}
\end{figure}

\Cref{fig:all} shows the results when the policy overapproximation heuristic is used in addition to the worst-case transition heuristic.
Beyond the splits due to the worst-case transition heuristic, this heuristic results in splits in overapproximated regions at higher values of $\tau$. There is no significant difference in the visualization of probabilities between \cref{fig:dynamics} and \cref{fig:all}, and the maximum probability of NMAC at $\tau = 40$ seconds drops from 0.0305 to 0.0268. The addition of this splitting heuristic did not add a significant benefit over the worst-case transition heuristic, possibly due to the fact that adaptive verification already addresses policy overapproximation error.

\Cref{tab:error_reduc_summary} summarizes the effect of each overapproximation error reduction technique on the maximum overapproximated probability of NMAC for cells at $\tau = 40$ seconds. We also provide the number of cells in the final partitioning and the time required to perform the model checking to illustrate the added complexity of each technique. All time trials for this experiment were performed on 5 cores of a 4.20 GHz Intel Core i7 processor.

\begin{table}[htb]
    \caption{Effect of error reduction methods on overapproximated probability of NMAC \label{tab:error_reduc_summary}}
    \begin{tabular}{@{}llllll@{}}
         \toprule
         \textbf{Error Reduction} & \makecell[cl]{\textbf{Transition} \\ \textbf{Threshold}} & \makecell[cl]{\textbf{Action} \\ \textbf{Threshold}} & \textbf{Cells} & \textbf{Time (s)} & \textbf{Probability}\\
         \midrule
         Adaptive verification & - & - & \num{3.4e5} & 157 & 1.0 \\
         Worst-case transition & 0.5 & - & \num{6.9e5} & 218 & 1.0  \\
         Worst-case transition & 0.1 & - & \num{2.3e6} & 565 & 0.372  \\
         Worst-case transition & 0.005 & - & \num{8.9e6} & 1546 & 0.0305  \\
         All techniques & 0.005 & 0.005 & \num{1.2e7} & 6336 & 0.0268 \\
         \bottomrule
    \end{tabular}
\end{table}

The adaptive verification technique alone is not enough to reduce overapproximation error to obtain a meaningful estimate of the probability of NMAC. Adaptive verification addresses only policy overapproximation error but does nothing to reduce worst-case transition error. When we add the online splitting heuristic to address worst-case transition error and decrease the splitting threshold, the probability estimate decreases to a more meaningful result. We can guarantee that the probability of NMAC is less than 0.0305. Adding the online splitting heuristic to further reduce policy overapproximation in key areas of the state space does not have as much of an effect on the resulting probability. Nevertheless, it changes the guarantee to a 0.0268 probability of NMAC.

While the overapproximation error in the probability estimate decreases when we apply the online splitting heuristics, the number of cells in the final partition and time required to perform the model checking increases. \Cref{tab:error_reduc_summary} therefore shows a tradeoff between the error in the final estimate and the overall complexity of the algorithm. However, the online splitting heuristics still allow us to exploit the structure of the problem to ensure that splits only occur in critical parts of the state space. If we were to instead na\"ively partition the state space uniformly into cells of the minimum cell size, our final partition would have \num{2.36e8} cells. Comparing to the results in \cref{tab:error_reduc_summary}, even the partition that uses all overapproximation error reduction techniques has less than 5\% of the cells required for a uniform partition.

\subsection{Full Scale Model}
After gaining intuition using the two-dimensional model, we applied our method to determine the maximum probability of collision on the full scale model. For computational reasons, the action range threshold was slowly increased throughout the solving process. To analyze the quality of our probability estimates, we generate two baseline probability comparisons. The first comparison is the estimated probability of NMAC when using the large numeric lookup table that the neural network is meant to approximate. This probability is calculated by performing traditional MDP model checking on the table policy using the method outlined in \cref{sec:model_check} with the same model used to evaluate the neural network. We use multilinear interpolation to determine probabilities at points in the state space that do not correspond directly to table entries, so the result is not guaranteed to be an overapproximation. We also compare with the probability of NMAC detected through 1,000 Monte Carlo simulations from various points in the state space. While Monte Carlo simulations cannot provide any formal guarantees, they provide a good approximation to the probability of NMAC. 

\begin{figure}[htb]
    \begin{tikzpicture}
    \begin{groupplot}[group style={horizontal sep = 0.4cm, vertical sep = 2.5cm, group size=3 by 1}]
    \nextgroupplot [height = {5cm}, ylabel = {$h$ (ft)}, title = {Lookup Table}, xmin = {0.0}, xmax = {40.0}, ymax = {1000}, xlabel = {$\tau$ (s)}, ymin = {-1000}, width = {4.35cm}, enlargelimits = false, axis on top]\addplot [point meta min=0, point meta max=3] graphics [xmin=0.0, xmax=40.0, ymin=-1000, ymax=1000] {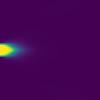};
    \nextgroupplot [height = {5cm}, title = {Monte Carlo}, xmin = {0.0}, xmax = {40.0}, ymax = {1000}, xlabel = {$\tau$ (s)}, ymin = {-1000}, width = {4.35cm}, enlargelimits = false, axis on top, yticklabels={,,}]\addplot [point meta min=0, point meta max=3] graphics [xmin=0.0, xmax=40.0, ymin=-1000, ymax=1000] {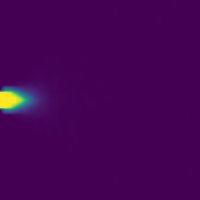};
    \nextgroupplot [height = {5cm}, title = {Model Checking}, xmin = {0.0}, xmax = {40.0}, ymax = {1000}, xlabel = {$\tau$ (s)}, ymin = {-1000}, width = {4.35cm}, enlargelimits = false, axis on top, yticklabels={,,}, colorbar style={
                width=0.3cm}, colormap={mycolormap}{ rgb(0cm)=(0.267004,0.004874,0.329415) rgb(1cm)=(0.26851,0.009605,0.335427) rgb(2cm)=(0.269944,0.014625,0.341379) rgb(3cm)=(0.271305,0.019942,0.347269) rgb(4cm)=(0.272594,0.025563,0.353093) rgb(5cm)=(0.273809,0.031497,0.358853) rgb(6cm)=(0.274952,0.037752,0.364543) rgb(7cm)=(0.276022,0.044167,0.370164) rgb(8cm)=(0.277018,0.050344,0.375715) rgb(9cm)=(0.277941,0.056324,0.381191) rgb(10cm)=(0.278791,0.062145,0.386592) rgb(11cm)=(0.279566,0.067836,0.391917) rgb(12cm)=(0.280267,0.073417,0.397163) rgb(13cm)=(0.280894,0.078907,0.402329) rgb(14cm)=(0.281446,0.08432,0.407414) rgb(15cm)=(0.281924,0.089666,0.412415) rgb(16cm)=(0.282327,0.094955,0.417331) rgb(17cm)=(0.282656,0.100196,0.42216) rgb(18cm)=(0.28291,0.105393,0.426902) rgb(19cm)=(0.283091,0.110553,0.431554) rgb(20cm)=(0.283197,0.11568,0.436115) rgb(21cm)=(0.283229,0.120777,0.440584) rgb(22cm)=(0.283187,0.125848,0.44496) rgb(23cm)=(0.283072,0.130895,0.449241) rgb(24cm)=(0.282884,0.13592,0.453427) rgb(25cm)=(0.282623,0.140926,0.457517) rgb(26cm)=(0.28229,0.145912,0.46151) rgb(27cm)=(0.281887,0.150881,0.465405) rgb(28cm)=(0.281412,0.155834,0.469201) rgb(29cm)=(0.280868,0.160771,0.472899) rgb(30cm)=(0.280255,0.165693,0.476498) rgb(31cm)=(0.279574,0.170599,0.479997) rgb(32cm)=(0.278826,0.17549,0.483397) rgb(33cm)=(0.278012,0.180367,0.486697) rgb(34cm)=(0.277134,0.185228,0.489898) rgb(35cm)=(0.276194,0.190074,0.493001) rgb(36cm)=(0.275191,0.194905,0.496005) rgb(37cm)=(0.274128,0.199721,0.498911) rgb(38cm)=(0.273006,0.20452,0.501721) rgb(39cm)=(0.271828,0.209303,0.504434) rgb(40cm)=(0.270595,0.214069,0.507052) rgb(41cm)=(0.269308,0.218818,0.509577) rgb(42cm)=(0.267968,0.223549,0.512008) rgb(43cm)=(0.26658,0.228262,0.514349) rgb(44cm)=(0.265145,0.232956,0.516599) rgb(45cm)=(0.263663,0.237631,0.518762) rgb(46cm)=(0.262138,0.242286,0.520837) rgb(47cm)=(0.260571,0.246922,0.522828) rgb(48cm)=(0.258965,0.251537,0.524736) rgb(49cm)=(0.257322,0.25613,0.526563) rgb(50cm)=(0.255645,0.260703,0.528312) rgb(51cm)=(0.253935,0.265254,0.529983) rgb(52cm)=(0.252194,0.269783,0.531579) rgb(53cm)=(0.250425,0.27429,0.533103) rgb(54cm)=(0.248629,0.278775,0.534556) rgb(55cm)=(0.246811,0.283237,0.535941) rgb(56cm)=(0.244972,0.287675,0.53726) rgb(57cm)=(0.243113,0.292092,0.538516) rgb(58cm)=(0.241237,0.296485,0.539709) rgb(59cm)=(0.239346,0.300855,0.540844) rgb(60cm)=(0.237441,0.305202,0.541921) rgb(61cm)=(0.235526,0.309527,0.542944) rgb(62cm)=(0.233603,0.313828,0.543914) rgb(63cm)=(0.231674,0.318106,0.544834) rgb(64cm)=(0.229739,0.322361,0.545706) rgb(65cm)=(0.227802,0.326594,0.546532) rgb(66cm)=(0.225863,0.330805,0.547314) rgb(67cm)=(0.223925,0.334994,0.548053) rgb(68cm)=(0.221989,0.339161,0.548752) rgb(69cm)=(0.220057,0.343307,0.549413) rgb(70cm)=(0.21813,0.347432,0.550038) rgb(71cm)=(0.21621,0.351535,0.550627) rgb(72cm)=(0.214298,0.355619,0.551184) rgb(73cm)=(0.212395,0.359683,0.55171) rgb(74cm)=(0.210503,0.363727,0.552206) rgb(75cm)=(0.208623,0.367752,0.552675) rgb(76cm)=(0.206756,0.371758,0.553117) rgb(77cm)=(0.204903,0.375746,0.553533) rgb(78cm)=(0.203063,0.379716,0.553925) rgb(79cm)=(0.201239,0.38367,0.554294) rgb(80cm)=(0.19943,0.387607,0.554642) rgb(81cm)=(0.197636,0.391528,0.554969) rgb(82cm)=(0.19586,0.395433,0.555276) rgb(83cm)=(0.1941,0.399323,0.555565) rgb(84cm)=(0.192357,0.403199,0.555836) rgb(85cm)=(0.190631,0.407061,0.556089) rgb(86cm)=(0.188923,0.41091,0.556326) rgb(87cm)=(0.187231,0.414746,0.556547) rgb(88cm)=(0.185556,0.41857,0.556753) rgb(89cm)=(0.183898,0.422383,0.556944) rgb(90cm)=(0.182256,0.426184,0.55712) rgb(91cm)=(0.180629,0.429975,0.557282) rgb(92cm)=(0.179019,0.433756,0.55743) rgb(93cm)=(0.177423,0.437527,0.557565) rgb(94cm)=(0.175841,0.44129,0.557685) rgb(95cm)=(0.174274,0.445044,0.557792) rgb(96cm)=(0.172719,0.448791,0.557885) rgb(97cm)=(0.171176,0.45253,0.557965) rgb(98cm)=(0.169646,0.456262,0.55803) rgb(99cm)=(0.168126,0.459988,0.558082) rgb(100cm)=(0.166617,0.463708,0.558119) rgb(101cm)=(0.165117,0.467423,0.558141) rgb(102cm)=(0.163625,0.471133,0.558148) rgb(103cm)=(0.162142,0.474838,0.55814) rgb(104cm)=(0.160665,0.47854,0.558115) rgb(105cm)=(0.159194,0.482237,0.558073) rgb(106cm)=(0.157729,0.485932,0.558013) rgb(107cm)=(0.15627,0.489624,0.557936) rgb(108cm)=(0.154815,0.493313,0.55784) rgb(109cm)=(0.153364,0.497,0.557724) rgb(110cm)=(0.151918,0.500685,0.557587) rgb(111cm)=(0.150476,0.504369,0.55743) rgb(112cm)=(0.149039,0.508051,0.55725) rgb(113cm)=(0.147607,0.511733,0.557049) rgb(114cm)=(0.14618,0.515413,0.556823) rgb(115cm)=(0.144759,0.519093,0.556572) rgb(116cm)=(0.143343,0.522773,0.556295) rgb(117cm)=(0.141935,0.526453,0.555991) rgb(118cm)=(0.140536,0.530132,0.555659) rgb(119cm)=(0.139147,0.533812,0.555298) rgb(120cm)=(0.13777,0.537492,0.554906) rgb(121cm)=(0.136408,0.541173,0.554483) rgb(122cm)=(0.135066,0.544853,0.554029) rgb(123cm)=(0.133743,0.548535,0.553541) rgb(124cm)=(0.132444,0.552216,0.553018) rgb(125cm)=(0.131172,0.555899,0.552459) rgb(126cm)=(0.129933,0.559582,0.551864) rgb(127cm)=(0.128729,0.563265,0.551229) rgb(128cm)=(0.127568,0.566949,0.550556) rgb(129cm)=(0.126453,0.570633,0.549841) rgb(130cm)=(0.125394,0.574318,0.549086) rgb(131cm)=(0.124395,0.578002,0.548287) rgb(132cm)=(0.123463,0.581687,0.547445) rgb(133cm)=(0.122606,0.585371,0.546557) rgb(134cm)=(0.121831,0.589055,0.545623) rgb(135cm)=(0.121148,0.592739,0.544641) rgb(136cm)=(0.120565,0.596422,0.543611) rgb(137cm)=(0.120092,0.600104,0.54253) rgb(138cm)=(0.119738,0.603785,0.5414) rgb(139cm)=(0.119512,0.607464,0.540218) rgb(140cm)=(0.119423,0.611141,0.538982) rgb(141cm)=(0.119483,0.614817,0.537692) rgb(142cm)=(0.119699,0.61849,0.536347) rgb(143cm)=(0.120081,0.622161,0.534946) rgb(144cm)=(0.120638,0.625828,0.533488) rgb(145cm)=(0.12138,0.629492,0.531973) rgb(146cm)=(0.122312,0.633153,0.530398) rgb(147cm)=(0.123444,0.636809,0.528763) rgb(148cm)=(0.12478,0.640461,0.527068) rgb(149cm)=(0.126326,0.644107,0.525311) rgb(150cm)=(0.128087,0.647749,0.523491) rgb(151cm)=(0.130067,0.651384,0.521608) rgb(152cm)=(0.132268,0.655014,0.519661) rgb(153cm)=(0.134692,0.658636,0.517649) rgb(154cm)=(0.137339,0.662252,0.515571) rgb(155cm)=(0.14021,0.665859,0.513427) rgb(156cm)=(0.143303,0.669459,0.511215) rgb(157cm)=(0.146616,0.67305,0.508936) rgb(158cm)=(0.150148,0.676631,0.506589) rgb(159cm)=(0.153894,0.680203,0.504172) rgb(160cm)=(0.157851,0.683765,0.501686) rgb(161cm)=(0.162016,0.687316,0.499129) rgb(162cm)=(0.166383,0.690856,0.496502) rgb(163cm)=(0.170948,0.694384,0.493803) rgb(164cm)=(0.175707,0.6979,0.491033) rgb(165cm)=(0.180653,0.701402,0.488189) rgb(166cm)=(0.185783,0.704891,0.485273) rgb(167cm)=(0.19109,0.708366,0.482284) rgb(168cm)=(0.196571,0.711827,0.479221) rgb(169cm)=(0.202219,0.715272,0.476084) rgb(170cm)=(0.20803,0.718701,0.472873) rgb(171cm)=(0.214,0.722114,0.469588) rgb(172cm)=(0.220124,0.725509,0.466226) rgb(173cm)=(0.226397,0.728888,0.462789) rgb(174cm)=(0.232815,0.732247,0.459277) rgb(175cm)=(0.239374,0.735588,0.455688) rgb(176cm)=(0.24607,0.73891,0.452024) rgb(177cm)=(0.252899,0.742211,0.448284) rgb(178cm)=(0.259857,0.745492,0.444467) rgb(179cm)=(0.266941,0.748751,0.440573) rgb(180cm)=(0.274149,0.751988,0.436601) rgb(181cm)=(0.281477,0.755203,0.432552) rgb(182cm)=(0.288921,0.758394,0.428426) rgb(183cm)=(0.296479,0.761561,0.424223) rgb(184cm)=(0.304148,0.764704,0.419943) rgb(185cm)=(0.311925,0.767822,0.415586) rgb(186cm)=(0.319809,0.770914,0.411152) rgb(187cm)=(0.327796,0.77398,0.40664) rgb(188cm)=(0.335885,0.777018,0.402049) rgb(189cm)=(0.344074,0.780029,0.397381) rgb(190cm)=(0.35236,0.783011,0.392636) rgb(191cm)=(0.360741,0.785964,0.387814) rgb(192cm)=(0.369214,0.788888,0.382914) rgb(193cm)=(0.377779,0.791781,0.377939) rgb(194cm)=(0.386433,0.794644,0.372886) rgb(195cm)=(0.395174,0.797475,0.367757) rgb(196cm)=(0.404001,0.800275,0.362552) rgb(197cm)=(0.412913,0.803041,0.357269) rgb(198cm)=(0.421908,0.805774,0.35191) rgb(199cm)=(0.430983,0.808473,0.346476) rgb(200cm)=(0.440137,0.811138,0.340967) rgb(201cm)=(0.449368,0.813768,0.335384) rgb(202cm)=(0.458674,0.816363,0.329727) rgb(203cm)=(0.468053,0.818921,0.323998) rgb(204cm)=(0.477504,0.821444,0.318195) rgb(205cm)=(0.487026,0.823929,0.312321) rgb(206cm)=(0.496615,0.826376,0.306377) rgb(207cm)=(0.506271,0.828786,0.300362) rgb(208cm)=(0.515992,0.831158,0.294279) rgb(209cm)=(0.525776,0.833491,0.288127) rgb(210cm)=(0.535621,0.835785,0.281908) rgb(211cm)=(0.545524,0.838039,0.275626) rgb(212cm)=(0.555484,0.840254,0.269281) rgb(213cm)=(0.565498,0.84243,0.262877) rgb(214cm)=(0.575563,0.844566,0.256415) rgb(215cm)=(0.585678,0.846661,0.249897) rgb(216cm)=(0.595839,0.848717,0.243329) rgb(217cm)=(0.606045,0.850733,0.236712) rgb(218cm)=(0.616293,0.852709,0.230052) rgb(219cm)=(0.626579,0.854645,0.223353) rgb(220cm)=(0.636902,0.856542,0.21662) rgb(221cm)=(0.647257,0.8584,0.209861) rgb(222cm)=(0.657642,0.860219,0.203082) rgb(223cm)=(0.668054,0.861999,0.196293) rgb(224cm)=(0.678489,0.863742,0.189503) rgb(225cm)=(0.688944,0.865448,0.182725) rgb(226cm)=(0.699415,0.867117,0.175971) rgb(227cm)=(0.709898,0.868751,0.169257) rgb(228cm)=(0.720391,0.87035,0.162603) rgb(229cm)=(0.730889,0.871916,0.156029) rgb(230cm)=(0.741388,0.873449,0.149561) rgb(231cm)=(0.751884,0.874951,0.143228) rgb(232cm)=(0.762373,0.876424,0.137064) rgb(233cm)=(0.772852,0.877868,0.131109) rgb(234cm)=(0.783315,0.879285,0.125405) rgb(235cm)=(0.79376,0.880678,0.120005) rgb(236cm)=(0.804182,0.882046,0.114965) rgb(237cm)=(0.814576,0.883393,0.110347) rgb(238cm)=(0.82494,0.88472,0.106217) rgb(239cm)=(0.83527,0.886029,0.102646) rgb(240cm)=(0.845561,0.887322,0.099702) rgb(241cm)=(0.85581,0.888601,0.097452) rgb(242cm)=(0.866013,0.889868,0.095953) rgb(243cm)=(0.876168,0.891125,0.09525) rgb(244cm)=(0.886271,0.892374,0.095374) rgb(245cm)=(0.89632,0.893616,0.096335) rgb(246cm)=(0.906311,0.894855,0.098125) rgb(247cm)=(0.916242,0.896091,0.100717) rgb(248cm)=(0.926106,0.89733,0.104071) rgb(249cm)=(0.935904,0.89857,0.108131) rgb(250cm)=(0.945636,0.899815,0.112838) rgb(251cm)=(0.9553,0.901065,0.118128) rgb(252cm)=(0.964894,0.902323,0.123941) rgb(253cm)=(0.974417,0.90359,0.130215) rgb(254cm)=(0.983868,0.904867,0.136897) rgb(255cm)=(0.993248,0.906157,0.143936) }, colorbar]\addplot [point meta min=0.0, point meta max=1.0] graphics [xmin=0.0, xmax=40.0, ymin=-1000, ymax=1000] {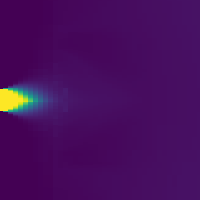};
    \end{groupplot}
    
    \end{tikzpicture}
    \caption{Comparison of lookup table probabilities of NMAC estimated using traditional MDP model checking, neural network probabilities of NMAC estimated using Monte Carlo simulations, and overapproximated neural network probabilities of NMAC using the model checking formulation in this work. The ownship and intruder vertical rates are both 0 ft/s, and the previous advisory is $\textsc{coc}$. \label{fig:monte_carlo_comp}}
\end{figure}

The results for one slice of the state space are shown in \cref{fig:monte_carlo_comp}.
All three plots show similar trends. The model checking outputs for the table and neural network are similar with slightly higher probabilities for the neural network. The probabilities are also similar to the Monte Carlo probabilities with the same cells showing high probability of collision; however, the region of high probability is slightly larger in the model checking results. It is clear that the model checking probabilities represent an overapproximation.

\begin{figure}[htb]
    \begin{tikzpicture}
    \begin{axis}[height = {6cm}, ylabel = {Probability of NMAC}, xlabel = {$\tau$ (s)}, width = {7cm}, xmin = {0.0}, xmax = {40.0}]\addplot+ [mark = {none}, style={thick, black}]coordinates {
    (0, 1.0)
    (1, 1.000000003259629)
    (2, 1.0000000012722456)
    (3, 0.9999840003904046)
    (4, 0.9711354560321244)
    (5, 0.8102937846246531)
    (6, 0.5866548562103804)
    (7, 0.40452021163894186)
    (8, 0.2755899324161582)
    (9, 0.18574869193502688)
    (10, 0.1253973366226162)
    (11, 0.08453738915862637)
    (12, 0.05712172288377555)
    (13, 0.03851713299876188)
    (14, 0.026187843147882806)
    (15, 0.01801539638121246)
    (16, 0.012586574866307215)
    (17, 0.008787238712389971)
    (18, 0.006249618452299826)
    (19, 0.004568962681229616)
    (20, 0.0033746387902389616)
    (21, 0.002495831305374725)
    (22, 0.0018869168717364367)
    (23, 0.0014824401781075055)
    (24, 0.0014248234215460589)
    (25, 0.0013642427563419577)
    (26, 0.0013501275047024758)
    (27, 0.0013236614538639346)
    (28, 0.0012869133659599511)
    (29, 0.0012421335140934287)
    (30, 0.0011917026754584858)
    (31, 0.0011505479765193149)
    (32, 0.0011210778542779694)
    (33, 0.0010846569464477368)
    (34, 0.0010433327585261927)
    (35, 0.0009989622094593166)
    (36, 0.0009530831145992647)
    (37, 0.0009069373346373182)
    (38, 0.0008614505526743553)
    (39, 0.0008172721496309245)
    (40, 0.0007748369266731784)
    };
    \addplot+ [mark = {none}, style={thick, black!60}]coordinates {
    (0, 1.0)
    (1, 1.0)
    (2, 1.0)
    (3, 1.0)
    (4, 0.936)
    (5, 0.666)
    (6, 0.342)
    (7, 0.209)
    (8, 0.109)
    (9, 0.054)
    (10, 0.029)
    (11, 0.02)
    (12, 0.015)
    (13, 0.008)
    (14, 0.005)
    (15, 0.006)
    (16, 0.003)
    (17, 0.003)
    (18, 0.003)
    (19, 0.002)
    (20, 0.002)
    (21, 0.002)
    (22, 0.001)
    (23, 0.002)
    (24, 0.001)
    (25, 0.001)
    (26, 0.001)
    (27, 0.001)
    (28, 0.001)
    (29, 0.001)
    (30, 0.0)
    (31, 0.001)
    (32, 0.002)
    (33, 0.001)
    (34, 0.001)
    (35, 0.002)
    (36, 0.0)
    (37, 0.002)
    (38, 0.002)
    (39, 0.001)
    (40, 0.001)
    };
    \addplot+ [mark = {none}, style={thick,red}]coordinates {
    (0, 1.0)
    (1, 1.0)
    (2, 1.0)
    (3, 1.0)
    (4, 1.0)
    (5, 0.9614775196195985)
    (6, 0.7802454376058908)
    (7, 0.5483628522371836)
    (8, 0.3757669938001884)
    (9, 0.24579752069221691)
    (10, 0.15840823779199323)
    (11, 0.1154248711060435)
    (12, 0.08754488067783564)
    (13, 0.0747702473569079)
    (14, 0.05897235072962319)
    (15, 0.05322369971759498)
    (16, 0.04919927373925621)
    (17, 0.046452811449264865)
    (18, 0.04384899026734663)
    (19, 0.04280022449952902)
    (20, 0.04134249727471106)
    (21, 0.040585851187600414)
    (22, 0.04002175206126235)
    (23, 0.03926973715316563)
    (24, 0.03816597338507771)
    (25, 0.036132484428898624)
    (26, 0.03424517303088939)
    (27, 0.03300453235380484)
    (28, 0.03378501764857921)
    (29, 0.034797168781518305)
    (30, 0.0361414565372205)
    (31, 0.03729542077109216)
    (32, 0.03854924744006209)
    (33, 0.03922904742017115)
    (34, 0.040552770719066225)
    (35, 0.041627756763435705)
    (36, 0.042551957158505106)
    (37, 0.043909184766302456)
    (38, 0.0450006085264426)
    (39, 0.0466485918350328)
    (40, 0.04732218579963933)
    };
    \legend{Lookup Table, Monte Carlo, Model Checking}
    \end{axis}
    
    \end{tikzpicture}
    \caption{Probability of NMAC when previous advisory is $\textsc{coc}$ as time to loss of separation increases. The ownship and intruder vertical rates are fixed at 0 ft/s. \label{fig:tau_probs}}
\end{figure}
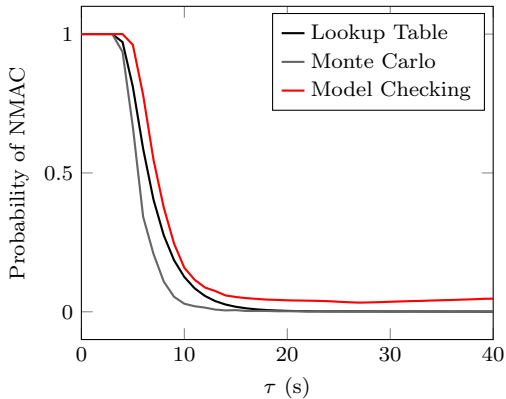

\Cref{fig:tau_probs} shows the maximum probability of collision among all cells as $\tau$ is increased for the same slice of the state space shown in \cref{fig:monte_carlo_comp}.
At $\tau = 0$ seconds, the maximum probability of NMAC is one, corresponding to cells in the NMAC region. As $\tau$ increases for each method, there is a sharp dropoff in probability of NMAC. The overapproximated model checking probabilities are greater than or equal to both the lookup table and Monte Carlo probabilities at all points. The curve, however, remains close to the other curves and represents a tight overapproximation. Furthermore, unlike the Monte Carlo estimate and lookup table model checking estimate, the neural network model checking estimate represents a guarantee on the performance of the neural network policy. The final model checking probability is 0.0473 at $\tau = 40$ seconds. Therefore, with respect to the stochastic dynamics model in \cref{sec:CAS}, we can guarantee that the probability of NMAC is less than 0.0473 when the intruder and ownship are in level flight.

\begin{figure}[htb]
    \begin{tikzpicture}
    \begin{groupplot}[group style={horizontal sep = 0.4cm, vertical sep = 2.5cm, group size=3 by 1}]
    \nextgroupplot [height = {5cm}, ylabel = {$h$ (ft)}, title = {Lookup Table}, xmin = {0.0}, xmax = {40.0}, ymax = {4000}, xlabel = {$\tau$ (s)}, ymin = {-4000}, width = {4.35cm}, enlargelimits = false, axis on top]\addplot [point meta min=0, point meta max=3] graphics [xmin=0.0, xmax=40.0, ymin=-4000, ymax=4000] {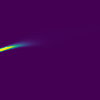};
    \nextgroupplot [height = {5cm}, title = {Monte Carlo}, xmin = {0.0}, xmax = {40.0}, ymax = {4000}, xlabel = {$\tau$ (s)}, ymin = {-4000}, width = {4.35cm}, enlargelimits = false, axis on top, yticklabels={,,}]\addplot [point meta min=0, point meta max=3] graphics [xmin=0.0, xmax=40.0, ymin=-4000, ymax=4000] {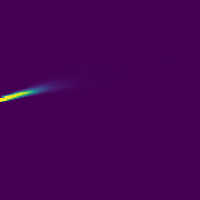};
    \nextgroupplot [height = {5cm}, title = {Model Checking}, xmin = {0.0}, xmax = {40.0}, ymax = {4000}, xlabel = {$\tau$ (s)}, ymin = {-4000}, width = {4.35cm}, enlargelimits = false, axis on top, yticklabels={,,}, colorbar style={
                width=0.3cm}, colormap={mycolormap}{ rgb(0cm)=(0.267004,0.004874,0.329415) rgb(1cm)=(0.26851,0.009605,0.335427) rgb(2cm)=(0.269944,0.014625,0.341379) rgb(3cm)=(0.271305,0.019942,0.347269) rgb(4cm)=(0.272594,0.025563,0.353093) rgb(5cm)=(0.273809,0.031497,0.358853) rgb(6cm)=(0.274952,0.037752,0.364543) rgb(7cm)=(0.276022,0.044167,0.370164) rgb(8cm)=(0.277018,0.050344,0.375715) rgb(9cm)=(0.277941,0.056324,0.381191) rgb(10cm)=(0.278791,0.062145,0.386592) rgb(11cm)=(0.279566,0.067836,0.391917) rgb(12cm)=(0.280267,0.073417,0.397163) rgb(13cm)=(0.280894,0.078907,0.402329) rgb(14cm)=(0.281446,0.08432,0.407414) rgb(15cm)=(0.281924,0.089666,0.412415) rgb(16cm)=(0.282327,0.094955,0.417331) rgb(17cm)=(0.282656,0.100196,0.42216) rgb(18cm)=(0.28291,0.105393,0.426902) rgb(19cm)=(0.283091,0.110553,0.431554) rgb(20cm)=(0.283197,0.11568,0.436115) rgb(21cm)=(0.283229,0.120777,0.440584) rgb(22cm)=(0.283187,0.125848,0.44496) rgb(23cm)=(0.283072,0.130895,0.449241) rgb(24cm)=(0.282884,0.13592,0.453427) rgb(25cm)=(0.282623,0.140926,0.457517) rgb(26cm)=(0.28229,0.145912,0.46151) rgb(27cm)=(0.281887,0.150881,0.465405) rgb(28cm)=(0.281412,0.155834,0.469201) rgb(29cm)=(0.280868,0.160771,0.472899) rgb(30cm)=(0.280255,0.165693,0.476498) rgb(31cm)=(0.279574,0.170599,0.479997) rgb(32cm)=(0.278826,0.17549,0.483397) rgb(33cm)=(0.278012,0.180367,0.486697) rgb(34cm)=(0.277134,0.185228,0.489898) rgb(35cm)=(0.276194,0.190074,0.493001) rgb(36cm)=(0.275191,0.194905,0.496005) rgb(37cm)=(0.274128,0.199721,0.498911) rgb(38cm)=(0.273006,0.20452,0.501721) rgb(39cm)=(0.271828,0.209303,0.504434) rgb(40cm)=(0.270595,0.214069,0.507052) rgb(41cm)=(0.269308,0.218818,0.509577) rgb(42cm)=(0.267968,0.223549,0.512008) rgb(43cm)=(0.26658,0.228262,0.514349) rgb(44cm)=(0.265145,0.232956,0.516599) rgb(45cm)=(0.263663,0.237631,0.518762) rgb(46cm)=(0.262138,0.242286,0.520837) rgb(47cm)=(0.260571,0.246922,0.522828) rgb(48cm)=(0.258965,0.251537,0.524736) rgb(49cm)=(0.257322,0.25613,0.526563) rgb(50cm)=(0.255645,0.260703,0.528312) rgb(51cm)=(0.253935,0.265254,0.529983) rgb(52cm)=(0.252194,0.269783,0.531579) rgb(53cm)=(0.250425,0.27429,0.533103) rgb(54cm)=(0.248629,0.278775,0.534556) rgb(55cm)=(0.246811,0.283237,0.535941) rgb(56cm)=(0.244972,0.287675,0.53726) rgb(57cm)=(0.243113,0.292092,0.538516) rgb(58cm)=(0.241237,0.296485,0.539709) rgb(59cm)=(0.239346,0.300855,0.540844) rgb(60cm)=(0.237441,0.305202,0.541921) rgb(61cm)=(0.235526,0.309527,0.542944) rgb(62cm)=(0.233603,0.313828,0.543914) rgb(63cm)=(0.231674,0.318106,0.544834) rgb(64cm)=(0.229739,0.322361,0.545706) rgb(65cm)=(0.227802,0.326594,0.546532) rgb(66cm)=(0.225863,0.330805,0.547314) rgb(67cm)=(0.223925,0.334994,0.548053) rgb(68cm)=(0.221989,0.339161,0.548752) rgb(69cm)=(0.220057,0.343307,0.549413) rgb(70cm)=(0.21813,0.347432,0.550038) rgb(71cm)=(0.21621,0.351535,0.550627) rgb(72cm)=(0.214298,0.355619,0.551184) rgb(73cm)=(0.212395,0.359683,0.55171) rgb(74cm)=(0.210503,0.363727,0.552206) rgb(75cm)=(0.208623,0.367752,0.552675) rgb(76cm)=(0.206756,0.371758,0.553117) rgb(77cm)=(0.204903,0.375746,0.553533) rgb(78cm)=(0.203063,0.379716,0.553925) rgb(79cm)=(0.201239,0.38367,0.554294) rgb(80cm)=(0.19943,0.387607,0.554642) rgb(81cm)=(0.197636,0.391528,0.554969) rgb(82cm)=(0.19586,0.395433,0.555276) rgb(83cm)=(0.1941,0.399323,0.555565) rgb(84cm)=(0.192357,0.403199,0.555836) rgb(85cm)=(0.190631,0.407061,0.556089) rgb(86cm)=(0.188923,0.41091,0.556326) rgb(87cm)=(0.187231,0.414746,0.556547) rgb(88cm)=(0.185556,0.41857,0.556753) rgb(89cm)=(0.183898,0.422383,0.556944) rgb(90cm)=(0.182256,0.426184,0.55712) rgb(91cm)=(0.180629,0.429975,0.557282) rgb(92cm)=(0.179019,0.433756,0.55743) rgb(93cm)=(0.177423,0.437527,0.557565) rgb(94cm)=(0.175841,0.44129,0.557685) rgb(95cm)=(0.174274,0.445044,0.557792) rgb(96cm)=(0.172719,0.448791,0.557885) rgb(97cm)=(0.171176,0.45253,0.557965) rgb(98cm)=(0.169646,0.456262,0.55803) rgb(99cm)=(0.168126,0.459988,0.558082) rgb(100cm)=(0.166617,0.463708,0.558119) rgb(101cm)=(0.165117,0.467423,0.558141) rgb(102cm)=(0.163625,0.471133,0.558148) rgb(103cm)=(0.162142,0.474838,0.55814) rgb(104cm)=(0.160665,0.47854,0.558115) rgb(105cm)=(0.159194,0.482237,0.558073) rgb(106cm)=(0.157729,0.485932,0.558013) rgb(107cm)=(0.15627,0.489624,0.557936) rgb(108cm)=(0.154815,0.493313,0.55784) rgb(109cm)=(0.153364,0.497,0.557724) rgb(110cm)=(0.151918,0.500685,0.557587) rgb(111cm)=(0.150476,0.504369,0.55743) rgb(112cm)=(0.149039,0.508051,0.55725) rgb(113cm)=(0.147607,0.511733,0.557049) rgb(114cm)=(0.14618,0.515413,0.556823) rgb(115cm)=(0.144759,0.519093,0.556572) rgb(116cm)=(0.143343,0.522773,0.556295) rgb(117cm)=(0.141935,0.526453,0.555991) rgb(118cm)=(0.140536,0.530132,0.555659) rgb(119cm)=(0.139147,0.533812,0.555298) rgb(120cm)=(0.13777,0.537492,0.554906) rgb(121cm)=(0.136408,0.541173,0.554483) rgb(122cm)=(0.135066,0.544853,0.554029) rgb(123cm)=(0.133743,0.548535,0.553541) rgb(124cm)=(0.132444,0.552216,0.553018) rgb(125cm)=(0.131172,0.555899,0.552459) rgb(126cm)=(0.129933,0.559582,0.551864) rgb(127cm)=(0.128729,0.563265,0.551229) rgb(128cm)=(0.127568,0.566949,0.550556) rgb(129cm)=(0.126453,0.570633,0.549841) rgb(130cm)=(0.125394,0.574318,0.549086) rgb(131cm)=(0.124395,0.578002,0.548287) rgb(132cm)=(0.123463,0.581687,0.547445) rgb(133cm)=(0.122606,0.585371,0.546557) rgb(134cm)=(0.121831,0.589055,0.545623) rgb(135cm)=(0.121148,0.592739,0.544641) rgb(136cm)=(0.120565,0.596422,0.543611) rgb(137cm)=(0.120092,0.600104,0.54253) rgb(138cm)=(0.119738,0.603785,0.5414) rgb(139cm)=(0.119512,0.607464,0.540218) rgb(140cm)=(0.119423,0.611141,0.538982) rgb(141cm)=(0.119483,0.614817,0.537692) rgb(142cm)=(0.119699,0.61849,0.536347) rgb(143cm)=(0.120081,0.622161,0.534946) rgb(144cm)=(0.120638,0.625828,0.533488) rgb(145cm)=(0.12138,0.629492,0.531973) rgb(146cm)=(0.122312,0.633153,0.530398) rgb(147cm)=(0.123444,0.636809,0.528763) rgb(148cm)=(0.12478,0.640461,0.527068) rgb(149cm)=(0.126326,0.644107,0.525311) rgb(150cm)=(0.128087,0.647749,0.523491) rgb(151cm)=(0.130067,0.651384,0.521608) rgb(152cm)=(0.132268,0.655014,0.519661) rgb(153cm)=(0.134692,0.658636,0.517649) rgb(154cm)=(0.137339,0.662252,0.515571) rgb(155cm)=(0.14021,0.665859,0.513427) rgb(156cm)=(0.143303,0.669459,0.511215) rgb(157cm)=(0.146616,0.67305,0.508936) rgb(158cm)=(0.150148,0.676631,0.506589) rgb(159cm)=(0.153894,0.680203,0.504172) rgb(160cm)=(0.157851,0.683765,0.501686) rgb(161cm)=(0.162016,0.687316,0.499129) rgb(162cm)=(0.166383,0.690856,0.496502) rgb(163cm)=(0.170948,0.694384,0.493803) rgb(164cm)=(0.175707,0.6979,0.491033) rgb(165cm)=(0.180653,0.701402,0.488189) rgb(166cm)=(0.185783,0.704891,0.485273) rgb(167cm)=(0.19109,0.708366,0.482284) rgb(168cm)=(0.196571,0.711827,0.479221) rgb(169cm)=(0.202219,0.715272,0.476084) rgb(170cm)=(0.20803,0.718701,0.472873) rgb(171cm)=(0.214,0.722114,0.469588) rgb(172cm)=(0.220124,0.725509,0.466226) rgb(173cm)=(0.226397,0.728888,0.462789) rgb(174cm)=(0.232815,0.732247,0.459277) rgb(175cm)=(0.239374,0.735588,0.455688) rgb(176cm)=(0.24607,0.73891,0.452024) rgb(177cm)=(0.252899,0.742211,0.448284) rgb(178cm)=(0.259857,0.745492,0.444467) rgb(179cm)=(0.266941,0.748751,0.440573) rgb(180cm)=(0.274149,0.751988,0.436601) rgb(181cm)=(0.281477,0.755203,0.432552) rgb(182cm)=(0.288921,0.758394,0.428426) rgb(183cm)=(0.296479,0.761561,0.424223) rgb(184cm)=(0.304148,0.764704,0.419943) rgb(185cm)=(0.311925,0.767822,0.415586) rgb(186cm)=(0.319809,0.770914,0.411152) rgb(187cm)=(0.327796,0.77398,0.40664) rgb(188cm)=(0.335885,0.777018,0.402049) rgb(189cm)=(0.344074,0.780029,0.397381) rgb(190cm)=(0.35236,0.783011,0.392636) rgb(191cm)=(0.360741,0.785964,0.387814) rgb(192cm)=(0.369214,0.788888,0.382914) rgb(193cm)=(0.377779,0.791781,0.377939) rgb(194cm)=(0.386433,0.794644,0.372886) rgb(195cm)=(0.395174,0.797475,0.367757) rgb(196cm)=(0.404001,0.800275,0.362552) rgb(197cm)=(0.412913,0.803041,0.357269) rgb(198cm)=(0.421908,0.805774,0.35191) rgb(199cm)=(0.430983,0.808473,0.346476) rgb(200cm)=(0.440137,0.811138,0.340967) rgb(201cm)=(0.449368,0.813768,0.335384) rgb(202cm)=(0.458674,0.816363,0.329727) rgb(203cm)=(0.468053,0.818921,0.323998) rgb(204cm)=(0.477504,0.821444,0.318195) rgb(205cm)=(0.487026,0.823929,0.312321) rgb(206cm)=(0.496615,0.826376,0.306377) rgb(207cm)=(0.506271,0.828786,0.300362) rgb(208cm)=(0.515992,0.831158,0.294279) rgb(209cm)=(0.525776,0.833491,0.288127) rgb(210cm)=(0.535621,0.835785,0.281908) rgb(211cm)=(0.545524,0.838039,0.275626) rgb(212cm)=(0.555484,0.840254,0.269281) rgb(213cm)=(0.565498,0.84243,0.262877) rgb(214cm)=(0.575563,0.844566,0.256415) rgb(215cm)=(0.585678,0.846661,0.249897) rgb(216cm)=(0.595839,0.848717,0.243329) rgb(217cm)=(0.606045,0.850733,0.236712) rgb(218cm)=(0.616293,0.852709,0.230052) rgb(219cm)=(0.626579,0.854645,0.223353) rgb(220cm)=(0.636902,0.856542,0.21662) rgb(221cm)=(0.647257,0.8584,0.209861) rgb(222cm)=(0.657642,0.860219,0.203082) rgb(223cm)=(0.668054,0.861999,0.196293) rgb(224cm)=(0.678489,0.863742,0.189503) rgb(225cm)=(0.688944,0.865448,0.182725) rgb(226cm)=(0.699415,0.867117,0.175971) rgb(227cm)=(0.709898,0.868751,0.169257) rgb(228cm)=(0.720391,0.87035,0.162603) rgb(229cm)=(0.730889,0.871916,0.156029) rgb(230cm)=(0.741388,0.873449,0.149561) rgb(231cm)=(0.751884,0.874951,0.143228) rgb(232cm)=(0.762373,0.876424,0.137064) rgb(233cm)=(0.772852,0.877868,0.131109) rgb(234cm)=(0.783315,0.879285,0.125405) rgb(235cm)=(0.79376,0.880678,0.120005) rgb(236cm)=(0.804182,0.882046,0.114965) rgb(237cm)=(0.814576,0.883393,0.110347) rgb(238cm)=(0.82494,0.88472,0.106217) rgb(239cm)=(0.83527,0.886029,0.102646) rgb(240cm)=(0.845561,0.887322,0.099702) rgb(241cm)=(0.85581,0.888601,0.097452) rgb(242cm)=(0.866013,0.889868,0.095953) rgb(243cm)=(0.876168,0.891125,0.09525) rgb(244cm)=(0.886271,0.892374,0.095374) rgb(245cm)=(0.89632,0.893616,0.096335) rgb(246cm)=(0.906311,0.894855,0.098125) rgb(247cm)=(0.916242,0.896091,0.100717) rgb(248cm)=(0.926106,0.89733,0.104071) rgb(249cm)=(0.935904,0.89857,0.108131) rgb(250cm)=(0.945636,0.899815,0.112838) rgb(251cm)=(0.9553,0.901065,0.118128) rgb(252cm)=(0.964894,0.902323,0.123941) rgb(253cm)=(0.974417,0.90359,0.130215) rgb(254cm)=(0.983868,0.904867,0.136897) rgb(255cm)=(0.993248,0.906157,0.143936) }, colorbar]\addplot [point meta min=0.0, point meta max=1.0] graphics [xmin=0.0, xmax=40.0, ymin=-4000, ymax=4000] {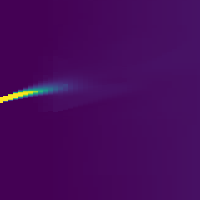};
    \end{groupplot}
    
    \end{tikzpicture}
    \caption{Comparison of lookup table probabilities of NMAC estimated using traditional MDP model checking, neural network probabilities of NMAC estimated using Monte Carlo simulations, and overapproximated neural network probabilities of NMAC using the model checking formulation in this work. The ownship vertical rate is 60 ft/s, the intruder vertical rate is 0 ft/s, and the previous advisory is $\textsc{coc}$. \label{fig:monte_carlo_ownRate}}
\end{figure}

\begin{figure}[htb]
    \begin{tikzpicture}[]
    \begin{groupplot}[group style={horizontal sep = 0.4cm, vertical sep = 2.5cm, group size=3 by 1}]
    \nextgroupplot [height = {5cm}, ylabel = {$h$ (ft)}, title = {Lookup Table}, xmin = {0.0}, xmax = {40.0}, ymax = {4000}, xlabel = {$\tau$ (s)}, ymin = {-4000}, width = {4.35cm}, enlargelimits = false, axis on top]\addplot [point meta min=0, point meta max=3] graphics [xmin=0.0, xmax=40.0, ymin=-4000, ymax=4000] {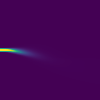};
    \nextgroupplot [height = {5cm}, title = {Monte Carlo}, xmin = {0.0}, xmax = {40.0}, ymax = {4000}, xlabel = {$\tau$ (s)}, ymin = {-4000}, width = {4.35cm}, enlargelimits = false, axis on top, yticklabels={,,}]\addplot [point meta min=0, point meta max=3] graphics [xmin=0.0, xmax=40.0, ymin=-4000, ymax=4000] {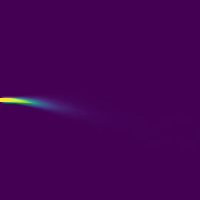};
    \nextgroupplot [height = {5cm}, title = {Model Checking}, xmin = {0.0}, xmax = {40.0}, ymax = {4000}, xlabel = {$\tau$ (s)}, ymin = {-4000}, width = {4.35cm}, enlargelimits = false, axis on top, yticklabels={,,}, colorbar style={
                width=0.3cm}, colormap={mycolormap}{ rgb(0cm)=(0.267004,0.004874,0.329415) rgb(1cm)=(0.26851,0.009605,0.335427) rgb(2cm)=(0.269944,0.014625,0.341379) rgb(3cm)=(0.271305,0.019942,0.347269) rgb(4cm)=(0.272594,0.025563,0.353093) rgb(5cm)=(0.273809,0.031497,0.358853) rgb(6cm)=(0.274952,0.037752,0.364543) rgb(7cm)=(0.276022,0.044167,0.370164) rgb(8cm)=(0.277018,0.050344,0.375715) rgb(9cm)=(0.277941,0.056324,0.381191) rgb(10cm)=(0.278791,0.062145,0.386592) rgb(11cm)=(0.279566,0.067836,0.391917) rgb(12cm)=(0.280267,0.073417,0.397163) rgb(13cm)=(0.280894,0.078907,0.402329) rgb(14cm)=(0.281446,0.08432,0.407414) rgb(15cm)=(0.281924,0.089666,0.412415) rgb(16cm)=(0.282327,0.094955,0.417331) rgb(17cm)=(0.282656,0.100196,0.42216) rgb(18cm)=(0.28291,0.105393,0.426902) rgb(19cm)=(0.283091,0.110553,0.431554) rgb(20cm)=(0.283197,0.11568,0.436115) rgb(21cm)=(0.283229,0.120777,0.440584) rgb(22cm)=(0.283187,0.125848,0.44496) rgb(23cm)=(0.283072,0.130895,0.449241) rgb(24cm)=(0.282884,0.13592,0.453427) rgb(25cm)=(0.282623,0.140926,0.457517) rgb(26cm)=(0.28229,0.145912,0.46151) rgb(27cm)=(0.281887,0.150881,0.465405) rgb(28cm)=(0.281412,0.155834,0.469201) rgb(29cm)=(0.280868,0.160771,0.472899) rgb(30cm)=(0.280255,0.165693,0.476498) rgb(31cm)=(0.279574,0.170599,0.479997) rgb(32cm)=(0.278826,0.17549,0.483397) rgb(33cm)=(0.278012,0.180367,0.486697) rgb(34cm)=(0.277134,0.185228,0.489898) rgb(35cm)=(0.276194,0.190074,0.493001) rgb(36cm)=(0.275191,0.194905,0.496005) rgb(37cm)=(0.274128,0.199721,0.498911) rgb(38cm)=(0.273006,0.20452,0.501721) rgb(39cm)=(0.271828,0.209303,0.504434) rgb(40cm)=(0.270595,0.214069,0.507052) rgb(41cm)=(0.269308,0.218818,0.509577) rgb(42cm)=(0.267968,0.223549,0.512008) rgb(43cm)=(0.26658,0.228262,0.514349) rgb(44cm)=(0.265145,0.232956,0.516599) rgb(45cm)=(0.263663,0.237631,0.518762) rgb(46cm)=(0.262138,0.242286,0.520837) rgb(47cm)=(0.260571,0.246922,0.522828) rgb(48cm)=(0.258965,0.251537,0.524736) rgb(49cm)=(0.257322,0.25613,0.526563) rgb(50cm)=(0.255645,0.260703,0.528312) rgb(51cm)=(0.253935,0.265254,0.529983) rgb(52cm)=(0.252194,0.269783,0.531579) rgb(53cm)=(0.250425,0.27429,0.533103) rgb(54cm)=(0.248629,0.278775,0.534556) rgb(55cm)=(0.246811,0.283237,0.535941) rgb(56cm)=(0.244972,0.287675,0.53726) rgb(57cm)=(0.243113,0.292092,0.538516) rgb(58cm)=(0.241237,0.296485,0.539709) rgb(59cm)=(0.239346,0.300855,0.540844) rgb(60cm)=(0.237441,0.305202,0.541921) rgb(61cm)=(0.235526,0.309527,0.542944) rgb(62cm)=(0.233603,0.313828,0.543914) rgb(63cm)=(0.231674,0.318106,0.544834) rgb(64cm)=(0.229739,0.322361,0.545706) rgb(65cm)=(0.227802,0.326594,0.546532) rgb(66cm)=(0.225863,0.330805,0.547314) rgb(67cm)=(0.223925,0.334994,0.548053) rgb(68cm)=(0.221989,0.339161,0.548752) rgb(69cm)=(0.220057,0.343307,0.549413) rgb(70cm)=(0.21813,0.347432,0.550038) rgb(71cm)=(0.21621,0.351535,0.550627) rgb(72cm)=(0.214298,0.355619,0.551184) rgb(73cm)=(0.212395,0.359683,0.55171) rgb(74cm)=(0.210503,0.363727,0.552206) rgb(75cm)=(0.208623,0.367752,0.552675) rgb(76cm)=(0.206756,0.371758,0.553117) rgb(77cm)=(0.204903,0.375746,0.553533) rgb(78cm)=(0.203063,0.379716,0.553925) rgb(79cm)=(0.201239,0.38367,0.554294) rgb(80cm)=(0.19943,0.387607,0.554642) rgb(81cm)=(0.197636,0.391528,0.554969) rgb(82cm)=(0.19586,0.395433,0.555276) rgb(83cm)=(0.1941,0.399323,0.555565) rgb(84cm)=(0.192357,0.403199,0.555836) rgb(85cm)=(0.190631,0.407061,0.556089) rgb(86cm)=(0.188923,0.41091,0.556326) rgb(87cm)=(0.187231,0.414746,0.556547) rgb(88cm)=(0.185556,0.41857,0.556753) rgb(89cm)=(0.183898,0.422383,0.556944) rgb(90cm)=(0.182256,0.426184,0.55712) rgb(91cm)=(0.180629,0.429975,0.557282) rgb(92cm)=(0.179019,0.433756,0.55743) rgb(93cm)=(0.177423,0.437527,0.557565) rgb(94cm)=(0.175841,0.44129,0.557685) rgb(95cm)=(0.174274,0.445044,0.557792) rgb(96cm)=(0.172719,0.448791,0.557885) rgb(97cm)=(0.171176,0.45253,0.557965) rgb(98cm)=(0.169646,0.456262,0.55803) rgb(99cm)=(0.168126,0.459988,0.558082) rgb(100cm)=(0.166617,0.463708,0.558119) rgb(101cm)=(0.165117,0.467423,0.558141) rgb(102cm)=(0.163625,0.471133,0.558148) rgb(103cm)=(0.162142,0.474838,0.55814) rgb(104cm)=(0.160665,0.47854,0.558115) rgb(105cm)=(0.159194,0.482237,0.558073) rgb(106cm)=(0.157729,0.485932,0.558013) rgb(107cm)=(0.15627,0.489624,0.557936) rgb(108cm)=(0.154815,0.493313,0.55784) rgb(109cm)=(0.153364,0.497,0.557724) rgb(110cm)=(0.151918,0.500685,0.557587) rgb(111cm)=(0.150476,0.504369,0.55743) rgb(112cm)=(0.149039,0.508051,0.55725) rgb(113cm)=(0.147607,0.511733,0.557049) rgb(114cm)=(0.14618,0.515413,0.556823) rgb(115cm)=(0.144759,0.519093,0.556572) rgb(116cm)=(0.143343,0.522773,0.556295) rgb(117cm)=(0.141935,0.526453,0.555991) rgb(118cm)=(0.140536,0.530132,0.555659) rgb(119cm)=(0.139147,0.533812,0.555298) rgb(120cm)=(0.13777,0.537492,0.554906) rgb(121cm)=(0.136408,0.541173,0.554483) rgb(122cm)=(0.135066,0.544853,0.554029) rgb(123cm)=(0.133743,0.548535,0.553541) rgb(124cm)=(0.132444,0.552216,0.553018) rgb(125cm)=(0.131172,0.555899,0.552459) rgb(126cm)=(0.129933,0.559582,0.551864) rgb(127cm)=(0.128729,0.563265,0.551229) rgb(128cm)=(0.127568,0.566949,0.550556) rgb(129cm)=(0.126453,0.570633,0.549841) rgb(130cm)=(0.125394,0.574318,0.549086) rgb(131cm)=(0.124395,0.578002,0.548287) rgb(132cm)=(0.123463,0.581687,0.547445) rgb(133cm)=(0.122606,0.585371,0.546557) rgb(134cm)=(0.121831,0.589055,0.545623) rgb(135cm)=(0.121148,0.592739,0.544641) rgb(136cm)=(0.120565,0.596422,0.543611) rgb(137cm)=(0.120092,0.600104,0.54253) rgb(138cm)=(0.119738,0.603785,0.5414) rgb(139cm)=(0.119512,0.607464,0.540218) rgb(140cm)=(0.119423,0.611141,0.538982) rgb(141cm)=(0.119483,0.614817,0.537692) rgb(142cm)=(0.119699,0.61849,0.536347) rgb(143cm)=(0.120081,0.622161,0.534946) rgb(144cm)=(0.120638,0.625828,0.533488) rgb(145cm)=(0.12138,0.629492,0.531973) rgb(146cm)=(0.122312,0.633153,0.530398) rgb(147cm)=(0.123444,0.636809,0.528763) rgb(148cm)=(0.12478,0.640461,0.527068) rgb(149cm)=(0.126326,0.644107,0.525311) rgb(150cm)=(0.128087,0.647749,0.523491) rgb(151cm)=(0.130067,0.651384,0.521608) rgb(152cm)=(0.132268,0.655014,0.519661) rgb(153cm)=(0.134692,0.658636,0.517649) rgb(154cm)=(0.137339,0.662252,0.515571) rgb(155cm)=(0.14021,0.665859,0.513427) rgb(156cm)=(0.143303,0.669459,0.511215) rgb(157cm)=(0.146616,0.67305,0.508936) rgb(158cm)=(0.150148,0.676631,0.506589) rgb(159cm)=(0.153894,0.680203,0.504172) rgb(160cm)=(0.157851,0.683765,0.501686) rgb(161cm)=(0.162016,0.687316,0.499129) rgb(162cm)=(0.166383,0.690856,0.496502) rgb(163cm)=(0.170948,0.694384,0.493803) rgb(164cm)=(0.175707,0.6979,0.491033) rgb(165cm)=(0.180653,0.701402,0.488189) rgb(166cm)=(0.185783,0.704891,0.485273) rgb(167cm)=(0.19109,0.708366,0.482284) rgb(168cm)=(0.196571,0.711827,0.479221) rgb(169cm)=(0.202219,0.715272,0.476084) rgb(170cm)=(0.20803,0.718701,0.472873) rgb(171cm)=(0.214,0.722114,0.469588) rgb(172cm)=(0.220124,0.725509,0.466226) rgb(173cm)=(0.226397,0.728888,0.462789) rgb(174cm)=(0.232815,0.732247,0.459277) rgb(175cm)=(0.239374,0.735588,0.455688) rgb(176cm)=(0.24607,0.73891,0.452024) rgb(177cm)=(0.252899,0.742211,0.448284) rgb(178cm)=(0.259857,0.745492,0.444467) rgb(179cm)=(0.266941,0.748751,0.440573) rgb(180cm)=(0.274149,0.751988,0.436601) rgb(181cm)=(0.281477,0.755203,0.432552) rgb(182cm)=(0.288921,0.758394,0.428426) rgb(183cm)=(0.296479,0.761561,0.424223) rgb(184cm)=(0.304148,0.764704,0.419943) rgb(185cm)=(0.311925,0.767822,0.415586) rgb(186cm)=(0.319809,0.770914,0.411152) rgb(187cm)=(0.327796,0.77398,0.40664) rgb(188cm)=(0.335885,0.777018,0.402049) rgb(189cm)=(0.344074,0.780029,0.397381) rgb(190cm)=(0.35236,0.783011,0.392636) rgb(191cm)=(0.360741,0.785964,0.387814) rgb(192cm)=(0.369214,0.788888,0.382914) rgb(193cm)=(0.377779,0.791781,0.377939) rgb(194cm)=(0.386433,0.794644,0.372886) rgb(195cm)=(0.395174,0.797475,0.367757) rgb(196cm)=(0.404001,0.800275,0.362552) rgb(197cm)=(0.412913,0.803041,0.357269) rgb(198cm)=(0.421908,0.805774,0.35191) rgb(199cm)=(0.430983,0.808473,0.346476) rgb(200cm)=(0.440137,0.811138,0.340967) rgb(201cm)=(0.449368,0.813768,0.335384) rgb(202cm)=(0.458674,0.816363,0.329727) rgb(203cm)=(0.468053,0.818921,0.323998) rgb(204cm)=(0.477504,0.821444,0.318195) rgb(205cm)=(0.487026,0.823929,0.312321) rgb(206cm)=(0.496615,0.826376,0.306377) rgb(207cm)=(0.506271,0.828786,0.300362) rgb(208cm)=(0.515992,0.831158,0.294279) rgb(209cm)=(0.525776,0.833491,0.288127) rgb(210cm)=(0.535621,0.835785,0.281908) rgb(211cm)=(0.545524,0.838039,0.275626) rgb(212cm)=(0.555484,0.840254,0.269281) rgb(213cm)=(0.565498,0.84243,0.262877) rgb(214cm)=(0.575563,0.844566,0.256415) rgb(215cm)=(0.585678,0.846661,0.249897) rgb(216cm)=(0.595839,0.848717,0.243329) rgb(217cm)=(0.606045,0.850733,0.236712) rgb(218cm)=(0.616293,0.852709,0.230052) rgb(219cm)=(0.626579,0.854645,0.223353) rgb(220cm)=(0.636902,0.856542,0.21662) rgb(221cm)=(0.647257,0.8584,0.209861) rgb(222cm)=(0.657642,0.860219,0.203082) rgb(223cm)=(0.668054,0.861999,0.196293) rgb(224cm)=(0.678489,0.863742,0.189503) rgb(225cm)=(0.688944,0.865448,0.182725) rgb(226cm)=(0.699415,0.867117,0.175971) rgb(227cm)=(0.709898,0.868751,0.169257) rgb(228cm)=(0.720391,0.87035,0.162603) rgb(229cm)=(0.730889,0.871916,0.156029) rgb(230cm)=(0.741388,0.873449,0.149561) rgb(231cm)=(0.751884,0.874951,0.143228) rgb(232cm)=(0.762373,0.876424,0.137064) rgb(233cm)=(0.772852,0.877868,0.131109) rgb(234cm)=(0.783315,0.879285,0.125405) rgb(235cm)=(0.79376,0.880678,0.120005) rgb(236cm)=(0.804182,0.882046,0.114965) rgb(237cm)=(0.814576,0.883393,0.110347) rgb(238cm)=(0.82494,0.88472,0.106217) rgb(239cm)=(0.83527,0.886029,0.102646) rgb(240cm)=(0.845561,0.887322,0.099702) rgb(241cm)=(0.85581,0.888601,0.097452) rgb(242cm)=(0.866013,0.889868,0.095953) rgb(243cm)=(0.876168,0.891125,0.09525) rgb(244cm)=(0.886271,0.892374,0.095374) rgb(245cm)=(0.89632,0.893616,0.096335) rgb(246cm)=(0.906311,0.894855,0.098125) rgb(247cm)=(0.916242,0.896091,0.100717) rgb(248cm)=(0.926106,0.89733,0.104071) rgb(249cm)=(0.935904,0.89857,0.108131) rgb(250cm)=(0.945636,0.899815,0.112838) rgb(251cm)=(0.9553,0.901065,0.118128) rgb(252cm)=(0.964894,0.902323,0.123941) rgb(253cm)=(0.974417,0.90359,0.130215) rgb(254cm)=(0.983868,0.904867,0.136897) rgb(255cm)=(0.993248,0.906157,0.143936) }, colorbar]\addplot [point meta min=0.0, point meta max=1.0] graphics [xmin=0.0, xmax=40.0, ymin=-4000, ymax=4000] {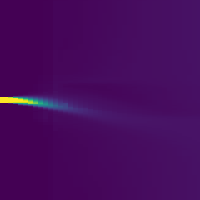};
    \end{groupplot}
    
    \end{tikzpicture}
    \caption{Comparison of lookup table probabilities of NMAC estimated using traditional MDP model checking, neural network probabilities of NMAC estimated using Monte Carlo simulations, and overapproximated neural network probabilities of NMAC using the model checking formulation in this work. The ownship and intruder vertical rates are both 60 ft/s, and the previous advisory is $\textsc{coc}$. \label{fig:monte_carlo_bothRate}}
\end{figure}

We also examined the probability of NMAC when the aircraft are climbing or descending. \Cref{fig:monte_carlo_ownRate} contains the results when the ownship is climbing at 60 ft/s in comparison to the table and Monte Carlo results, and \cref{fig:monte_carlo_bothRate} shows the vertical rates when both aircraft are climbing at a rate of 60 ft/s. The trends in the probabilities match the trends seen in the lookup table and Monte Carlo results. In \cref{fig:monte_carlo_ownRate}, the region of high probability of NMAC extends above the ownship because it starts in a climb.

When both aircraft are climbing at the same rate in \cref{fig:monte_carlo_bothRate}, the two aircraft are on a direct collision path when they are co-altitude. For this reason, probabilities are high in this region when $\tau$ is small. As the time to collision increases, the region shifts downward slightly since the intruder is not executing collision avoidance maneuvers and is therefore most dangerous at altitudes below the ownship. The overapproximated probabilities are again higher than the Monte Carlo probabilities, but the maximum probability of NMAC with these starting states decreases slightly to 0.0462 and 0.0437, respectively. 

The maximum probability of NMAC among all states at $\tau = 40$ is 0.0605 and occurs in the cell in which the intruder is descending at a vertical rate between 77.34 and 78.13 ft/s and is located between 656.25 and 664.06 ft below the ownship. The ownship is also descending rapidly between 99.22 and 100.0 ft/s. Therefore, we can guarantee that the probability of NMAC with respect to the dynamics model outlined in this work when following the neural network policy is less than 0.0605. Tuning parameters such as the online splitting thresholds and minimum cell size allowed us to tighten our bound on the probability. The full scale analysis was run on 9 cores of an Intel Xeon 2.20 GHz CPU and took approximately 10 days. With more computational resources, it is likely that we could tighten this bound further.


\section{Related Work}
The complexity of deep neural networks makes their performance difficult to verify, and previous research has addressed this problem from multiple perspectives. One perspective involves decreasing the complexity of deep neural networks by extracting simpler policies that are easier to verify \citep{bastani2018verifiable, koul2018learning, carr2020verifiable}. For instance, \citet{bastani2018verifiable} outline a method to learn decision tree policies from neural networks trained used deep reinforcement learning. Other works focus on extracting finite state controllers from recurrent neural network policies \citep{koul2018learning, carr2020verifiable}. While these works simplify the verification problem, the extracted policies may be less effective and more difficult to store. In fact, \Citet{Julian2019jgcd} found that neural network policies performed better and required significantly less storage than decision trees when applied to the aircraft collision avoidance problem.

Another perspective on ensuring the safety of neural networks involves verifying their policies directly without the need to extract a simpler policy. Recent work has used formal methods to verify input-output properties of neural networks \citep{Katz2017, katz2019marabou, reluval, tjeng2017evaluating}. \Citet{liu2019algorithms} provide an overview of neural verification methodologies that incorporate techniques from reachability, optimization, and search. In the context of neural network controllers, these tools can provide guarantees on the actions that the network could output in a given region of the state space. This methodology was used to prove properties of neural networks trained on an early prototype of ACAS Xu, the version of ACAS X developed for unmanned aircraft \citep{Katz2017, owen2019acas}. An example property that should hold for the ACAS Xu networks is that the system should always issue an alert if an intruder is directly in front of the ownship. \Citet{Katz2017} use the Reluplex algorithm for neural network verification to prove this property, along with a number of other intuitive properties that must hold for safe operation. Other verification tools have proven the same properties as a benchmark \citep{reluval, liu2019algorithms}.

While neural network verification tools provide guarantees on the performance of neural network controllers at a single instance in time, other work has extended these tools to evaluate their performance over time using a model of the system dynamics. \Citet{Julian2019vnn} define ``safeable'' regions for each action, in which even the worst case trajectory created by following the action could still be safe. This worst-case analysis, however, does not consider whether states are reachable when following the neural network policy, so states that it flags as unsafe may never be reached in the first place.

To better approximate the true closed-loop system, previous work has used various forms of reachability analysis \citep{Julian2019reach, Julian2019dasc, huang2019reachnn, xiang2018reachability, xiang2018reachable, xiang2019reachable, dutta2019reachability, ivanov2019verisig, claviere2020safety, manzanas2021verification}. These reachability approaches typically involve combining work in neural network verification with existing reachability methods from fields such as ordinary differential equations and hybrid systems. Other approaches involve combining the neural network controller with a neural network representation of the dynamics and performing the neural network verification on the entire closed-loop system \citep{Sidrane2019iclr, akintunde2018reachability}.

Previous works have applied reachability analysis to evaluate the closed-loop performance of aircraft collision avoidance neural networks \citep{manzanas2021verification, claviere2020safety, Julian2019dasc, akintunde2020formal}. While \citet{manzanas2021verification} and \citet{claviere2020safety} are able to show that the aircraft will not collide in a given set of scenarios, they do not take into account any uncertainty in the dynamics of either aircraft. \Citet{Julian2019dasc} and \citet{akintunde2020formal} account for uncertainty in the dynamics by allowing the ownship and intruder to follow a range of accelerations. While these approaches are able to provide deterministic guarantees that the aircraft will not collide, they require strong assumptions on the dynamics of the aircraft for the conclusion to hold. As these assumptions are relaxed, there is likely to exist a set of adversarial intruder and ownship accelerations that could result in a collision. Even though the aircraft are unlikely to follow this acceleration pattern, there is still a nonzero probability that they will follow it, and the reachability analysis would conclude that the system is unsafe.

In this work, we adapt methods in probabilistic model checking to provide probabilistic guarantees for neural network controllers \citep{baier2008principles, lahijanian2011control, Bouton2020aaai, d2001reachability}. There has been extensive previous work on model checking for finite state MDPs \citep{baier2008principles, lahijanian2011control}. MDP model checking can be used to determine the probability of satisfying an LTL formula when following a given policy from any state in the state space. However, these methods are limited to problems with discrete states and cannot be applied directly to continuous neural network policies. This work therefore extends these methods to handle continuous states. To provide guarantees, we ensure that the probability estimates using our method represent an overapproximation of the true probability of failure.

\section{Conclusion}
In this work, we have introduced an approach to generate probabilistic safety guarantees on a neural network controller and applied it to an open source collision avoidance system inspired by the ACAS X neural networks. We demonstrated how techniques in MDP model checking could be applied to verify an overapproximated neural network policy obtained using a neural network verification tool. We identified the major sources of overapproximation error in the model checking process and developed both offline and online error reduction techniques to address them.

Our adaptive verification technique efficiently partitions the input space to obtain an overapproximated neural network policy that minimizes overapproximation error. This method can be used outside of the context of model checking to analyze neural network policies and detect decision boundaries. By combining the adaptive verification results with online splitting heuristics inspired by MDP state abstraction, we are able to provide meaningful probabilistic safety guarantees that follow trends shown in both Monte Carlo analysis and model checking analysis performed on the lookup table that the neural network approximates.

The probabilistic dynamics model used in this work represents a conservative aircraft response model. In the future, other aircraft response models will be analyzed to determine the effect of the model selection on the probabilities. Furthermore, the results can be used to determine unsafe areas of the state space where the neural network policy may need adjustments. For example, at the outset of this work, the model checking method was able to easily detect a bounds error in the neural network policy generation that may have been otherwise easy to overlook. Although we used the method to determine the probability of an NMAC in this work, the general formulation allows us to determine the probability of satisfying any LTL specification. Future work will explore both generating guarantees on other aspects of the aircraft collision avoidance problem as well as for other safety-critical applications such as autonomous driving. The methodology presented here represents a step toward the ability to verify the performance of neural network controllers for use in safety-critical environments.

\section*{Declarations}
\textbf{Funding:} This research was supported by National Science Foundation Graduate Research Fellowship under Grant No. DGE--1656518. Any opinion, findings, and conclusions or recommendations expressed in this material are those of the authors and do not necessarily reflect the views of the National Science Foundation.
\\ \\
\textbf{Conflicts of interest/Competing interests:} The authors have no conflicts of interest to declare that are relevant to the content of this article.
\\ \\
\textbf{Availability of data and material:} The neural networks used in this research can be found in the ``networks'' folder of the repository located at \url{https://github.com/sisl/AdaptiveVerification}.
\\ \\
\textbf{Code availability:} The code for the adaptive verification portion of the work can be found at \url{https://github.com/sisl/AdaptiveVerification}, and the code for the model checking is located at \url{https://github.com/sisl/NeuralModelChecking}. The repository used to generate the networks used in this work is at \url{https://github.com/sisl/VerticalCAS}.

\section*{Conflict of interest}

The authors declare that they have no conflict of interest.

\bibliographystyle{spbasic}      
\bibliography{refs}   


\end{document}